\documentclass[acmsmall,prologue,dvipsnames]{acmart}

\AtBeginDocument{%
  \providecommand\BibTeX{{%
    \normalfont B\kern-0.5em{\scshape i\kern-0.25em b}\kern-0.8em\TeX}}}

\usepackage{caption}
\usepackage{array}
\usepackage{multirow}
\usepackage{graphicx}
\usepackage{tablefootnote}
\usepackage[flushleft]{threeparttable}
\usepackage[lofdepth,lotdepth]{subfig}
\newcommand{\symfootnote}[1]{%
\let\oldthefootnote=\thefootnote%
\stepcounter{mpfootnote}%
\addtocounter{footnote}{-1}%
\renewcommand{\thefootnote}{\fnsymbol{mpfootnote}}%
\footnote{#1}%
\let\thefootnote=\oldthefootnote%
}

\makeatletter
\DeclareRobustCommand{\cev}[1]{%
  {\mathpalette\do@cev{#1}}%
}
\newcommand{\do@cev}[2]{%
  \vbox{\offinterlineskip
    \sbox\z@{$\m@th#1 x$}%
    \ialign{##\cr
      \hidewidth\reflectbox{$\m@th#1\vec{}\mkern4mu$}\hidewidth\cr
      \noalign{\kern-\ht\z@}
      $\m@th#1#2$\cr
    }%
  }%
}
\makeatother

\definecolor{applegreen}{rgb}{0.25, 0.40, 0.0}

\setcopyright{acmcopyright}
\copyrightyear{2018}
\acmYear{2018}
\acmDOI{10.1145/1122445.1122456}

\acmJournal{JACM}
\acmVolume{37}
\acmNumber{4}
\acmArticle{111}
\acmMonth{8}

\begin{document}

\title{A Survey on Multi-modal Summarization}


\author{Anubhav Jangra}
\email{anubhav0603@gmail.com}
\orcid{0001-5571-6098}
\affiliation{
  \institution{Department of Computer Science, Indian Institute of Technology Patna}
  \city{Patna}
  \state{Bihar}
  \country{India}
  \postcode{801106}
}

\author{Sourajit Mukherjee}
\email{mailsourajit25@gmail.com}
\affiliation{
  \institution{Department of Mathematics, Indian Institute of Technology Patna}
  \city{Patna}
  \state{Bihar}
  \country{India}
}

\author{Adam Jatowt}
\affiliation{
  \institution{Department of Informatics \& DiSC, University of Innsbruck}
  \city{Innsbruck}
  \country{Austria}}
\email{jatowt@acm.org}

\author{Sriparna Saha}
\affiliation{
  \institution{Department of Computer Science, Indian Institute of Technology Patna}
  \city{Patna}
  \state{Bihar}
  \country{India}
}
\email{sriparna.saha@gmail.com}

\author{Mohammad Hasanuzzaman}
\affiliation{%
 \institution{Department of Computer Science, Cork Institute of Technology}
 \city{Bishopstown}
 \state{Cork}
 \country{Ireland}}

\renewcommand{\shortauthors}{Jangra et al.}

\begin{abstract} 
The new era of technology has brought us to the point where it is convenient for people to share their opinions
over an abundance of platforms. These platforms have a provision for the users to express themselves in
multiple forms of representations, including text, images, videos, and audio. This, however, makes it difficult for
users to obtain all the key information about a topic, making the task of automatic multi-modal summarization
(MMS) essential. In this paper, we present a comprehensive survey of the existing research in the area of MMS, covering various modalities like text, image, audio, and video. Apart from highlighting the different evaluation metrics and datasets used for the MMS task, our work also discusses the current challenges and future directions in this field. 
\end{abstract}



\maketitle

\section{Introduction} \label{sec:intro}
Everyday, the Internet is flooded with tons of new information coming from multiple sources. Due to the technological advancements, people can now share information in multiple formats with various modes of communication to be used at their disposal. This alarmingly increasing amount of content on the Internet makes it difficult for the users to receive useful information from the torrent of sources, necessitating research on the task of multi-modal summarization (MMS). Various studies have shown that including multi-modal data as input can indeed help improve the summary quality \cite{jangra2020text,li2017multi}. \citet{zhu2018msmo} claimed that on an average having a pictorial summary can improve the user satisfaction by 12.4\% over a plain text summary. The fact that nearly every content sharing platform has a provision to accompany an opinion or fact in multiple media forms, and every mobile phone has the feature to deliver that kind of facility are indicative of the superiority of a multi-modal means of communication in terms of ease in conveying and understanding information.

Information in the form of multi-modal inputs has been leveraged in many tasks other than summarization including multi-modal machine translation \cite{specia2018multi, caglayan2019probing, huang2016attention, elliott2018adversarial, elliott2017findings}, multi-modal movement prediction \cite{wang2018ajile, kirchner2014multimodal, cui2019multimodal}, multi-modal question answering \cite{singh2021mimoqa}, multi-modal lexico-semantic classification \cite{jha2022combining}, multi-modal keyword extraction \cite{verma2022maked}, product classification in e-commerce \cite{zahavy2016picture}, multi-modal interactive artificial intelligence frameworks \cite{kim2018smilee}, multi-modal emoji prediction \cite{barbieri2018multimodal, coman2018predicting}, multi-modal frame identification \cite{botschen2018multimodal}, multi-modal financial risk forecasting \cite{sawhney2020multimodal, li2020maec},  multi-modal sentiment analysis \cite{yadav2020deep, morency2011towards, rosas2013multimodal},  multi-modal named identity recognition \cite{moon2018multimodal, arshad2019aiding, zhang2018adaptive, moon2018multimodalb, yu2020improving, suman2020pay}, multi-modal video description generation \cite{ramanishka2016multimodal, hori2017attention, hori2018multimodal}, multi-modal product title compression \cite{miao2020multi} and multi-modal biometric authentication \cite{snelick2005large, fierrez2005discriminative, indovina2003multimodal}. The shear number of application possibilities for multi-modal information processing and retrieval tasks are quite impressive. Research on multi-modality can also be utilized in other closely related research problems like image-captioning \cite{chen2019news, chen2020news}, image-to-image translation \cite{huang2018multimodal}, seismic pavement testing \cite{ryden2004multimodal}, aesthetic assessment \cite{zhang2014perception, kostoulas2017films, liu2020aesthetic}, and visual question-answering \cite{kim2016multimodal}.


Text summarization is one of the oldest problems in the fields of natural language processing (NLP) and information retrieval (IR), that has attracted various researchers due to its challenging nature and potential for many applications. Research on text summarization can be traced back to more than six decades in the past \cite{luhn1958automatic}. The NLP and IR community have tackled research in text summarization for multiple applications by developing myriad of techniques and model architectures \cite{DBLP:journals/corr/SeeLM17, chen2018fast, jangra2020semantic, liu2022brio}. As an extension to this, the problem of multi-modal summarization adds another angle by incorporating visual and aural aspects into the mix, making the task more challenging and interesting to tackle. This extension of incorporating multiple modalities into a summarization problem expands the breadth of the problem, leading to wider application range for the task. 
In recent years, multi-modal summarization has experienced many new developments, including release of new datasets, advancements in techniques to tackle the MMS task, as well as proposals of more appropriate evaluation metrics. The idea of multi-modal summarization is a rather flexible one, embracing a broad range of possibilities for the input and output modalities, and also making it difficult to apprehend existing works on the MMS task with knowledge of uni-modal summarization techniques alone. This necessitates a survey on multi-modal summarization.

The MMS task, just like any uni-modal summarization task, is a demanding one, and existence of multiple correct solutions makes it very challenging. Humans creating a multi-modal summary have to use their prior understanding and external knowledge to produce the content. Establishing computer systems to mimic this behaviour becomes difficult given their inherent lack of human perception and knowledge, making the problem of automatic multi-modal summarization a non-trivial but interesting task.

Although quite a few survey papers were written for uni-modal summarization tasks including surveys on text summarization \cite{yao2017recent, gambhir2017recent, tas2007survey, nenkova2012survey, gupta2010survey, jain2022survey} and video summarization \cite{kini2019survey, sebastian2015survey, money2008video, hussain2020comprehensive, basavarajaiah2019survey}, and a few survey papers covering multi-modal research \cite{baltruvsaitis2018multimodal, soleymani2017survey, atrey2010multimodal, jaimes2007multimodal, ramachandram2017deep, sebe2005multimodal}. However, to the best of our knowledge, we are the first to present a survey on multi-modal summarization.  The closest work to ours is the work on multi-dimensional summarization by \citet{zhuge2016multi}, who proposes the method for  summarization of things in cyber-physical society through a multi-dimensional lens of semantic computing. However, our survey is distinct from that work as \citet{zhuge2016multi} focuses on how understanding human behaviour, psychology, and advances in cognitive sciences can help to improve the current summarization systems in the emerging cyber-physical society while in this manuscript we mostly focus on the direct applications and techniques adopted by the research community to tackle the MMS task.
Through this manuscript, we unify and systematize the information presented in related works, including the datasets, methodology, and evaluation techniques. With this survey, we aim to assist researchers familiarize with various techniques and resources available to proceed with research in the area of multi-modal summarization.


The rest of the paper is structured as follows: We formally define the MMS task in Section \ref{sec:mms}. In Section \ref{sec:seg}, we provide an extensive organization of existing works. In Section \ref{sec:method}, we give an overview about the techniques used for the task of MMS. In Section \ref{sec:data} we introduce the datasets available for the MMS task and evaluation techniques devised for the evaluation of multi-modal summaries, respectively. We discuss about possibilities of future work in Section \ref{sec:future} and conclude our paper in Section \ref{sec:conc}.

\section{Multi-modal Summarization task} \label{sec:mms}
In this section we formally define what classifies as a multi-modal summarization task. Before formalizing the multi-modal summarization we broadly define the term summarization\footnote{In this paper, summarization stands for automatic summarization unless specified otherwise.}. According to Wikipedia\footnote{\url{https://en.wikipedia.org/wiki/Automatic_summarization}}, automatic summarization is ``\textit{the process of shortening a set of data computationally, to create an abstract that represents the most important or relevant information within the original content}.'' Formally, summarization is the process of obtaining the set $X_{sum} = f(D)$ such that $length(X_{sum}) \leq length(D)$, where $X_{sum}$ is the output summary, $D$ is the input data, and function $f(.)$ is the summarization function. 

The multi-modal summarization task can be defined as a summarization task that takes more than one mode of information representation (termed as modality) as input, and depends on information sharing across different modalities to generate the final summary. Mathematically speaking, when the input dataset $D$ can be broken down into several partially disjoint sets of different modality information $\{M_1 \bigcup M_2 \bigcup ... \bigcup M_n\}$, where $n \geq 2$ and $\exists$ several pairs of $(M_i, M_j)$ for $(i,j) \in \{1,..,n\}$ such that the shared latent information between $(M_i, M_j)$ is not $\varnothing$ , then the task of obtaining the set $X_{sum} = f(D)$ is known as multi-modal summarization\footnote{The reason for restricting $n \geq 2$ for the task definition is limitation of current techniques, that are unable to successfully generate modalities other than text for multi-modal summarization. Even though there have been some recent breakthroughs in text-to-image generation (like Open AI's DALL-E \cite{ramesh2021zeroshot}), and text-to-speech synthesis (like Google's Duplex \cite{leviathan2018google}); they still lack the level of integrity and robustness to be used in a real-world application like MMS.}. If $n' \geq 2$ for $X_{sum} = \{M'_1 \bigcup M'_2 \bigcup ... \bigcup M'_{n'}\}$, then the output summary is multi-modal; otherwise, the output is a uni-modal summary. 

In this survey, we mainly focus on recent works that have natural language as the \textit{central modality}\footnote{We believe that the MMS models that have video as the \textit{central modality} tend to be closely related to the task of video summarization.}, where a \textit{central modality} (or \textit{key modality}) is selected according to the intuition: \textit{''For any information processing task in multi-modal scenarios, including content summarization, amongst all the modalities, there is often a preferable mode of representation based on the significance and ability to fulfill the task''} \cite{jangra2021multimodal}. Other modalities that aid the central modality to convey information are termed as \textit{adjacent modalities}.

\noindent\textbf{Various aspects of multi-modal summarization:} Literature has explored the MMS task for myriad of reasons and motives, and doing so, has lead to different challenges and variants of the task. Some of the most prominent and interesting ones are discussed below: 
\begin{itemize}
    \item \textbf{Combined complementary-supplementary multi-modal summarization task (CCS-MMS) \cite{jangra2021multimodal}:} \citet{jangra2021multimodal} proposed the CCS-MMS task of generating a multi-modal summary that considers text as the central modality, and images, audio and videos as the adjacent modality. The task is to generate the multi-modal summary such that it consists of both supplementary and complementary enhancements, which are defined as follows:
    \begin{itemize}
        \item \textbf{Supplementary enhancement:} When the adjacent modalities reinforce the facts and ideas presented in the central modality, the adjacent modalities are termed as \textit{supplementary enhancements}.
        \item \textbf{Complementary enhancement:} When the adjacent modalities complete the information by providing additional but relevant information that is not covered in the central modality, the adjacent modalities are termed as \textit{complementary enhancements}.
    \end{itemize}
    
    \item \textbf{Summarization objectives:} We can distinguish prior work based on summarization objectives they have used. For instance, \citet{li2017multi} uses weighted sum of three sub-modular objective functions to create an extractive text summarization system that is guided by multi-modal inputs. The chosen submodular functions are - salience of input text, image information covered by text summary, and non-redundancy in input text. \citet{jangra2020text} uses a single objective function for an ILP setup, that is the weighted average of uni-modal salience, and cross-modal correspondence. \citet{jangra2020multimodal} proposes two different sets of multi-modal objectives for the task of extractive multi-modal summary generation - a) summarization-based objective, and b) clustering-based objectives. For summarization-based objectives, they use the following three objectives - i) Salience(txt) / Redundancy(txt), ii) Salience(img) / Redundancy(img), and iii) cross-modal correspondence; while for clustering-based objectives, they use PBM \cite{pakhira2004validity}, a popular cluster validity index (a function of cluster compactness and separation) to evaluate the uni-modal clusters of image and text, giving the following set of objectives - i) PBM(txt), PBM(img), and cross-modal correspondence. Almost all the neural networks based multi-modal summarization frameworks \cite{li2018multi, palaskar2019multimodal, chen2018abstractive, chen2018extractive} on the other hand use the standard negative log-likelihood function over the output vocabulary as the training objective. Some works also use textual and visual coverage loss to prevent over-attending the input as well \cite{li2018multi, zhu2018msmo}.
    
    \item \textbf{Multi-modal social media event summarization:} Various works have been conducted on the social media data that consists of opinions and experiences of a diverse set of population. \citet{tiwari2018multimodal} proposes the problem of summarizing asynchronous information from multiple social media platforms like Twitter, Instagram, and Flickr to generate a summary of event that is widely covered by users of these platforms extensively. \citet{bian2013multimedia} propose multi-modal summarization of trending topics in microblogs. They use Sina Weibo\footnote{\url{http://www.weibo.com/}} microblogs for the experimentation, which is a very popular microblogging platform in China. \citet{qian2018multimodal} uses the Weibo platform information to summarize disaster events like train crash and earthquakes. 
\end{itemize}

\section{Organization of existing work} \label{sec:seg}
\begin{figure}[!ht]
\includegraphics[width=0.9\textwidth]{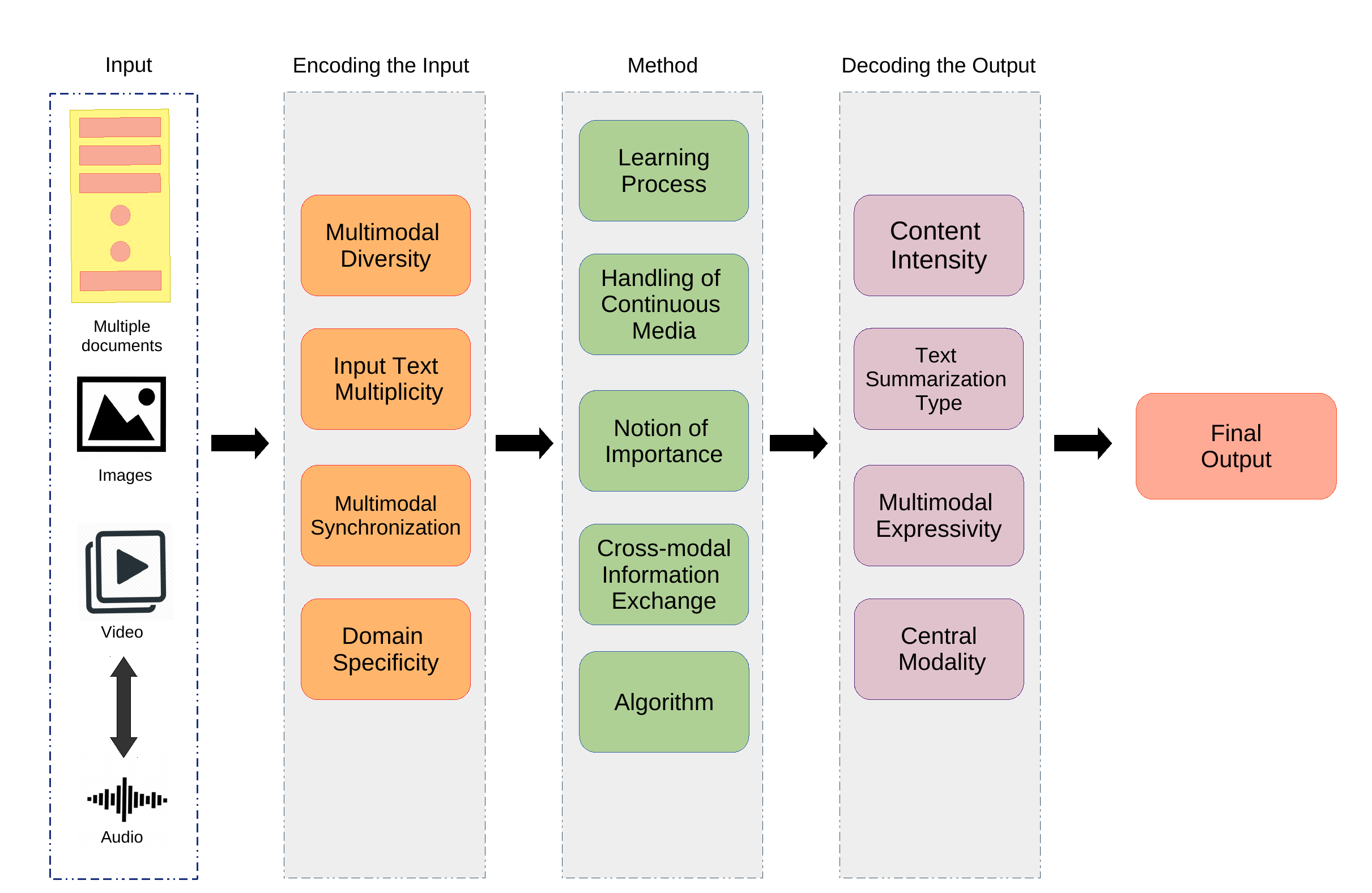}\centering
\caption{Generic multimodal summarization model flow diagram based on the defined taxonomy. Here the boxes at different stages of the flow diagram highlight the various types of factors that needs to be taken care of during that stage. "Orange" box for "Encoding the Input" stage, "Green" box for "Method" stage and "Purple" for "Decoding the output" stage. Based on these factors we have defined our taxonomy } \label{fig:model_generic}
\end{figure}

\begin{figure}[t]
\includegraphics[width=\textwidth]{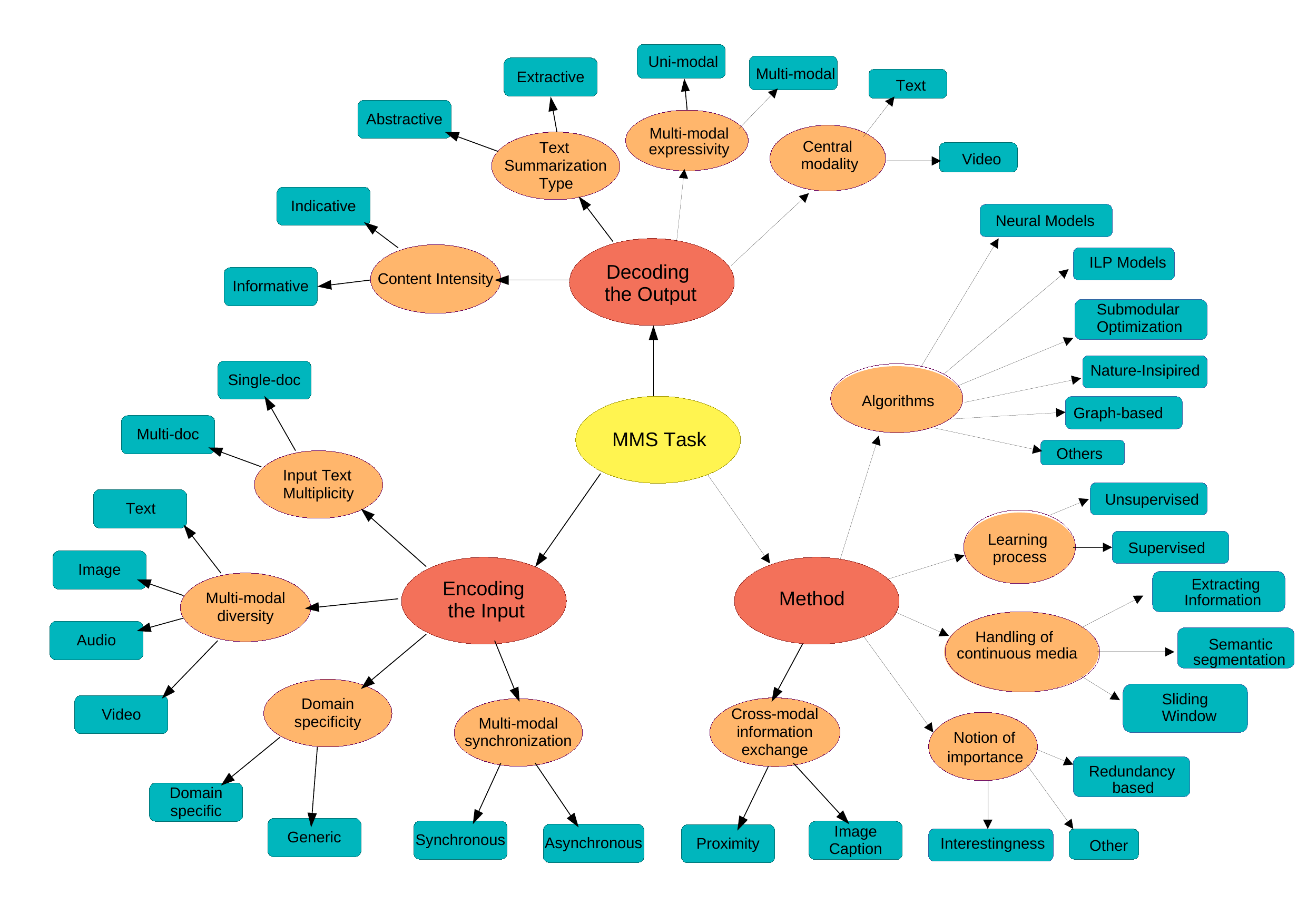}\centering
\caption{Visual representation of proposed taxonomy. The dark orange nodes coming out of the root (in yellow) represents the segregation based on input, output and adopted methodology, while the light orange nodes following them represent the respective characteristics of the research work on which the works can be distinguished. The teal colored rectangles in the leaf denote the various categories of each such characteristic.} \label{fig1}
\end{figure}

Different attempts have been made to solve the MMS task, and thus it is important to categorize the existing works to get a better understanding of the task. We categorize the prior works into three broad categories, depending upon encoding the input, the model architecture, and decoding the output. We have also illustrated these categorizations through a generic model diagram in Figure \ref{fig:model_generic}. A detailed pictorial representation of the taxonomy is shown in Figure \ref{fig1} and a comprehensive study is provided in Table \ref{tab:categ} (note that if some classifications are not marked in the table, then either the information about that category was not present, or is not applicable.).

\subsection{On the basis of encoding the input}
A multi-modal summarization task is highly driven by the kind of input it is given. Due to this dependency on diverse input modalities, the feature extraction and encoding strategies also differ for different multi-modal summarization systems. Existing works can be distinguished from others on the basis of the type of input and its encoding strategy in the following categories: 

\vspace{1.5mm}
\noindent\textbf{Multi-modal Diversity (MMD):} Different combinations of input (text, image,video \& audio) involve different preprocessing and encoding strategies. We can classify the existing works depending on the combination of modalities in which the input is represented. Various combinations within the input modalities like \textit{text-image} \cite{zhu2018msmo,li2018multi,chen2018abstractive}, \textit{text-video} \cite{fu2020multi, li2020vmsmo},  \textit{audio-video}\footnote{Note that \textit{audio-video} and \textit{text-audio-video} works are grouped together since in most of the existing works, automatic speech transcription is performed to obtain the textual modality part of data in the pre-processing step.} \cite{erol2003multimodal,evangelopoulos2013multimodal}, and \textit{text-image-audio-video} \cite{uzzaman2011multimodal,li2017multi,jangra2020text,jangra2020multimodal,jangra2021multimodal} have been explored in the literature of MMS. The different feature extraction strategies for individual modalities are described in Section \ref{subsec:preprocess}.

\vspace{1.5mm}
\noindent\textbf{Input Text Multiplicity (ITM):} Since a major focus of this survey is on MMS tasks with text as the \textit{central modality}, the number of text documents in input can also be one way of categorizing the related works. Depending upon whether the textual input is single-document \cite{chen2018extractive,li2018multi,zhu2018msmo} or multi-document \cite{li2017multi,jangra2020text,jangra2020multimodal,jangra2021multimodal}, the input preprocessing and the overall summarization strategies might differ. Having multiple documents makes the task a lot more challenging, since the degree of redundant information in input becomes a lot more prominent, making the data somewhat more noisy \cite{ma2020multidocument}.

\vspace{1.5mm}
\noindent\textbf{Multi-modal Synchronization (MMSy)\footnote{Note that the term synchronization is mostly used when there is a continuous media in consideration.}:} Synchronization refers to the interaction of two or more things at the same time or rate. For multi-modal summarization, having a synchronized input indicates that the multiple modalities have a coordination in terms of information flow, making them convey information in unison. We then classify input as synchronous \cite{erol2003multimodal,evangelopoulos2013multimodal} and asynchronous \cite{li2017multi,jangra2020text,jangra2020multimodal,tjondronegoro2011multi,jangra2021multimodal}.

\vspace{1.5mm}
\noindent\textbf{Domain Specificity (DS):} Domain can be defined as the specific area of cognition that is covered by any data, and depending upon the extent of domain coverage, we can classify works as \textit{domain-specific} or \textit{generic}.
The approach to summarize a \textit{domain-specific} input can differ from the \textit{generic} input greatly, since feature extraction in the former can be very particular in nature while not so in the latter, impacting the overall techniques immensely. Most of the news summarization tasks \cite{jangra2020multimodal,zhu2018msmo,li2017multi,chen2018abstractive,jangra2021multimodal} are \textit{generic} in nature, since news covers information about almost all the domains; whereas movie summarization \cite{evangelopoulos2013multimodal}, sports event summarization for tennis \cite{tjondronegoro2011multi} and soccer \cite{sanabria2019deep}, meetings recording summarization \cite{erol2003multimodal}, tutorial summarization \cite{libovicky2018multimodal}, social media event summarization \cite{tiwari2018multimodal} are examples of \textit{domain specific} tasks.

\subsubsection{Feature Extraction Strategies} \label{subsec:preprocess}
In a multi-modal setting, pre-processing \& feature extraction becomes vital step, since it involves extracting features from different modalities. Each input modality has been dealt with using modal-specific feature-extraction techniques. Even though some works tend to learn the semantic representation of data using their own proposed models, nearly all follow the same steps for feature extraction. Since the related works have different sets of input modalities, we describe feature extraction techniques for each modality individually.

\vspace{1.5mm}
\noindent\textbf{Text:} Traditionally, before the era of deep learning, Term Frequency-Document Inverse Frequency (TF-IDF) \cite{salton1989automatic} was used to identify relevant text segments \cite{erol2003multimodal,tjondronegoro2011multi,evangelopoulos2013multimodal}. Due to significant advancements in feature extraction, almost all the MMS tasks in the past five years either use pre-trained embeddings like word2vec \cite{mikolov2013distributed} or Glove \cite{pennington2014glove}. These pre-trained embeddings utilize the fact that the semantic information of a word is related to its contextual neighbors for training. Some works also train similar embeddings on their own datasets \cite{zhu2018msmo, zhu3multimodal} (refer to \textit{Feature Extraction} in Section \ref{subsec:neural}). Some works also adopt different pre-processing steps, depending upon the task specifications. For example, \citet{tiwari2018multimodal} applied a normalizer to handle the concept of expressive lengthening dealing with microblog datasets. Even though current MMS systems have not yet adopted them, it is worth mentioning Transformer-based word representations \cite{vaswani2017attention} like BERT, etc. that have achieved state-of-the-art performance in the vast majority of NLP and vision tasks. This achievement can be credited to their fast training due to parallelization, and ability to pre-train the language models on unlabelled corpora. We even have multi-lingual embeddings like LabSE \cite{feng2020languageagnostic}, and multi-modal text-image embeddings like UNITER \cite{chen2020uniter}, VilBERT \cite{lu2019vilbert}, VisualBERT \cite{li2019visualbert}, Pixel-BERT \cite{huang2020pixel}, etc.

\begin{table}[tp]  \centering
\scriptsize
\caption{\textbf{Comprehensive list of works that use specific pre-trained deep learning frameworks used to generate image embeddings.}}\label{tab:img}
\renewcommand{\arraystretch}{2}
\begin{tabular}{|p{0.2\textwidth}|p{0.6\textwidth}|}
\hline
\textbf{Pre-trained network} & \textbf{Works using this framework}\\
\hline
VGGNet \cite{Simonyan15} & \citet{li2017multi}, \citet{li2018multi}, \citet{zhu2018msmo}, \citet{chen2018abstractive}, \citet{zhu3multimodal}, \citet{jangra2020text}, \citet{jangra2020multimodal}, \citet{chen2018extractive}, \citet{modani2016summarizing}, \citet{jangra2021multimodal}
\\\hline

ResNet \cite{he2016deep} & \citet{fu2020multi}, \citet{li2020vmsmo}, \citet{li2020aspect} \\\hline

GoogleNet \cite{szegedy2015going} & \citet{sanabria2019deep} \\\hline

\end{tabular}
\end{table}

\vspace{1.5mm}
\noindent\textbf{Images:} Images, unlike text, are non-sequential and have a two-dimensional contextual span. Convolutional neural network (CNN) based deep neural network models have proven to be very promising in feature extraction tasks, but training these models requires large datasets, making it difficult to train features on MMS datasets. Hence, most of the existing works use pre-trained networks (e.g., ResNet \cite{he2016deep}, VGGNet \cite{Simonyan15}, GoogleNet \cite{szegedy2015going}) trained on large image classification datasets like ImageNet \cite{deng2009imagenet}. The technique of extracting local features (containing information about a confined patch of image) along with global features has shown promise in the MMS task as well \cite{zhu2018msmo}. A detailed list of frameworks that use pre-trained deep learning networks can be found in Table \ref{tab:img}. \citet{tiwari2018multimodal} uses Speeded-Up Robust Features (SURF) for each image, following a bag-of-word approach to creating a visual vocabulary. \citet{chen2018extractive} handle images by first extracting the Scale Invariant Feature Transform (SIFT) features. These SIFT features are fed to a hierarchical quantization module \cite{qian2014landmark} to obtain a 10,000-dimensional bag of the visual histogram. Having been inspired by the success of self-attention and Transformers \cite{vaswani2017attention} in effectively  modeling textual sequences, researchers in computer vision have adopted the techniques like self-attention, unsupervised pre-training, parallelizability of transformer architecture, etc. to better model the image representations\footnote{The readers are encouraged to read the extensive survey provided by \citet{khan2021transformers}.}. In order to adopt the self-attention layer dedicated to text sequences, \citet{parmar2018image} proposed a framework that restricts the self-attention to the local neighborhoods, thus significantly increasing the size of images that the model can process, despite maintaining larger receptive fields per layer than a CNN framework. \citet{dosovitskiy2020image} illustrated that usage of self-attention in conjunction with CNNs is not required, and a pure transformer applied to the sequence of image patches can also perform well on image classification tasks. \citet{touvron2021going} developed and optimized deep image transformer frameworks that do not saturate early with more depth.

To the best of our knowledge, none of the existing multi-modal summarization works use image transformers to encode the images. Since these large-scale models have a lot more capability to store more learned patterns from large-scale datasets due to the huge parameter space, they are bound to improve the overall summarization process by aiding in better image understanding.

\vspace{1.5mm}
\noindent\textbf{Audio and video:} Audio and video are usually present together as a single synchronized continuous media, and hence we discuss the pre-processing techniques used to extract features from them simultaneously. Continuous media has been processed in many diverse ways. Since audios and videos are susceptible to noise, it becomes of utmost importance to detect relevant segments before proceeding to the training phase\footnote{Note that some deep neural models like \citet{fu2020multi} or \citet{li2020vmsmo} prefer to encode individual frames using CNNs, and then use trainable RNNs to encode temporal information in videos. This CNN-RNN framework is not part of pre-processing, but instead, it belongs to the main model since these layers are also affected during training.}. While some works have adopted a naïve sliding window approach, making equal length cuts and further experimenting on these segments \cite{erol2003multimodal}, quite a few have done a modal conversion, changing the information media using automatic speech transcription to generate speech transcriptions and extracting key-frames from video using techniques like boundary shot-detection \cite{jangra2020multimodal,jangra2020text,li2017multi,tjondronegoro2011multi,jangra2021multimodal}. Some works have also taken into account the nature of the dataset, and performed semantic segmentation, getting better segment slices. For example, \citet{tjondronegoro2011multi} worked on a tennis dataset and used the information that the umpire requires the audience to remain quiet during the match point, performing segmentation consisting of a segment that begins with low audio activity followed by high audio energy levels as a result of the cheering and the commentary. If the audio and video are converted into another modality, then their pre-processing follows the same procedure as the new modalities, whereas, in the case of segmentation, various metrics like acoustic confidence, audio magnitude, sound localization for audio, motion detection, and Spatio-temporal features driven by intensity, color, and orientation for video have been explored to determine the salience and relevance of segments depending upon the task at hand \cite{erol2003multimodal,li2017multi,evangelopoulos2013multimodal}.

\vspace{1.5mm}
\noindent\textbf{Cross-modal correspondence:} Although the majority of works train their own shared embedding space for multiple modalities using the information from the target datasets \cite{li2018multi, zhu2018msmo, libovicky2018multimodal}, quite a few works \cite{jangra2020multimodal, jangra2020text, li2017multi, modani2016summarizing,jangra2021multimodal} also tend to use pre-trained neural network models \cite{wang2016learning, karpathy2014deep} trained on the image-caption datasets like Pascal1k \cite{rashtchian2010collecting}, Flickr8k \cite{hodosh2013framing}, Flick30k \cite{young2014image} etc. to leverage the information overlap amongst different modalities. This becomes a necessity for small datasets that are mostly used for extractive summarization. However even these pre-trained models cannot process raw data, and hence the text and image inputs are first pre-processed to desired embedding formats and then are fed to these models with pre-trained weights. For example,  \citet{wang2016learning} required a 6,000-dimensional sentence vector and 4,096-dimensional image vector generated by applying Principal Component Analysis (PCA) \cite{pearson1901liii} to the 18,000-dimensional output from the Hybrid Gaussian Laplacian mixture model (HGLMM) \cite{klein2014fisher} and extracting the weights from the final fully connected layer, $fc7$, from VGGNet \cite{Simonyan15}, respectively. In recent years, various Transformer-based \cite{vaswani2017attention} models have also been developed to correlate semantic information across textual and visual modalities. These BERT \cite{devlin2018bert} inspired models include ViLBERT \cite{lu2019vilbert}, VisualBERT \cite{li2019visualbert}, VideoBERT \cite{sun2019videobert}, VLP \cite{zhou2020unified}, Pixel-BERT \cite{huang2020pixel} etc. to name a few. There has also been some video-text representation learning like \cite{lin2021vx2text} and \cite{patrick2020support} that can be used to summarize multi-modal content with continuous modalities. However, none of the recent works on multi-modal summarization has utilized these transformer-based techniques in their system pipelines.

\vspace{1.5mm}
\noindent\textbf{Domain specific techniques:} Most of the systems proposed to solve the problem of multi-modal summarization are generic and can be adapted to other domains and problem statements as well. But, there do exist some works that benefit from the external knowledge of particular domains and problem settings to create better-performing systems. For instance, \citet{tjondronegoro2011multi} utilizes the fact that in tennis, the umpire always requires spectators to be silent before a serve, until the end of the point. The authors also pointed out that the end of the point is usually marked by a loud cheer from the supporters of the players in the audience. They used this fact to perform smooth segmentation of tennis clips using audio energy levels to indicate the start and end positions of a segment. Similar to this, \citet{sanabria2019deep} utilized atomic events in a game of soccer like a pass, goal, dribble, etc. to segment the video, which is later connected together to generate the summary. Other than sports, such domain-specific solutions have also been adopted in other domains. For example, \citet{erol2003multimodal}, when summarizing meeting recordings of a conference room, seeks out some visual activity like ``someone entering the room'' or ``someone standing up to write something on a whiteboard'' to detect some event likely to contain relevant information. In a different domain setting, people have benefited from other data pre-processing strategies, for instance, \citet{li2020aspect} extracts various key aspects of products like ``environmentally friendly refrigerators'' or ``energy efficient freezers'' to generate a captivating summary for Chinese e-commerce products.

\subsection{On the basis of method}
A lot of various approaches have been developed to solve the MMS task, and we can organize the existing works on the basis of proposed methodologies as follows:

\vspace{1.5mm}
\noindent\textbf{Learning process (LP):} A lot of work has been done in both supervised learning \cite{zhu3multimodal,libovicky2018multimodal,chen2018extractive,zhu2018msmo,li2018multi} and unsupervised learning \cite{jangra2020multimodal,jangra2020text,erol2003multimodal,evangelopoulos2013multimodal,li2017multi,jangra2021multimodal}. It can be observed that a large fraction of supervised techniques adopt deep neural networks to tackle the problem \cite{li2018multi,chen2018abstractive,zhu2018msmo,libovicky2018multimodal}, whereas in unsupervised techniques a large diversity of techniques have been adopted including deep neural networks \cite{chen2018extractive}, integer linear programming \cite{jangra2020text}, differential evolution \cite{jangra2020multimodal,jangra2021multimodal}, submodular optimization \cite{li2017multi} etc.

\vspace{1.5mm}
\noindent\textbf{Handling of continuous media (HCM):} We can also distinguish between works depending upon how the proposed models handle continuous media (audio and video in this case). There are three broad distinctions possible, a) \textit{extracting-information}, where the model extracts information from continuous media to get a discrete representation \cite{jangra2020multimodal,jangra2020text,li2017multi,jangra2021multimodal}, b) \textit{semantic-segmentation}, where a logical technique is proposed to slice out the continuous media \cite{tjondronegoro2011multi,evangelopoulos2013multimodal,evangelopoulos2009video}, and c) \textit{sliding window}, when a naïve fixed window based modeling is performed \cite{erol2003multimodal}.

\vspace{1.5mm}
\noindent\textbf{Notion of importance (NI):} One of the most significant distinctions would be the notion of importance used to generate the final summary. A diverse set of objectives ranging from interestingness \cite{tjondronegoro2011multi}, redundancy \cite{li2017multi}, cluster validity index \cite{jangra2020multimodal}, acoustic energy / visual illumination \cite{evangelopoulos2009video,evangelopoulos2013multimodal}, and social popularity \cite{sahuguet2013socially} have been explored in attempt to solve the MMS task.

\vspace{1.5mm}
\noindent\textbf{Cross-modal information exchange (CIE):} The most important part of a MMS model is the ability to extract and share information across multiple modalities. Most of the works either adopt a proximity-based approach \cite{xu2013cross, erol2003multimodal}, a pre-trained model on image-caption pairs based corpora for information overlap \cite{jangra2020multimodal,jangra2020text,li2017multi,jangra2021multimodal}, or learn the semantic overlap over uni-modal embeddings \cite{li2018multi,zhu2018msmo,zhu3multimodal,chen2018extractive,chen2018abstractive}.

\vspace{1.5mm}
\noindent\textbf{Algorithms (A):} The algorithm for the multimodal summarization task varies from traditional multiobjective optimization strategies to modern deep learning-based approaches. We can classify the existing works based on the different algorithms as Neural models (NN), Integer Linear Programming based models (ILP), Submodular Optimization-based models (SO), Nature-Inspired Algorithm based models (NIA), Graph-based models (G), and other algorithms (Oth). We have discussed these different methods in detail in Section \ref{sub:main-model}. The other algorithms comprise different clustering-based, LDA \cite{blei2003latent} - based and audio-video analysis-based techniques which were earlier used for performing multimodal summarization. 

\subsection{On the basis of decoding the output}
The summarization objective decides the desired type of output. For different summarization objectives, the type of output and decoding method vary. Depending on the type of output and the decoding method, we can categorize the existing works on the following basis: 

\vspace{1.5mm}
\noindent\textbf{Content intensity (CI):} The degree to which an output summary elaborates on a concept can hugely impact the overall modeling. The output summary can either be \textit{informative}, having detailed information about the input topic \cite{libovicky2018multimodal,yan2012visualizing}, or \textit{indicative}, only hinting at the most relevant information \cite{zhu2018msmo,chen2018abstractive}.

\vspace{1.5mm}
\noindent\textbf{Text Summarization Type (TST):} The most widely discussed distinction for text summarization works is the distinction of \textit{extractive} vs \textit{abstractive}. Abstractive summarization systems generally use a beam search or greedy search mechanism for decoding the output summary. While extractive systems during decoding use some scoring mechanism to identify the salient, non-redundant, and readable elements from the input for the final output. Depending on the nature of an output text summary, we can also classify the works in MMS tasks (containing text in the output) into \textit{extractive MMS} \cite{jangra2020multimodal,jangra2020text,chen2018extractive,li2017multi,jangra2021multimodal} and \textit{abstractive MMS} \cite{zhu3multimodal,chen2018abstractive,zhu2018msmo,li2018multi}\footnote{Note that other modalities beside text have been so far subject to only extractive approaches in MMS researches.}.

\vspace{1.5mm}
\noindent\textbf{Multi-modal expressivity (MME):} Whether the output is \textit{uni-modal} (comprising of one modality) \cite{libovicky2018multimodal,li2018multi,chen2018extractive,li2017multi,evangelopoulos2013multimodal} or \textit{multi-modal} (comprising of multiple modalities) \cite{jangra2020multimodal,jangra2020text,zhu3multimodal,chen2018abstractive,zhu2018msmo,tjondronegoro2011multi,jangra2021multimodal} is a major classification for the existing work. Mostly the systems producing multimodal output involve some post-processing steps for selecting the final output elements from the non-central modalities. 

\vspace{1.5mm}
\noindent\textbf{Central modality (CM):} Based on \textit{central-modality} (defined in Section \ref{sec:mms}), existing works can also be distinguished depending on the base modality around which the final output, as well as the model, are formulated. A large portion of the prior work adopts either a text-centric approach \cite{jangra2020text,libovicky2018multimodal,chen2018abstractive,zhu2018msmo,li2018multi,li2017multi,jangra2021multimodal} or a video-centric\footnote{Here audio is assumed to be a part of video since in all the existing works video and audio are synchronous to each other.} approach \cite{sahuguet2013socially,evangelopoulos2013multimodal,tjondronegoro2011multi,erol2003multimodal}. Few of the decoding methods followed popularly in neural models have been discussed in detail in Section \ref{subsec:neural}.

\begin{table}
 \setlength\extrarowheight{2pt}
 \centering
 \tiny
 \caption{\textbf{Comprehensive study of existing work using the proposed taxonomy (refer to Section \ref{sec:seg}).}}
 \label{tab:categ}
 \vspace{-3mm}
 \rotatebox{90}{
 \begin{tabular}{|  l*{30}{|c}|}
 \hline
   & \multicolumn{10}{c|}{\textbf{Input Based}} & \multicolumn{8}{c|}{\textbf{Output Based}} & \multicolumn{12}{c|}{\textbf{Method Based}} \\ \cline{2-31}
  
  \multirow{2}{*}{\textbf{\hspace{10mm} Papers}} & \multicolumn{4}{c|}{\textbf{MMD}} & \multicolumn{2}{c|}{\textbf{ITM}} & \multicolumn{2}{c|}{\textbf{MMSy}} & \multicolumn{2}{c|}{\textbf{DS}} & \multicolumn{2}{c|}{\textbf{CI}} & \multicolumn{2}{c|}{\textbf{TST}} & \multicolumn{2}{c|}{\textbf{MME}} & \multicolumn{2}{c|}{\textbf{CM}} & \multicolumn{2}{c|}{\textbf{LP}} & \multicolumn{3}{c|}{\textbf{HCM}} & \multicolumn{3}{c|}{\textbf{NI}} & \multicolumn{3}{c|}{\textbf{CIE}} & \multicolumn{1}{c|}
  {\textbf{A}} \\ \cline{2-31}
  
   & \rotatebox{270}{Text\phantom{.}} & \rotatebox{270}{Images\phantom{.}} & \rotatebox{270}{Audio\phantom{.}} &  \rotatebox{270}{Video\phantom{.}} & \rotatebox{270}{Single-doc\phantom{.}} & \rotatebox{270}{Multi-doc\phantom{.}} & \rotatebox{270}{Sync\phantom{.}} &   \rotatebox{270}{Async\phantom{.}} & \rotatebox{270}{Domain Specific\phantom{.}} & \rotatebox{270}{Generic\phantom{.}} & \rotatebox{270}{Informative\phantom{.}} & \rotatebox{270}{Indicative\phantom{.}} & \rotatebox{270}{Abstractive\phantom{.}} & \rotatebox{270}{Extractive\phantom{.}} & \rotatebox{270}{Uni-modal\phantom{.}} & \rotatebox{270}{Multi-modal\phantom{.}} & \rotatebox{270}{Text\phantom{.}} & \rotatebox{270}{Video\phantom{.}} & \rotatebox{270}{Unsupervised\phantom{.}} & \rotatebox{270}{Supervised\phantom{.}} & \rotatebox{270}{Extracting info\phantom{.}} & \rotatebox{270}{Semantic seg\phantom{.}} & \rotatebox{270}{Sliding window\phantom{.}} & \rotatebox{270}{Redundancy based\phantom{.}} & \rotatebox{270}{Interestingness\phantom{.}} & \rotatebox{270}{Other\phantom{.}} & \rotatebox{270}{Image Caption\phantom{.}} & \rotatebox{270}{Proximity\phantom{.}} & \rotatebox{270}{Uni-modal embedding\phantom{.}} & \rotatebox{270}{Algorithms\phantom{.}}\\ \hline

  \citet{erol2003multimodal} & \textcolor{applegreen}{\textbf{\checkmark}} & & \textcolor{applegreen}{\textbf{\checkmark}} & \textcolor{applegreen}{\textbf{\checkmark}} & \textcolor{applegreen}{\textbf{\checkmark}} & & \textcolor{applegreen}{\textbf{\checkmark}} & & \textcolor{applegreen}{\textbf{\checkmark}} & & & \textcolor{applegreen}{\textbf{\checkmark}} & & & & \textcolor{applegreen}{\textbf{\checkmark}} & & \textcolor{applegreen}{\textbf{\checkmark}} & \textcolor{applegreen}{\textbf{\checkmark}} & & \textcolor{applegreen}{\textbf{\checkmark}} & & & & & \textcolor{applegreen}{\textbf{\checkmark}} & & \textcolor{applegreen}{\textbf{\checkmark}} & & 
  Oth \\ \hline
  \citet{tjondronegoro2011multi} & \textcolor{applegreen}{\textbf{\checkmark}} & & \textcolor{applegreen}{\textbf{\checkmark}} & \textcolor{applegreen}{\textbf{\checkmark}} & \textcolor{applegreen}{\textbf{\checkmark}} & & \textcolor{applegreen}{\textbf{\checkmark}} & & \textcolor{applegreen}{\textbf{\checkmark}} & & \textcolor{applegreen}{\textbf{\checkmark}} & & & & & \textcolor{applegreen}{\textbf{\checkmark}} & & \textcolor{applegreen}{\textbf{\checkmark}} & \textcolor{applegreen}{\textbf{\checkmark}} & & & \textcolor{applegreen}{\textbf{\checkmark}} & & & \textcolor{applegreen}{\textbf{\checkmark}} & & & \textcolor{applegreen}{\textbf{\checkmark}} & & 
  Oth\\ \hline
  \citet{uzzaman2011multimodal}  & \textcolor{applegreen}{\textbf{\checkmark}} & \textcolor{applegreen}{\textbf{\checkmark}} & & & \textcolor{applegreen}{\textbf{\checkmark}} & & & \textcolor{applegreen}{\textbf{\checkmark}} & & \textcolor{applegreen}{\textbf{\checkmark}} & \textcolor{applegreen}{\textbf{\checkmark}} & & & & & \textcolor{applegreen}{\textbf{\checkmark}} & \textcolor{applegreen}{\textbf{\checkmark}} & & \textcolor{applegreen}{\textbf{\checkmark}} & & & & & & & \textcolor{applegreen}{\textbf{\checkmark}} & & & & 
  Oth\\ \hline
  \citet{evangelopoulos2013multimodal} & \textcolor{applegreen}{\textbf{\checkmark}} & & \textcolor{applegreen}{\textbf{\checkmark}} & \textcolor{applegreen}{\textbf{\checkmark}} & \textcolor{applegreen}{\textbf{\checkmark}} & & \textcolor{applegreen}{\textbf{\checkmark}} & & \textcolor{applegreen}{\textbf{\checkmark}} & & & \textcolor{applegreen}{\textbf{\checkmark}} & & & & \textcolor{applegreen}{\textbf{\checkmark}} & & \textcolor{applegreen}{\textbf{\checkmark}} & \textcolor{applegreen}{\textbf{\checkmark}} & & & \textcolor{applegreen}{\textbf{\checkmark}} & & & & \textcolor{applegreen}{\textbf{\checkmark}} & & \textcolor{applegreen}{\textbf{\checkmark}} & & 
  Oth\\ \hline
  \citet{li2017multi} & \textcolor{applegreen}{\textbf{\checkmark}} & \textcolor{applegreen}{\textbf{\checkmark}} & \textcolor{applegreen}{\textbf{\checkmark}} & \textcolor{applegreen}{\textbf{\checkmark}} & & \textcolor{applegreen}{\textbf{\checkmark}} & & \textcolor{applegreen}{\textbf{\checkmark}} & & \textcolor{applegreen}{\textbf{\checkmark}} & & \textcolor{applegreen}{\textbf{\checkmark}} & & \textcolor{applegreen}{\textbf{\checkmark}} & \textcolor{applegreen}{\textbf{\checkmark}} & & \textcolor{applegreen}{\textbf{\checkmark}} & & \textcolor{applegreen}{\textbf{\checkmark}} & & \textcolor{applegreen}{\textbf{\checkmark}} & & & \textcolor{applegreen}{\textbf{\checkmark}} & & & \textcolor{applegreen}{\textbf{\checkmark}} & & & 
  SO, G \\ \hline
  \citet{li2018multi} & \textcolor{applegreen}{\textbf{\checkmark}} & \textcolor{applegreen}{\textbf{\checkmark}} & & & \textcolor{applegreen}{\textbf{\checkmark}} & & & \textcolor{applegreen}{\textbf{\checkmark}} & & \textcolor{applegreen}{\textbf{\checkmark}} & & \textcolor{applegreen}{\textbf{\checkmark}} & \textcolor{applegreen}{\textbf{\checkmark}} & & \textcolor{applegreen}{\textbf{\checkmark}} & & \textcolor{applegreen}{\textbf{\checkmark}} & & & \textcolor{applegreen}{\textbf{\checkmark}} & & & & \textcolor{applegreen}{\textbf{\checkmark}} & & & & & \textcolor{applegreen}{\textbf{\checkmark}} & 
  NN \\ \hline
  \citet{zhu2018msmo} & \textcolor{applegreen}{\textbf{\checkmark}} & \textcolor{applegreen}{\textbf{\checkmark}} & & & \textcolor{applegreen}{\textbf{\checkmark}} & & & \textcolor{applegreen}{\textbf{\checkmark}} & & \textcolor{applegreen}{\textbf{\checkmark}} & & \textcolor{applegreen}{\textbf{\checkmark}} & \textcolor{applegreen}{\textbf{\checkmark}} & & & \textcolor{applegreen}{\textbf{\checkmark}} & \textcolor{applegreen}{\textbf{\checkmark}} & & & \textcolor{applegreen}{\textbf{\checkmark}} & & & & \textcolor{applegreen}{\textbf{\checkmark}} & & & & & \textcolor{applegreen}{\textbf{\checkmark}} & 
  NN \\ \hline
  \citet{chen2018abstractive} & \textcolor{applegreen}{\textbf{\checkmark}} & \textcolor{applegreen}{\textbf{\checkmark}} & & & \textcolor{applegreen}{\textbf{\checkmark}} & & & \textcolor{applegreen}{\textbf{\checkmark}} & & \textcolor{applegreen}{\textbf{\checkmark}} & & \textcolor{applegreen}{\textbf{\checkmark}} & \textcolor{applegreen}{\textbf{\checkmark}} & & & \textcolor{applegreen}{\textbf{\checkmark}} & \textcolor{applegreen}{\textbf{\checkmark}} & & & \textcolor{applegreen}{\textbf{\checkmark}} & & & & \textcolor{applegreen}{\textbf{\checkmark}} & & & & & \textcolor{applegreen}{\textbf{\checkmark}} & 
  NN \\ \hline
  \citet{libovicky2018multimodal} & \textcolor{applegreen}{\textbf{\checkmark}} & & \textcolor{applegreen}{\textbf{\checkmark}} & \textcolor{applegreen}{\textbf{\checkmark}} & \textcolor{applegreen}{\textbf{\checkmark}} & & \textcolor{applegreen}{\textbf{\checkmark}} & & \textcolor{applegreen}{\textbf{\checkmark}} & & & \textcolor{applegreen}{\textbf{\checkmark}} & \textcolor{applegreen}{\textbf{\checkmark}} & & \textcolor{applegreen}{\textbf{\checkmark}} & & & \textcolor{applegreen}{\textbf{\checkmark}} & & \textcolor{applegreen}{\textbf{\checkmark}} & & & \textcolor{applegreen}{\textbf{\checkmark}} & \textcolor{applegreen}{\textbf{\checkmark}} & & & & & \textcolor{applegreen}{\textbf{\checkmark}} & NN \\ \hline
  \citet{palaskar2019multimodal} & \textcolor{applegreen}{\textbf{\checkmark}} & & \textcolor{applegreen}{\textbf{\checkmark}} & \textcolor{applegreen}{\textbf{\checkmark}} & \textcolor{applegreen}{\textbf{\checkmark}} & & \textcolor{applegreen}{\textbf{\checkmark}} & & \textcolor{applegreen}{\textbf{\checkmark}} & & & \textcolor{applegreen}{\textbf{\checkmark}} & \textcolor{applegreen}{\textbf{\checkmark}} & & \textcolor{applegreen}{\textbf{\checkmark}} & & & \textcolor{applegreen}{\textbf{\checkmark}} & & \textcolor{applegreen}{\textbf{\checkmark}} & & & \textcolor{applegreen}{\textbf{\checkmark}} & \textcolor{applegreen}{\textbf{\checkmark}} & & & & & \textcolor{applegreen}{\textbf{\checkmark}} & NN \\ \hline
  \citet{zhu3multimodal} & \textcolor{applegreen}{\textbf{\checkmark}} & \textcolor{applegreen}{\textbf{\checkmark}} & & & \textcolor{applegreen}{\textbf{\checkmark}} & & & \textcolor{applegreen}{\textbf{\checkmark}} & & \textcolor{applegreen}{\textbf{\checkmark}} & & \textcolor{applegreen}{\textbf{\checkmark}} & \textcolor{applegreen}{\textbf{\checkmark}} & & & \textcolor{applegreen}{\textbf{\checkmark}} & \textcolor{applegreen}{\textbf{\checkmark}} & & & \textcolor{applegreen}{\textbf{\checkmark}} & & & & \textcolor{applegreen}{\textbf{\checkmark}} & & & & & \textcolor{applegreen}{\textbf{\checkmark}} & NN\\ \hline
  \citet{jangra2020text} & \textcolor{applegreen}{\textbf{\checkmark}} & \textcolor{applegreen}{\textbf{\checkmark}} & \textcolor{applegreen}{\textbf{\checkmark}} & \textcolor{applegreen}{\textbf{\checkmark}} & & \textcolor{applegreen}{\textbf{\checkmark}} & & \textcolor{applegreen}{\textbf{\checkmark}} & & \textcolor{applegreen}{\textbf{\checkmark}} & & \textcolor{applegreen}{\textbf{\checkmark}} & & \textcolor{applegreen}{\textbf{\checkmark}} & & \textcolor{applegreen}{\textbf{\checkmark}}  & \textcolor{applegreen}{\textbf{\checkmark}} & & \textcolor{applegreen}{\textbf{\checkmark}} & & \textcolor{applegreen}{\textbf{\checkmark}} & & & \textcolor{applegreen}{\textbf{\checkmark}} & & & \textcolor{applegreen}{\textbf{\checkmark}} & & & ILP \\ \hline
  \citet{jangra2020multimodal} & \textcolor{applegreen}{\textbf{\checkmark}} & \textcolor{applegreen}{\textbf{\checkmark}} & \textcolor{applegreen}{\textbf{\checkmark}} & \textcolor{applegreen}{\textbf{\checkmark}} & & \textcolor{applegreen}{\textbf{\checkmark}} & & \textcolor{applegreen}{\textbf{\checkmark}} & & \textcolor{applegreen}{\textbf{\checkmark}} & & \textcolor{applegreen}{\textbf{\checkmark}} & & \textcolor{applegreen}{\textbf{\checkmark}} & & \textcolor{applegreen}{\textbf{\checkmark}}  & \textcolor{applegreen}{\textbf{\checkmark}} & & \textcolor{applegreen}{\textbf{\checkmark}} & & \textcolor{applegreen}{\textbf{\checkmark}} & & & \textcolor{applegreen}{\textbf{\checkmark}} & & & \textcolor{applegreen}{\textbf{\checkmark}} & &  & NIA\\ \hline
  \citet{jangra2021multimodal} & \textcolor{applegreen}{\textbf{\checkmark}} & \textcolor{applegreen}{\textbf{\checkmark}} & \textcolor{applegreen}{\textbf{\checkmark}} & \textcolor{applegreen}{\textbf{\checkmark}} & & \textcolor{applegreen}{\textbf{\checkmark}} & & \textcolor{applegreen}{\textbf{\checkmark}} & & \textcolor{applegreen}{\textbf{\checkmark}} & & \textcolor{applegreen}{\textbf{\checkmark}} & & \textcolor{applegreen}{\textbf{\checkmark}} & & \textcolor{applegreen}{\textbf{\checkmark}}  & \textcolor{applegreen}{\textbf{\checkmark}} & & \textcolor{applegreen}{\textbf{\checkmark}} & & \textcolor{applegreen}{\textbf{\checkmark}} & & & \textcolor{applegreen}{\textbf{\checkmark}} & & & \textcolor{applegreen}{\textbf{\checkmark}} & & & NIA \\ \hline
  \citet{xu2013cross} & \textcolor{applegreen}{\textbf{\checkmark}} & \textcolor{applegreen}{\textbf{\checkmark}} & & & & \textcolor{applegreen}{\textbf{\checkmark}} & \textcolor{applegreen}{\textbf{\checkmark}} & & & \textcolor{applegreen}{\textbf{\checkmark}} & \textcolor{applegreen}{\textbf{\checkmark}} & & & \textcolor{applegreen}{\textbf{\checkmark}} & & \textcolor{applegreen}{\textbf{\checkmark}} & \textcolor{applegreen}{\textbf{\checkmark}} & & \textcolor{applegreen}{\textbf{\checkmark}} & & & & & & & \textcolor{applegreen}{\textbf{\checkmark}} & & \textcolor{applegreen}{\textbf{\checkmark}} & & NIA\\ \hline
  \citet{sahuguet2013socially} & \textcolor{applegreen}{\textbf{\checkmark}} & \textcolor{applegreen}{\textbf{\checkmark}} & \textcolor{applegreen}{\textbf{\checkmark}} & \textcolor{applegreen}{\textbf{\checkmark}} & & \textcolor{applegreen}{\textbf{\checkmark}} & \textcolor{applegreen}{\textbf{\checkmark}} & & \textcolor{applegreen}{\textbf{\checkmark}} & & & \textcolor{applegreen}{\textbf{\checkmark}} & & \textcolor{applegreen}{\textbf{\checkmark}} & & \textcolor{applegreen}{\textbf{\checkmark}} & & \textcolor{applegreen}{\textbf{\checkmark}} & \textcolor{applegreen}{\textbf{\checkmark}} & & & & & & & \textcolor{applegreen}{\textbf{\checkmark}} & & \textcolor{applegreen}{\textbf{\checkmark}} & & Oth\\ \hline
  \citet{tiwari2018multimodal} & \textcolor{applegreen}{\textbf{\checkmark}} & \textcolor{applegreen}{\textbf{\checkmark}} & & & & \textcolor{applegreen}{\textbf{\checkmark}} & \textcolor{applegreen}{\textbf{\checkmark}} & & & \textcolor{applegreen}{\textbf{\checkmark}} & & \textcolor{applegreen}{\textbf{\checkmark}} & & \textcolor{applegreen}{\textbf{\checkmark}} & & \textcolor{applegreen}{\textbf{\checkmark}} & \textcolor{applegreen}{\textbf{\checkmark}} & & \textcolor{applegreen}{\textbf{\checkmark}} & & & & & \textcolor{applegreen}{\textbf{\checkmark}} & & & & \textcolor{applegreen}{\textbf{\checkmark}} & & SO\\ \hline
  \citet{bian2013multimedia} & \textcolor{applegreen}{\textbf{\checkmark}} & \textcolor{applegreen}{\textbf{\checkmark}} & & & & \textcolor{applegreen}{\textbf{\checkmark}} & & & \textcolor{applegreen}{\textbf{\checkmark}} & & \textcolor{applegreen}{\textbf{\checkmark}} & & & \textcolor{applegreen}{\textbf{\checkmark}} & & \textcolor{applegreen}{\textbf{\checkmark}} & \textcolor{applegreen}{\textbf{\checkmark}} & & & \textcolor{applegreen}{\textbf{\checkmark}} & & & & \textcolor{applegreen}{\textbf{\checkmark}} & & & & & & Oth\\ \hline
  \citet{yan2012visualizing} & \textcolor{applegreen}{\textbf{\checkmark}} & \textcolor{applegreen}{\textbf{\checkmark}} & & & & \textcolor{applegreen}{\textbf{\checkmark}} & & & \textcolor{applegreen}{\textbf{\checkmark}} & & \textcolor{applegreen}{\textbf{\checkmark}} & & & \textcolor{applegreen}{\textbf{\checkmark}} & & \textcolor{applegreen}{\textbf{\checkmark}} & \textcolor{applegreen}{\textbf{\checkmark}} & & & \textcolor{applegreen}{\textbf{\checkmark}} & & & & & & \textcolor{applegreen}{\textbf{\checkmark}} & & & & G\\ \hline
  \citet{qian2019social} & \textcolor{applegreen}{\textbf{\checkmark}} & \textcolor{applegreen}{\textbf{\checkmark}} & & & & \textcolor{applegreen}{\textbf{\checkmark}} & & \textcolor{applegreen}{\textbf{\checkmark}} & & & \textcolor{applegreen}{\textbf{\checkmark}} & \textcolor{applegreen}{\textbf{\checkmark}} & & \textcolor{applegreen}{\textbf{\checkmark}} & & \textcolor{applegreen}{\textbf{\checkmark}} & \textcolor{applegreen}{\textbf{\checkmark}} & \textcolor{applegreen}{\textbf{\checkmark}} & & & & & \textcolor{applegreen}{\textbf{\checkmark}} & & & & & & & Oth\\ \hline
  \citet{chen2018extractive} & \textcolor{applegreen}{\textbf{\checkmark}} & \textcolor{applegreen}{\textbf{\checkmark}} & & & \textcolor{applegreen}{\textbf{\checkmark}} & & & \textcolor{applegreen}{\textbf{\checkmark}} & & \textcolor{applegreen}{\textbf{\checkmark}} & & \textcolor{applegreen}{\textbf{\checkmark}} & & \textcolor{applegreen}{\textbf{\checkmark}} & & \textcolor{applegreen}{\textbf{\checkmark}} & \textcolor{applegreen}{\textbf{\checkmark}} & & & \textcolor{applegreen}{\textbf{\checkmark}} & & & & \textcolor{applegreen}{\textbf{\checkmark}} & & & & & \textcolor{applegreen}{\textbf{\checkmark}} & NN\\ \hline
  \citet{evangelopoulos2009video} & \textcolor{applegreen}{\textbf{\checkmark}} & & \textcolor{applegreen}{\textbf{\checkmark}} & \textcolor{applegreen}{\textbf{\checkmark}} & \textcolor{applegreen}{\textbf{\checkmark}} & & \textcolor{applegreen}{\textbf{\checkmark}} & & \textcolor{applegreen}{\textbf{\checkmark}} & & & \textcolor{applegreen}{\textbf{\checkmark}} & & & & \textcolor{applegreen}{\textbf{\checkmark}} & & \textcolor{applegreen}{\textbf{\checkmark}} & \textcolor{applegreen}{\textbf{\checkmark}} & & & \textcolor{applegreen}{\textbf{\checkmark}} & & & & \textcolor{applegreen}{\textbf{\checkmark}} & & \textcolor{applegreen}{\textbf{\checkmark}} & & Oth\\ \hline
  \citet{bian2014multimedia} & \textcolor{applegreen}{\textbf{\checkmark}} & \textcolor{applegreen}{\textbf{\checkmark}} & & & & \textcolor{applegreen}{\textbf{\checkmark}} & & & \textcolor{applegreen}{\textbf{\checkmark}} & & \textcolor{applegreen}{\textbf{\checkmark}} & & & \textcolor{applegreen}{\textbf{\checkmark}} & & \textcolor{applegreen}{\textbf{\checkmark}} & \textcolor{applegreen}{\textbf{\checkmark}} & & & \textcolor{applegreen}{\textbf{\checkmark}} & & & & \textcolor{applegreen}{\textbf{\checkmark}} & & & & & & Oth\\ \hline
  \citet{fu2020multi} & \textcolor{applegreen}{\textbf{\checkmark}} & & \textcolor{applegreen}{\textbf{\checkmark}} & \textcolor{applegreen}{\textbf{\checkmark}} & \textcolor{applegreen}{\textbf{\checkmark}} & & & \textcolor{applegreen}{\textbf{\checkmark}} & & \textcolor{applegreen}{\textbf{\checkmark}} & & \textcolor{applegreen}{\textbf{\checkmark}} & \textcolor{applegreen}{\textbf{\checkmark}} & & \textcolor{applegreen}{\textbf{\checkmark}} & & \textcolor{applegreen}{\textbf{\checkmark}} & & & \textcolor{applegreen}{\textbf{\checkmark}} & \textcolor{applegreen}{\textbf{\checkmark}} & & & \textcolor{applegreen}{\textbf{\checkmark}} & & & & & \textcolor{applegreen}{\textbf{\checkmark}} & NN\\ \hline
  \citet{li2020vmsmo} & \textcolor{applegreen}{\textbf{\checkmark}} & & & \textcolor{applegreen}{\textbf{\checkmark}} & \textcolor{applegreen}{\textbf{\checkmark}} & & & \textcolor{applegreen}{\textbf{\checkmark}} & & \textcolor{applegreen}{\textbf{\checkmark}} & & \textcolor{applegreen}{\textbf{\checkmark}} & \textcolor{applegreen}{\textbf{\checkmark}} & & & \textcolor{applegreen}{\textbf{\checkmark}} & \textcolor{applegreen}{\textbf{\checkmark}} & & & \textcolor{applegreen}{\textbf{\checkmark}} & \textcolor{applegreen}{\textbf{\checkmark}} & & \textcolor{applegreen}{\textbf{\checkmark}} & \textcolor{applegreen}{\textbf{\checkmark}} & & & & & \textcolor{applegreen}{\textbf{\checkmark}} & NN\\ \hline
  \citet{li2020aspect} & \textcolor{applegreen}{\textbf{\checkmark}} & \textcolor{applegreen}{\textbf{\checkmark}} & & & \textcolor{applegreen}{\textbf{\checkmark}} & & & \textcolor{applegreen}{\textbf{\checkmark}} & \textcolor{applegreen}{\textbf{\checkmark}} & & & \textcolor{applegreen}{\textbf{\checkmark}} & \textcolor{applegreen}{\textbf{\checkmark}} & & \textcolor{applegreen}{\textbf{\checkmark}} & & \textcolor{applegreen}{\textbf{\checkmark}} & & & \textcolor{applegreen}{\textbf{\checkmark}} & & & & \textcolor{applegreen}{\textbf{\checkmark}} & & & & & \textcolor{applegreen}{\textbf{\checkmark}} & NN\\ \hline
  \citet{modani2016summarizing} & \textcolor{applegreen}{\textbf{\checkmark}} & \textcolor{applegreen}{\textbf{\checkmark}} & & & \textcolor{applegreen}{\textbf{\checkmark}} & & & \textcolor{applegreen}{\textbf{\checkmark}} & & \textcolor{applegreen}{\textbf{\checkmark}} & & \textcolor{applegreen}{\textbf{\checkmark}} & & \textcolor{applegreen}{\textbf{\checkmark}} & & \textcolor{applegreen}{\textbf{\checkmark}} & \textcolor{applegreen}{\textbf{\checkmark}} & & \textcolor{applegreen}{\textbf{\checkmark}} & & & & & \textcolor{applegreen}{\textbf{\checkmark}} & & & \textcolor{applegreen}{\textbf{\checkmark}} & & & SO, G\\ \hline
  \citet{sanabria2019deep} & \textcolor{applegreen}{\textbf{\checkmark}} & & \textcolor{applegreen}{\textbf{\checkmark}} & \textcolor{applegreen}{\textbf{\checkmark}} & \textcolor{applegreen}{\textbf{\checkmark}} & & \textcolor{applegreen}{\textbf{\checkmark}} & & \textcolor{applegreen}{\textbf{\checkmark}} & & & \textcolor{applegreen}{\textbf{\checkmark}} & \textcolor{applegreen}{\textbf{\checkmark}} & & \textcolor{applegreen}{\textbf{\checkmark}} & & & \textcolor{applegreen}{\textbf{\checkmark}} & & \textcolor{applegreen}{\textbf{\checkmark}} & \textcolor{applegreen}{\textbf{\checkmark}} & \textcolor{applegreen}{\textbf{\checkmark}} & & & \textcolor{applegreen}{\textbf{\checkmark}} & \textcolor{applegreen}{\textbf{\checkmark}} & & \textcolor{applegreen}{\textbf{\checkmark}} & & NN\\ \hline
 \end{tabular}
}
\end{table}

\section{Overview of Methods} \label{sec:method}
A lot of works have attempted to solve the MMS task using supervised and unsupervised techniques. In this section, we attempt to describe the MMS frameworks in a generalized manner, elucidating the nuances of different approaches. Since the variety of inputs, outputs and techniques that were used span a large spectrum of possibilities, we describe each one individually. We have broken down this section into three stages, \textit{pre-processing}, \textit{main model}, and \textit{post-processing}.

\begin{figure}[!ht]
\includegraphics[width=0.7\textwidth]{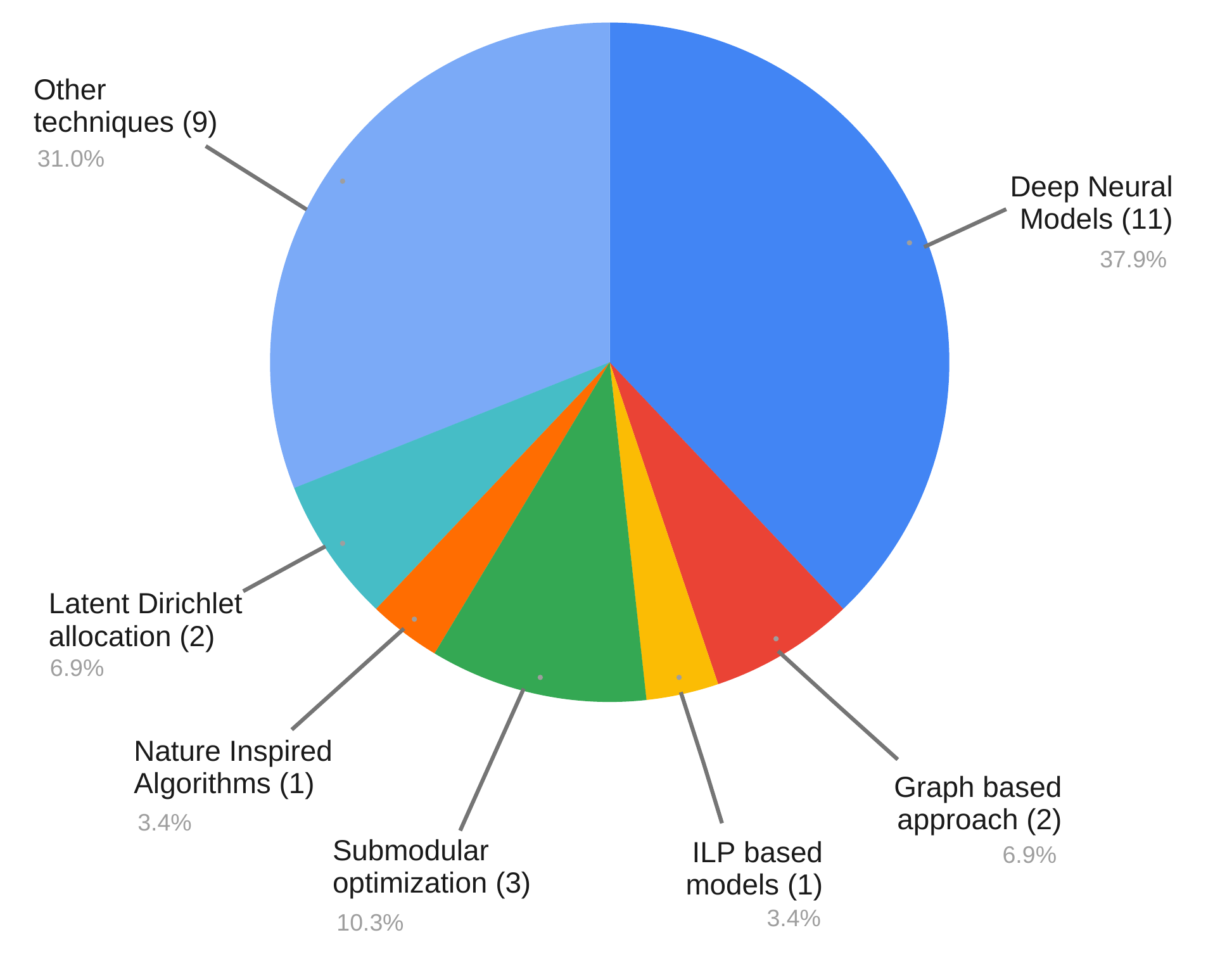}\centering
\caption{Illustration of techniques adopted to solve the MMS task.} \label{fig:model_cat}
\end{figure}

\subsection{Main Model}\label{sub:main-model}
A lot of different techniques have been adopted to perform the MMS task using extracted features. Figure \ref{fig:model_cat} illustrates the analysis of techniques adopted by researchers to solve the MMS task. We have tried to cover almost all the recent architectures that mainly focus on text-centric output summaries. In the approaches that have text as the central modality, the adjacent modalities are treated as a supplement to the text summaries, often getting selected at the post-processing step (Section \ref{sec:post-proc}).

\subsubsection{Neural Models} \label{subsec:neural}
\begin{figure}[t]
\includegraphics[width=\textwidth]{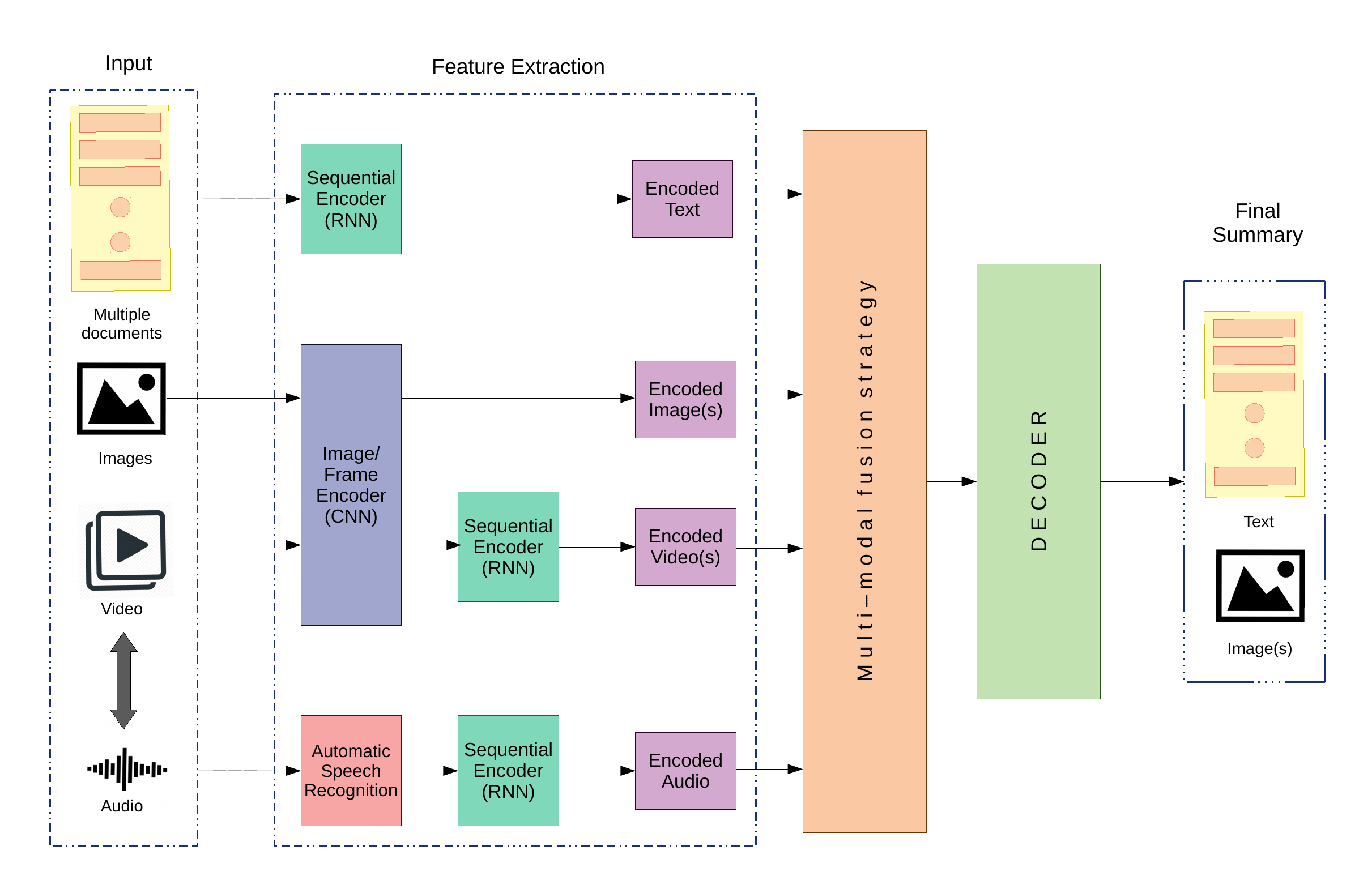}\centering
\caption{A generic framework portraying existing neural models.} \label{fig:neural}
\end{figure}

A few extractive summarization models \cite{chen2018extractive} and almost all of the abstractive text summarization based MMS architectures \cite{li2018multi,zhu2018msmo,chen2018abstractive,libovicky2018multimodal,zhu3multimodal} use Neural Networks (NN) in one form or another. Obtaining annotated dataset with sufficient instances to train these supervised techniques is the most difficult step for any Deep learning based MMS framework. The existing datasets satisfying these conditions belong to news domain, and have \textit{text-image} type input (refer to datasets \#4, \#5, \#6, \#7, \#19 in Table \ref{tab:datasets}) or \textit{text-audio-video} type input (refer to datasets \#17, \#18 in Table \ref{tab:datasets}). All these frameworks utilize the effectiveness of seq2seq RNN models for language processing and generation, and encoding temporal aspects in videos;  CNN networks are also adopted to encode discrete visual information in form of images \cite{zhu2018msmo, chen2018abstractive} and video frames \cite{li2020vmsmo, fu2020multi}. All the neural models have an encoder-decoder architecture at their heart, having three key elements: 1) a \textit{feature extraction module (encoder)}, 2) a \textit{summary generation module (decoder)}, and 3) a \textit{multi-modal fusion module}. Fig. \ref{fig:neural} describes a generic neural model to generate \textit{text-image}\footnote{We formulate \textit{text-image} summaries in our generic model since the existing neural models only output text \cite{li2018multi, chen2018extractive, fu2020multi} or text-image \cite{li2020vmsmo, zhu3multimodal, chen2018abstractive, zhu2018msmo} output.} summaries for multi-modal input. 

\vspace{1.5mm}
\noindent\underline{Feature Extraction (Encoder):} \textit{Encoder} is a generic term that entails both textual encoders as well as visual encoders. Various \textit{encoders} have been explored to encode contextual information in textual modality, ranging from \textit{sentence-level encoders} \cite{li2018multi} to \textit{hierarchical document level encoders} \cite{chen2018abstractive} with \textit{Long Short Term Memory (LSTM) units} \cite{hochreiter1997long} or \textit{Gated Recurrent Units (GRU)} \cite{cho2014learning} as the underlying RNN architecture. Most of the visual encoders do not train the parameter weights from scratch, but rather prefer to use CNN based pre-trained embeddings (refer to Section \ref{subsec:preprocess}). However, notably, in order to capture the contextual information of images, \citet{chen2018abstractive} used a bi-directional GRU unit to encode information from multiple images (encoded using VGGNet \cite{Simonyan15}) into one context vector, which is a unique approach for discrete image inputs. However, this RNN-CNN based encoding strategy is very standard approach adopted to  encoding video input. \citet{fu2020multi} and \citet{li2020vmsmo} in their respective works use pre-trained CNNs to encode individual frames, and then feed them as input to randomly initialized bi-directional RNNs to capture the temporal dependencies across these frames. \citet{libovicky2018multimodal} and \citet{palaskar2019multimodal} use ResNeXt-101 3D Convolutional Neural Network \cite{hara2018can} trained to recognize 400 diverse human actions on the Kinetics dataset \cite{kay2017kinetics} to tackle the problem of generating text summaries for tutorial videos from How2 dataset \cite{sanabria2018how2}.

\vspace{1.5mm}
\noindent\underline{Multi-modal fusion strategies:} A lot of fusion techniques have been developed in the field of MMS. However, most of the works that take text-image based inputs focus on \textit{multi-modal attention} to facilitate a smooth information flow across the two modalities. Attention strategies has proven to be a very useful technique to help discard noise and focus on relevant information \cite{vaswani2017attention}. The attention mechanism has been adopted by all the neural models that attempt to solve the MMS task. It has been applied to modal-specific information (\textit{uni-modal attention}), as well as at the information sharing step in form of \textit{multi-modal attention} to determine the degree of involvement of a specific modality for each input individually. \citet{li2018multi} proposed the hierarchical multi-modal attention for the first time to solve the task of multi-modal summarization of long sentences. The attention module comprises of individual text and image attention layers, followed by a subsequent layer of modality attention layer. Although multi-modal attention has shown great promise in text-image summarization tasks, it itself is not sufficient for text-video-audio summarization tasks \cite{li2020vmsmo}. Hence, to overcome this weakness, \citet{fu2020multi} proposed \textit{bi-hop attention} as an extension of bi-linear attention \cite{kim2016hadamard}, and \citet{li2020vmsmo} developed a novel \textit{conditional self-attention mechanism} module to capture local semantic information of video conditioned on the input text information. Both of these techniques were backed empirically, and established state-of-the-art in their respective problems.

\vspace{1.5mm}
\noindent\underline{Decoder:} Depending on the encoding strategy used, the textual decoders also vary from plain \textit{unidirectional RNN} \cite{zhu2018msmo} generating a word at a time to \textit{hierarchical RNN decoders} \cite{chen2018abstractive} performing this step in multiple levels of granularity. Although a vast majority of neural models focus only on generating textual summary using multi-modal information as input \cite{li2018multi, libovicky2018multimodal, chen2018extractive, palaskar2019multimodal, li2020aspect}, some work also output images as an supplement to the generated summary \cite{zhu2018msmo, chen2018abstractive, li2020vmsmo, zhu3multimodal, fu2020multi}; reinforcing the textual information and improving the user experience. These works either use a post-processing strategy to select the image(s) to become a part of final multi-modal summary \cite{zhu2018msmo, chen2018abstractive}, or they incorporate this functionality in their proposed model \cite{zhu3multimodal, fu2020multi, li2020vmsmo}. All the three frameworks that have implicit text-image summary generation characteristic adapt the final loss to be a weighted average of text generation loss together with the image selection loss. \citet{zhu3multimodal} treats the image selection as a classification task and adopts cross-entropy loss to train the image selector. \citet{fu2020multi} also treats the image selection process as a classification problem, and adopts an unsupervised learning technique that uses RL methods \cite{zhou2017deep}. The proposed technique uses representativeness and diversity as the two reward functions for the RL learning. \citet{li2020vmsmo} perceives the proposes a \textit{cover frame selector}\footnote{Model proposed by \citet{li2020vmsmo} only selects one image per input, chosen from the video frames.} that selects one image based on the hierarchical CNN-RNN based video encoding conditioned on article semantics using a \textit{conditional self-attention module}. \citet{li2020vmsmo} uses pairwise hinge loss to measure the loss during the model training.

Although the encoder-decoder model acts as the basic skeleton for the neural models solving MMS task, a lot of variations have been made, depending upon the input and output specifics. \citet{zhu2018msmo} proposes a visual coverage mechanism to mitigate the repetition of visual information. 
\citet{li2018multi} uses two image filters, namely \textit{image attention filter} and \textit{image context filter} to avoid noise introduction, filtering out useful information. \citet{zhu3multimodal} proposes a multi-modal objective function, that generates multi-modal summary at the end of this step, avoiding any statistical post-processing step for image selection. \citet{fu2020multi} utilizes the fact that audio and video are synchronous, and audio can easily be converted to textual format, utilizing these speech transcriptions as the the bridge across the asynchronous modalities of text and video. They also formulate various fusion techniques including \textit{early fusion} (concatenation of multi-modal embeddings), \textit{tensor fusion} \cite{zadeh2017tensor}, and \textit{late fusion} \cite{liu2018learn} to enhance the information representation in the latent space. 

\subsubsection{ILP-based Models}
Integer linear programming (ILP) has been used for text summarization in the past \cite{alguliev2010multi,galanis2012extractive}, primarily for extractive summarization. \citet{jangra2020text} has shown that if properly formulated, ILP can also be used to tackle the MMS task. More specifically, \citet{jangra2020text} attempt to solve the problem of generating multi-modal summaries from a multi-document multi-modal news dataset by extracting necessary sentences, images and videos. They propose a Joint Integer Linear Programming framework that optimizes weighted average of uni-modal salience and cross-modal correspondence. The model takes pre-trained joint embedding of sentences and images as input, and performs a shared clustering, generating $k_{txt}$ text clusters and $k_{img}$ image clusters. A recommendation-based setting is used to create the most optimal clusters. The text cluster centers are chosen to be the extractive text summary, and a multi-modal summary containing text, images and videos is generated at the post-processing step.

\subsubsection{Submodular Optimization based Models} \label{subsec:submod}
Sub-modular functions have been quite useful for text summarization tasks \cite{lin2010multi,sipos2012large} thanks to their assurance that the local optima is never worse than $1-\frac{1}{e}$ ($\approx$63\%) of the global optima \cite{nemhauser1978analysis}. A greedy algorithm having time complexity of $O(\textit{n} log\textit{n})$ is sufficient to optimize the functions. \citet{tiwari2018multimodal},  \citet{li2017multi}, and  \citet{modani2016summarizing} have also utilized these properties of submodular functions in order to solve the MMS task. \citet{tiwari2018multimodal} uses \textit{coverage}, \textit{novelty} and \textit{significance} as the submodular functions to extract the most significant documents for the task of timeline generation of a social media event in a multi-modal setting. \citet{li2017multi} proposes a linear combination of submodular functions (salience of text, redundancy and visual coverage in this case) under a budget constraint to obtain near-optimal solutions at a sentence level to obtain an extractive text summary using news input comprising of text, images, videos and audio. Modani et al. \cite{modani2016summarizing} uses a weighted sum of five submodular functions (coverage of input text/images, diversity of text/images in final summary, and coherence of text part and image part of the final summary) to generate a summary comprising of text and images.


\subsubsection{Nature Inspired Algorithms}
Genetic algorithms \cite{saini2019extractive2} and other nature inspired meta-heuristic optimization algorithms like the Grey Wolf Optimizer \cite{mirjalili2014grey} and Water Cycle algorithm \cite{eskandar2012water} have shown great promise for extractive text summarization \cite{saini2019extractive}.  \citet{jangra2020multimodal} has illustrated that such algorithms can also be useful in multi-modal scenarios by experimenting with a multi-objective setting using differential evolution as the underlying guidance strategy. For the multi-objective optimization setup, the authors have proposed two different sets of objectives: one redundancy based (including uni-modal salience, redundancy and cross-modal correspondence) and one using cluster validity indices (PBM index \cite{pakhira2004validity} was used in this case). Both of these settings have performed better than the baselines. The optimization setup outputs the top most suitable sentences and images, which follow similar post-processing procedure as \citet{jangra2020text}. \citet{jangra2021multimodal} on the other hand used Grey Wolf Optimizer \cite{mirjalili2014grey} based multi-objective optimization strategy to obtain the combined complementary-supplementary multi-modal summaries. The proposed approach was split into two key steps: a) global coverage text format (GCTF) - obtaining extractive text summaries using Grey Wolf Optimizer over all the input modalities in a clustering setup, b) visual enhanced text summaries (VETS) - using one-shot population based strategy to enhance the obtained text summaries with visual modalities to obtain the complementary and supplementary enhancements in a data-driven manner. The overall pipeline adopted similar pre-processing and post-processing steps as \citet{jangra2020multimodal}.

\subsubsection{Graph based Models}
Graph based techniques have been widely adopted in extractive text summarization frameworks \cite{mihalcea2004graph, mihalcea2004textrank, erkan2004lexrank, modani2015creating}. These techniques involve graph formulation of text documents where nodes are represented by document sentences and the edge weights are formulated using similarity across two sentences. Extending this idea to a multi-modal set up, Modani et al. \cite{modani2016summarizing} proposed a graph based approach to generate text-image summaries. A graph was constructed using \textit{content segments} (representing either sentences or images) as the nodes, and each node is given a weight depending on its information content. For sentences, this weight is computed as the sum of \#nouns, \#adverbs, \#adjectives, \#verbs, and half the \#pronouns, while an images node's weight is given by the average similarity score with all other image segments. An edge weight for an edge connecting two sentences is computed as the cosine similarity of sentence embeddings (evaluated using auto-encoders), edge weight connecting two images is computed as the cosine similarity of image embeddings (evaluated using VGGNet \cite{Simonyan15}) and the edge weight connecting a sentence and an image is computed as the cosine similarity of sentence embedding and image embedding projected in a shared vector space (using Deep Fragment embeddings \cite{karpathy2014deep}). After graph construction, an iterative greedy strategy \cite{modani2015creating} is adopted to select appropriate content segments and generate the \textit{text-image summary}. 

Li et al. \cite{li2017multi} also use a graph based technique to evaluate the salience of text to generate an extractive text summary using multi-modal input (containing text documents, images, videos). A guided LexRank \cite{erkan2004lexrank} was proposed to evaluate the salience score of the text unit (comprising of document sentences and speech transcriptions). The guidance strategy proposed by Li et al. \cite{li2017multi} had bidirectional connections for sentences belonging to documents, but only unidirectional connections were made for speech transcriptions with only outward edges to follow on their assumption that speech transcriptions might not always be grammatically correct, and hence should only be used for guidance and not for summary generation. This textual score was then used as a submodular function for the final model (refer to Sec \ref{subsec:submod}). 

\subsection{Post-processing} \label{sec:post-proc}
Most of the existing works are not capable of generating multi-modal summaries\footnote{Although, all the surveyed methods are "multi-modal summarization" approaches, i.e. they all summarize multi-modal information, however, most of them summarize it to generate uni-modal outputs.}. The systems that do generate multi-modal summaries either have an inbuilt system capable to generating multi-modal output (mainly by generating text using seq2seq mechanisms and selecting relevant images) \cite{li2020vmsmo, zhu3multimodal} or they adopt some post-processing steps to obtain the visual and vocal supplements of the generated textual summaries \cite{jangra2020text, zhu2018msmo}. Neural network models that use multi-modal attention mechanisms to determine the relevance of modality for each input case have been used for selecting the most suitable image \cite{zhu2018msmo,chen2018abstractive}. More precisely, the visual coverage scores (after the last decoding step), i.e. the summation of attention values while generating the text summary, are used to determine the most relevant images. Depending upon the needs of the task, a single image \cite{zhu2018msmo} as well as multiple images \cite{chen2018extractive} can be extracted to supplement the text. 

\begin{figure}[!ht]
  \centering
  \subfloat[Input image distribution]{\includegraphics[width=0.5\textwidth]{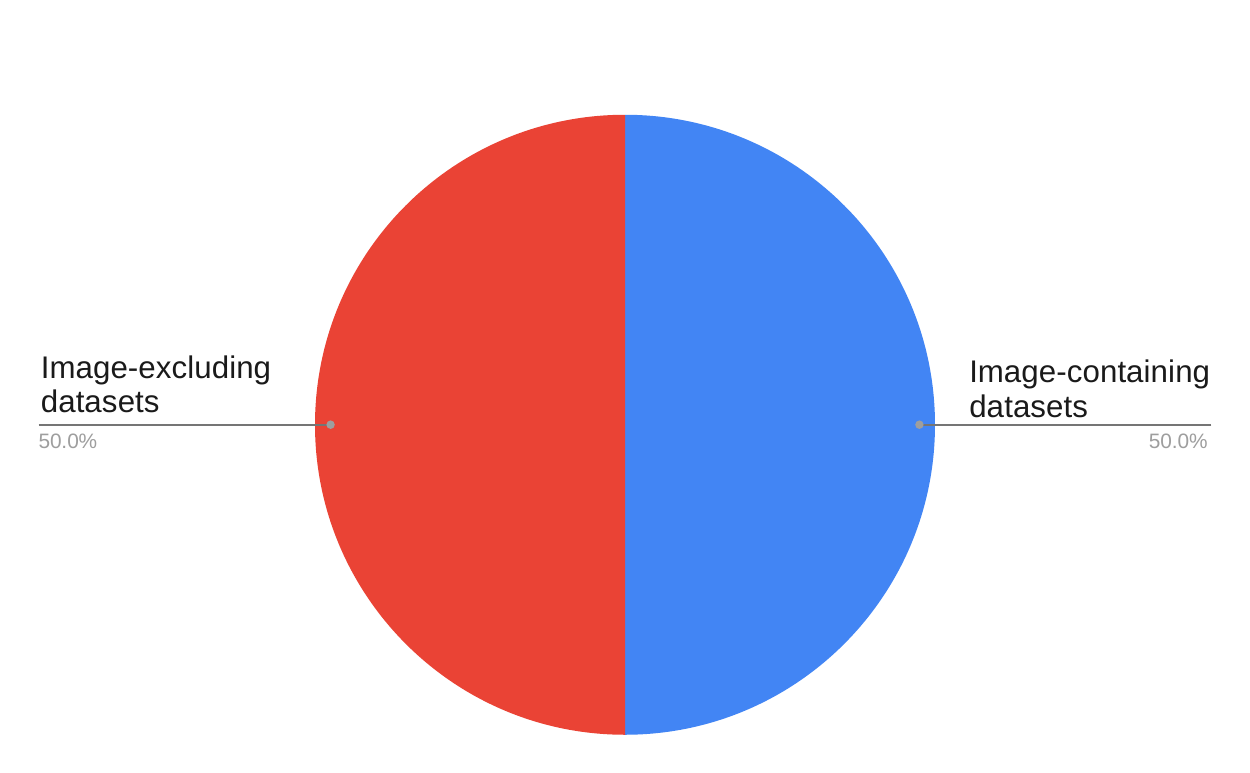}}
  \subfloat[Input audio/video distribution]{\includegraphics[width=0.5\textwidth]{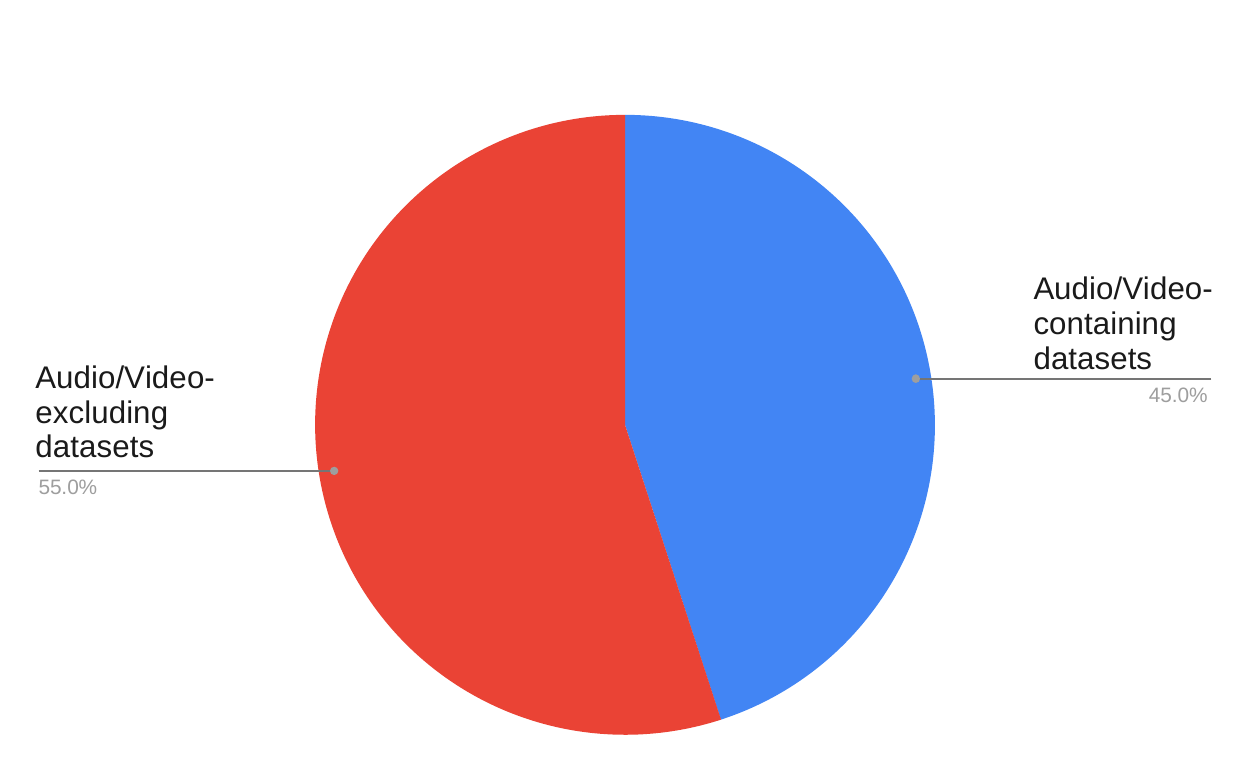}}
  \\
  \subfloat[Language distribution in datasets.]{\includegraphics[width=0.5\textwidth]{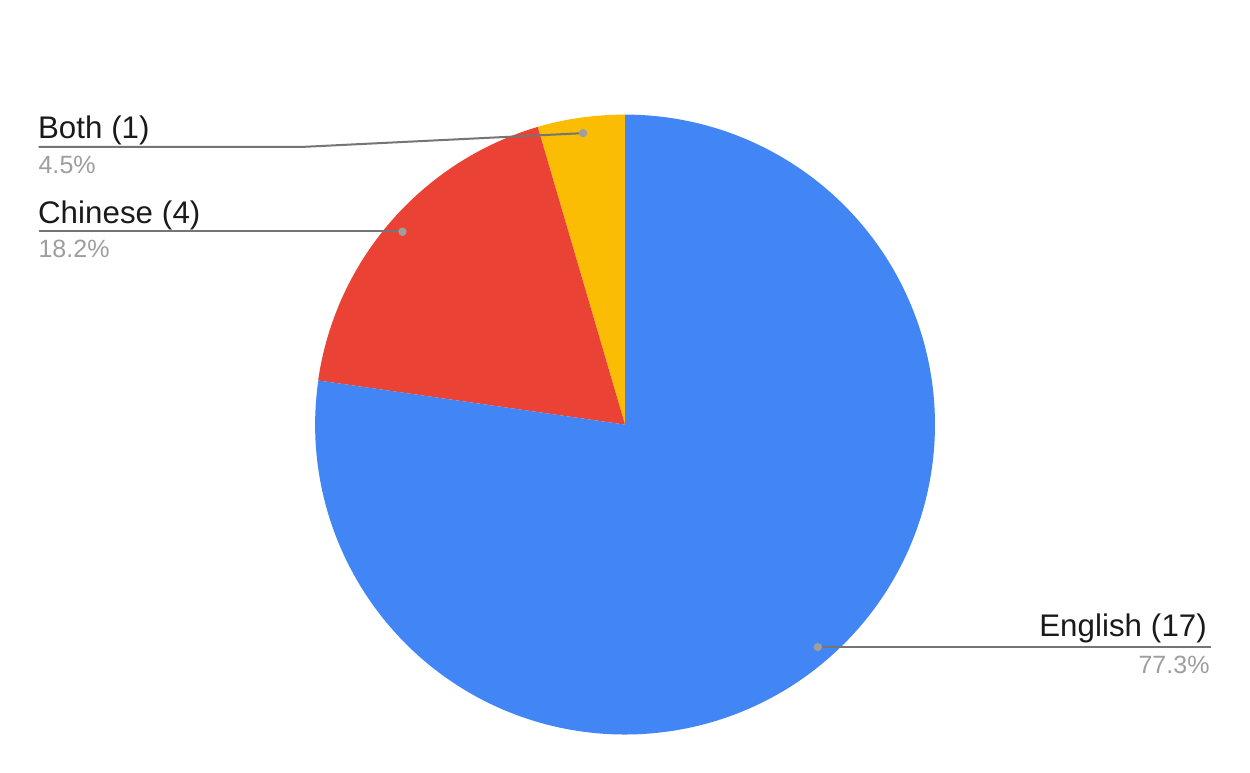}}
  \subfloat[Abstractive/Extractive text output]{\includegraphics[width=0.5\textwidth]{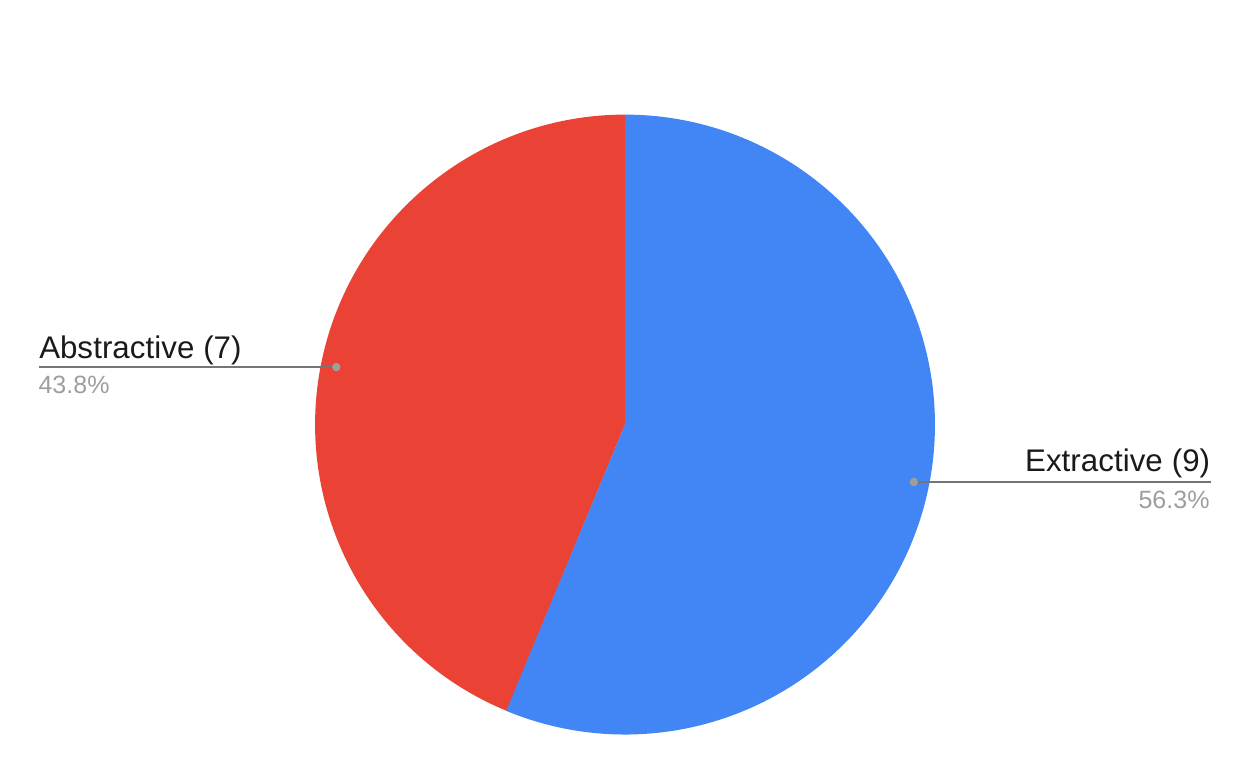}}
  \\
  \caption{Dataset statistics.} \label{fig:data}
\end{figure}

\citet{jangra2020text} proposes a text-image-video summary generation task, which as the name suggests, outputs all possible modalities in the final summary. Having extracted most important sentences and images (containing video key-frames as well) using the ILP framework, the images are separated from the key-frames, and are supplemented with other images from the input set that have a moderate similarity, with a pre-determined threshold and upper bound to avoid noisy and redundant information. Cosine similarity of global image features is used as the underlying similarity matrix in this case. A weighted average of verbal scores and visual scores is used to determine the most suitable video for the multi-modal summary. \textit{verbal score} is defined as the information overlap between speech transcriptions and generated text summary, while the \textit{visual score} is defined as the information overlap between the key-frames of a video with the generated image summary. 

\section{Datasets and Evaluation Techniques}
\label{sec:data}
Due to the flexible nature of the MMS task, with a large variety of input-output modalities, the MMS task does not have a standard dataset used as a common evaluation benchmark for all approaches to this date. Nonetheless, we have collected information about datasets used in the previous works, and a comprehensive study of 20 datasets can be found at Table \ref{tab:datasets}. 
It was found that out of these 21 datasets, 12 datasets are of news-related origin \cite{jangra2020text, li2018multi, zhu2018msmo, fu2020multi, yan2012visualizing}, and including the dataset on video tutorials by \citet{sanabria2018how2}, there are 13 datasets that are domain-independent, thus suitable to test out domain generic models. 6 out of the 21 datasets produce text-only summaries using multi-modal input; out of these six datasets, 2 datasets' output comprises of extracted text summaries \cite{chen2018extractive, li2017multi} and 4 datasets' output contains abstractive summaries \cite{li2018multi, li2020aspect, fu2020multi, sanabria2018how2}. One the other hand, there are 8 datasets that output text-image summaries, which can further be divided into 6 extractive text-image summary generation datasets \cite{xu2013cross, tiwari2018multimodal, bian2013multimedia} and 2 abstractive text-image summary generation datasets \cite{li2020vmsmo, chen2018abstractive}. Datasets \#19 (\cite{jangra2020text}) and \#20 (\cite{jangra2021multimodal}) are the only two datasets that comprise of text, image, audio and video in the output. However, these datasets are small, and thus limited to extractive summarization techniques. Meanwhile, dataset \#20 (\cite{jangra2021multimodal}) is the only existing dataset that comprises of both complementary and supplementary enhancements in the multi-modal summary (refer to Section \ref{sec:mms} for the definition). Out of the 21 datasets, 17 datasets contain text in the multi-modal summary, 11 contain images as well, 3 comprises solely of audio-video outputs \cite{evangelopoulos2009video, evangelopoulos2013multimodal, sanabria2019deep}, and 1 dataset has a fixed template\footnote{\citet{tjondronegoro2011multi} focusing on summarizing tennis matches, and thus the output has a fixed template comprising of three different summarization tasks: a) summarization of entire tournament, b) summarization of a match and c) summarization of a tennis player.} as output \cite{tjondronegoro2011multi}. Of these 17 text-containing datasets, 10 datasets contain extractive text summaries \cite{xu2013cross, tiwari2018multimodal, bian2013multimedia, jangra2020text} and the rest 7 datasets contain abstractive summaries \cite{li2020vmsmo, chen2018abstractive, fu2020multi, sanabria2018how2}. It is interesting to note that 5 out of these 7 abstractive datasets belong to the news-domain \cite{li2020vmsmo, chen2018abstractive, fu2020multi, li2018multi, zhu2018msmo}, while the other two focus on e-commerce product summarization \cite{li2020aspect} and tutorial summarization \cite{sanabria2018how2}.Out of the $21$ datasets only $4$ \citep{li2017multi,sanabria2018how2,jangra2021multimodal,zhu2018msmo} of them are publicly available.

\begin{figure}[ht] \ContinuedFloat
  \subfloat[Single vs Multiple document distribution for datasets]{\includegraphics[width=0.5\textwidth]{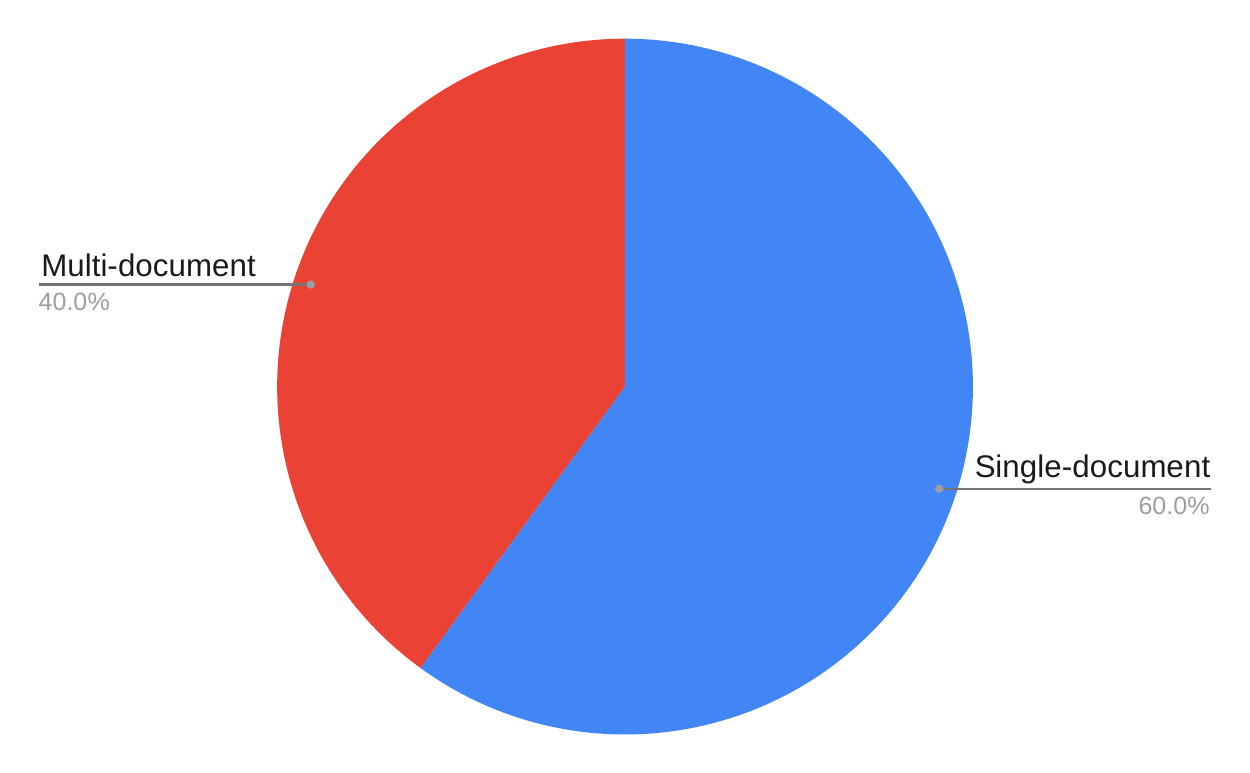}}
  \subfloat[Domain distribution for datasets]{\includegraphics[width=0.5\textwidth]{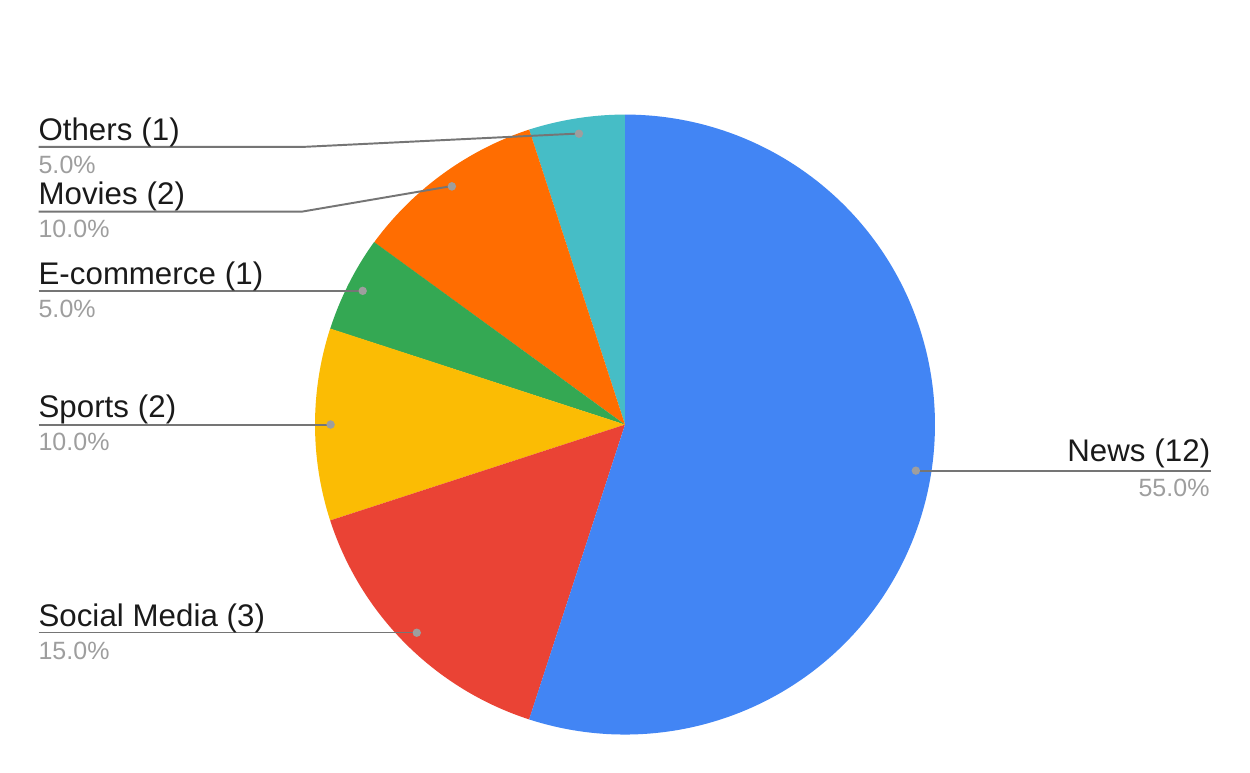}}
  \\
  \subfloat[Input modality distribution]{\includegraphics[width=0.5\textwidth]{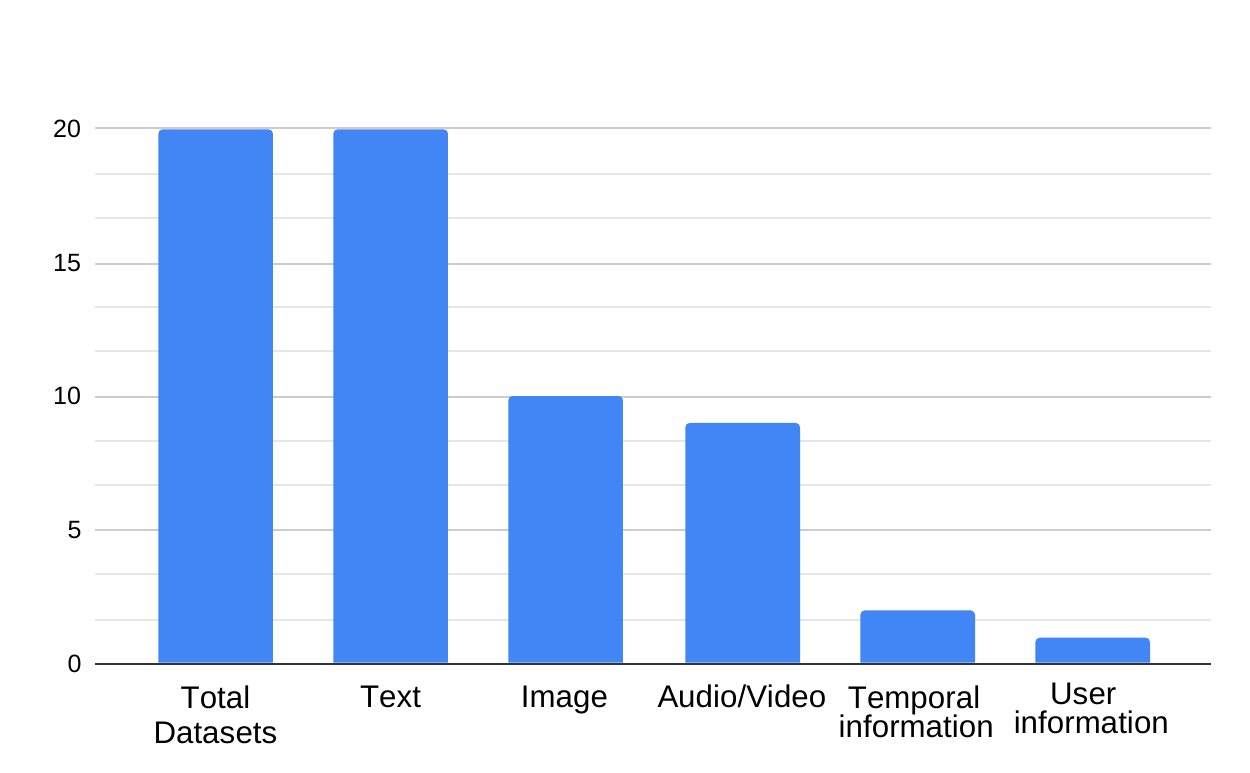}}
  \subfloat[Output modality distribution]{\includegraphics[width=0.5\textwidth]{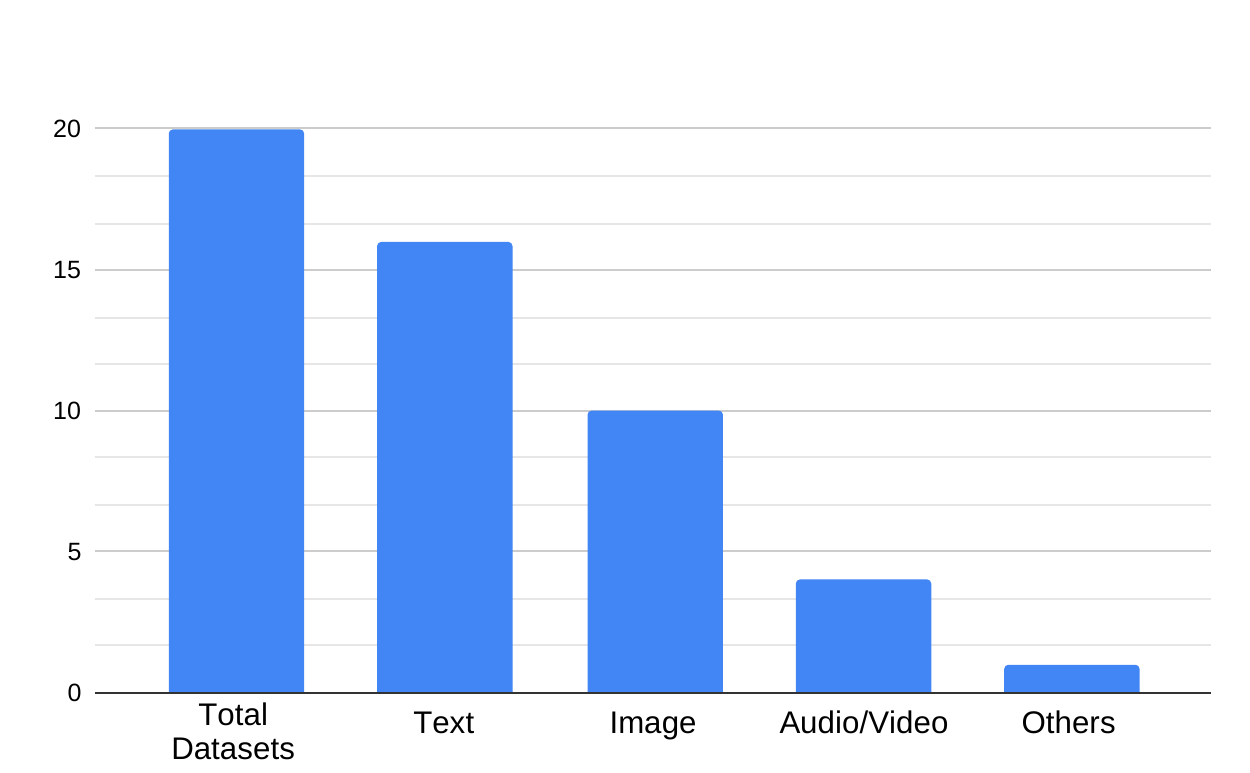}}
  \caption{Dataset statistics (cont.).}
\end{figure}

Depending on the input, we can also divide the 20 datasets based on the presence/absence of video in the input. There are 10 datasets that contain videos, whereas the rest 11 mostly work with text-image inputs. Due to the nature of this survey (the main focus on text modality), all 21 datasets in consideration contain text as input. A majority of these text sources are single documents \cite{li2020vmsmo, chen2018abstractive, li2020aspect, zhu2018msmo}, but there are 6 datasets that have multiple documents in the input \cite{li2017multi, jangra2020text, xu2013cross, bian2013multimedia, bian2014multimedia, jangra2021multimodal}. \citet{sanabria2019deep}, \citet{evangelopoulos2009video} and \citet{evangelopoulos2013multimodal} however do not contain text documents, but the speech transcriptions from corresponding audio inputs. While most of these datasets comprise of multi-sentence summaries generated from input documents, \citet{li2018multi} contains a single sentence as the source as well as the reference summary. Most of these datasets use English-based text and audio, but there are 3 datasets that contain Chinese text \cite{bian2013multimedia, li2017multi, bian2014multimedia, li2020vmsmo}. 

There are some datasets that have inputs other than text, image, audio and video. For instance, \citet{tiwari2018multimodal} and \citet{yan2012visualizing} contain temporal information about the dataset for the task of multi-modal timelines generation. \citet{qian2018multimodal} also utilize user information including demographics like gender, birthday, user profile (short biography), and other information including user name, nickname, number of followers, number of microblogs posted, profile registration time, and user's level of interests in different topics for generating summaries of an event based on social  media content. Detailed plots for selected statistics on the datasets covered in this study can be found at Figure \ref{fig:data}.

\begin{table}[ht]
\centering
\tiny
\caption{\textbf{A study on datasets available for multi-modal summarization.} `T' stands for English text, `TC' stands for Chinese text, `TF' stands for text (template filling), `TE' stands for text (extractive), `TA' stands for text (abstractive), `I' stands for images, `V' stands for video, `A' stands for audio, `U' signifies user information, and `TM' denotes existence of temporal information about the data such as publication date. The `\textbf{*}' denotes publicly available datasets and the `-' denotes the unavailability of details.}\label{tab:datasets}
\renewcommand{\arraystretch}{2}
\begin{tabular}{|l|p{0.06\textwidth}|p{0.08\textwidth}| p{0.09\textwidth}| p{0.3\textwidth}|p{0.08\textwidth}|}
\hline
\textbf{ID \& Paper} &
\textbf{Used In Paper} &
\textbf{Input Modalities} & \textbf{Output Modalities} & \textbf{Data Statistics} & \textbf{Domain} \\
\hline
\#1: \citet{li2018multi} (2018) & \cite{li2018multi} & T, I & TA & 66,000 triplets (sentence, image and summary) & News \\\hline
\#2: \citet{zhu2018msmo}(2018)\textbf{*}  &\cite{zhu2018msmo, zhu3multimodal} & T, I & TA, I & 313k documents,  2.0m images & News \\\hline
\#3: \citet{chen2018abstractive} (2018) &\cite{chen2018abstractive}  & T, I & TA, I & 219k documents & News \\\hline
\#4: \citet{xu2013cross} (2013) & \cite{xu2013cross} & T, I & TE, I & 8 topics (each containing 150+ documents) & News \\\hline
\#5: \citet{bian2013multimedia} (2013) & \cite{bian2013multimedia} & TC, I & TE, I & 10 topics (127k microblogs and 48k images) & Social Media \\\hline
\#6: \citet{bian2014multimedia} (2014) & \cite{bian2014multimedia} & TC, I & TE, I & 20 topics (310k documents, 114k images) & Social Media \\\hline
\#7: \citet{li2020aspect} (2020) & \cite{li2020aspect} & TC, I & TA & 1,375,453 instances from home appliances, clothing, and cases \& bags categories & E-commerce \\\hline
\#8: \citet{chen2018extractive} (2018) & \cite{chen2018extractive} & T, A & TE & - & News \\\hline
\#9: \citet{tiwari2018multimodal} (2018) & \cite{tiwari2018multimodal} & T, I, TM & TE, I & 6 topics & Social Media \\\hline
\#10: \citet{yan2012visualizing} (2012) & \cite{yan2012visualizing} & T, I, TM & TE, I & 4 topics (6k documents, 2k images) & News \\\hline
\#11: \citet{qian2019social} (2019) & \cite{qian2019social} & T, I, U & TE, I & 12 topics (9.1m documents,  2.2m users, 15m images) & News (disasters) \\\hline
\#12: \citet{tjondronegoro2011multi} (2011) & \cite{tjondronegoro2011multi} & T, A, V & TF & 66 hrs video (33 matches), 1,250 articles related to Australian Open 2010 tennis tournament & Sports (Tennis) \\\hline
\#13: \citet{sanabria2018how2} (2018)\textbf{*} & \cite{libovicky2018multimodal} & T, A, V & TA & 2,000 hrs video & Multiple domains \\\hline
\#14: \citet{fu2020multi} (2020) & \cite{fu2020multi} & T, A, V & TA & 1970 articles from Daily Mail (avg. video length 81.96 secs), and 203 articles from CNN (avg. video length 368.19 secs) & News \\\hline
\#15: \citet{li2020vmsmo} (2020) & \cite{li2020vmsmo} & T, A, V & TA, I & 184,920  articles (Weibo) with avg. video duration 1 min, avg. article length 96.84 words, avg. summary length 11.19 words & News \\\hline
\#16: \citet{sanabria2019deep}  (2019) & \cite{sanabria2019deep}& T, A, V & A, V & 20 complete soccer games from 2017-2018 season of French Ligue 1 & Sports (Soccer / Football) \\\hline
\#17: \citet{evangelopoulos2009video} (2009) &\cite{evangelopoulos2009video}  & T, A, V & A, V & 3 movie segments (5-7 min each) & Movies \\\hline
\#18: \citet{evangelopoulos2013multimodal} (2013) & \cite{evangelopoulos2013multimodal} & T, A, V & A, V & 7 half hour segments of movies & Movies \\\hline 
\#19: \citet{jangra2020text} (2020) & \cite{jangra2020text,jangra2020multimodal} & T, I, A, V & TE, I, A, V & 25 topics (500 documents, 151 images, 139 videos)  & News \\\hline
\#20: \citet{jangra2021multimodal} (2021)\textbf{*} & \cite{jangra2021multimodal} & T, I, A, V & TE, I, A, V & 25 topics (contains complementary and supplementary multi-modal references) & News \\\hline
\#21: \citet{li2017multi} (2017)\textbf{*} & \cite{li2017multi}& T, TC, I, A, V & TE & 25 documents in English, 25 documents in Chinese & News \\\hline
\end{tabular}
\end{table}

These datasets span a wide variety of domains, including sports like tennis \cite{tjondronegoro2011multi} and football \cite{sanabria2019deep}, movies \cite{evangelopoulos2009video, evangelopoulos2013multimodal}, social media-based information \cite{bian2013multimedia, bian2014multimedia, tiwari2018multimodal}, e-commerce \cite{li2020aspect}. In the coming days, we are likely bound to see more large-scale domain specific datasets to advance this field. 

Although there have been a lot of innovative attempts in solving the MMS task, the same does not go for the evaluation techniques used to showcase the quality of generated summaries. Most of the existing works use uni-modal evaluation metrics, including ROUGE scores \cite{lin-2004-rouge} to evaluate the text summaries, accuracy and precision-recall based metrics to evaluate the image and video parts of generated summaries. A few works have also reported \textit{True Positives} and \textit{False Positives} as well \cite{sahuguet2013socially}. The best way to evaluate the quality of a summary is to perform extensive human evaluations. Various techniques have been used to get the best user performance evaluations including the quiz method \cite{erol2003multimodal},
and user-satisfaction test \cite{zhu2018msmo}. These manual evaluation techniques are mainly of two kinds: a) simple scoring of summary quality based on input \cite{zhu3multimodal, zhu2018msmo, jangra2021multimodal}, and b) answering 
the questions based on input to quantify the information retention of input data instance \cite{li2020vmsmo}. However, one major issue with these manual evaluations is that they cannot be conducted for the entire dataset, and are hence performed on a subset of the test dataset. There are a lot of uncertainties involving this subset, as well as the mental conditions of the human evaluators while performing these quality checks. Hence it can be unreliable to compare two results of separate human evaluation experiments, even for the same task.

\subsection{Text Summary Evaluation Techniques}
Since the scope of this work is mostly limited to text-centric MMS techniques,
it is important to discuss evaluation of text summaries separately and in tandem to other modalities. Even though quite a few MMS works generate uni-modal text summaries from multi-modal inputs \cite{li2018multi, li2017multi, libovicky2018multimodal}, they still use very basic string based n-gram overlap metrics like ROUGE \cite{lin-2004-rouge} to conduct the evaluation. Through this survey, we want to direct the researchers to not just focus on ROUGE, but also look at other aspects of text summarization as well. For instance, \citet{fabbri2021summeval} proposes four key-characteristics that an ideal summary must have:
\begin{enumerate}
    \item \textbf{Coherence} the quality of smooth transition between different summary sentences such that sentences are not completely unrelated or completely same.
    \item \textbf{Consistency} the factual correctness of summary with respect to input document.
    \item \textbf{Fluency} the grammatical correctness and readability of sentences.
    \item \textbf{Relevance} the ability of a summary to capture important and relevant information from the input document.
\end{enumerate}

\citet{fabbri2021summeval} also illustrated how ROUGE is not capable of gauging the quality of generated summaries by doing an n-gram overlap with human written reference summaries. There are other metrics out there that use more advanced strategies to do the evaluation, such as n-gram based metric like WIDAR \cite{10.1007/978-3-030-99736-6_21}, embedding-based metrics like ROUGE-WE \cite{ng-abrecht-2015-better}, MoverScore \cite{zhao-etal-2019-moverscore} and Sentence Mover Similarity (SMS) \cite{clark-etal-2019-sentence} or neural model-based metrics like BERTScore \cite{Zhang*2020BERTScore:}, SUPERT \cite{gao2020supert}, BIANC \cite{Lita2005BLANCLE} and S$^3$ \cite{peyrard-etal-2017-learning}. These evaluation metrics have been proven to be empirically better at captivating the above-mentioned characteristics of a summary \cite{fabbri2021summeval}, and hence upcoming research works should also report performance on some of these metrics along with ROUGE to have more accurate analysis of the generated summaries.

\subsection{Multi-modal Summary Evaluation Techniques}

In an attempt to evaluate the multi-modal summaries, \citet{zhu2018msmo} proposes a multi-modal automatic evaluation (MMAE) technique that jointly considers uni-modal salience and cross-modal relevance. In their case, the final summary comprises of text and images, and the final objective function is formulated as a mapping of three objective functions: 1) salience of text, 2) salience of images, and 3) text-image relevance. This mapping function is learnt using supervised techniques (Linear Regression, Logistic Regression and Multi-layer Perceptron in their case) to minimize training loss with human judgement scores. Although the metric seems promising, there are a lot of conditions that must be met in order to perform the evaluation. 

The MMAE metric does not effectively evaluate the information integrity\footnote{Information integrity is the dependability or trustworthiness of information. In context of a multi-modal summary evaluation task, it refers to the ability to make a judgement that is unbiased towards any modality (i.e. an ideal evaluation metric does not give higher importance to information from one modality (e.g. text) over other modality (e.g. images)).} of a multi-modal summary, since it uses uni-modal salience scores as a feature in the overall judgement making process, leading to a cognitive bias.  \citet{zhu3multimodal} improves upon this by proposing an evaluation metric based on joint multi-modal representation (termed as MMAE++), projecting the generated summaries and the ground truth summaries into a joint semantic space. In contrast to other multi-modal evaluation metrics, they attempt to look at the multi-modal summaries as a whole entity, than a combination of piece-wise significant elements. A neural network based model is used to train this joint representation. Images of two image caption pairs are swapped to obtain two image-text pairs that are semantically close to each other to obtain the training data for joint representation automatically. The evaluation model is trained using a multi-modal attention mechanism \cite{li2018multi} to fuse the text and image vectors, using max-margin loss as loss function.

\citet{modani2016summarizing} propose a novel evaluation technique termed by them as \textit{Multimedia Summary Quality (MuSQ)}. Just like other multi-modal summarization metrics described above, \textit{MuSQ} is also limited to text-image summaries. However, unlike the majority of previous evaluation metrics for multi-modal summarization or document summarization techniques \cite{ermakova2019survey}, \textit{MuSQ} does not require a ground truth to evaluate the quality of generated summary. \textit{MuSQ} is a naive coverage based evaluation metric denoted as $\mu_M$, and is defined as:
\begin{gather}
    \mu_M = \mu_T + \mu_I + \sigma_{T,I}\\
    \mu_T = \sum_{v\in T} R_v * max_{u\in S}\{Sim(u,v)\}\\
    \mu_I = \sum_{w\in V} \hat{R_w} * max_{x\in I}\{Sim(w,x)\}\\
    \sigma_{T,I} = \sum_{v\in S}\sum_{w\in I} \{Sim(w,x) * R_v * \hat{R_w}\}
\end{gather}

where $\mu_T$ denotes the degree of coverage of input text document $T$ by text summary $S$, $\mu_I$ denotes the degree of coverage of input images $V$ by the image summary $I$. $\sigma_{T,I}$ measures the cohesion across the text sentences and images of final multi-modal summary. $R_v$ and $\hat{R_w}$ are respectively the individual reward values for each input sentence and input image that denote the extent of information content in each content fragment (a text sentence or an image).

\begin{table}[ht]\centering
\scriptsize
\caption{\textbf{Comparative study of evaluation techniques for evaluation techniques on multi-modal summarization.}}\label{tab:evaluation}
\renewcommand{\arraystretch}{2}
\begin{threeparttable}
\begin{tabular}{|p{0.3\textwidth}| p{0.6\textwidth}|}
\hline
\textbf{Metric name \& corresponding paper} & \textbf{Pros \& Cons} \\
\hline

\multirow{5}{0.2\textwidth}{Multi-modal Automatic Evaluation (MMAE). \citet{zhu2018msmo}} & \textbf{Advantages} \\
& - MMAE shows high correlation with human judgement scores. \\
& \textbf{Disadvantages} \\
& - Requires a substantial manually annotated dataset. \\
& - Might perform ambiguously$^\ddagger$ for evaluation of other new domains. \\
\hline

\multirow{5}{0.2\textwidth}{MMAE++. \citet{zhu3multimodal}} & \textbf{Advantages} \\
& - Utilizes joint multi-modal representation of sentence-image pairs to better improve the correlation scores over MMAE metric \cite{zhu2018msmo}. \\
& \textbf{Disadvantages} \\
& - Requires a substantial manually annotated dataset. \\
& - Might perform ambiguously\tnote{1} for evaluation of other new domains. \\
\hline

\multirow{5}{0.2\textwidth}{Multimedia Summary Quality (MuSQ). \citet{modani2016summarizing}} & \textbf{Advantages} \\
& - Does not require manually created gold summaries. \\
& \textbf{Disadvantages} \\
& - The technique is very naive, and only considers coverage of input information and text-image cohesiveness. \\
& - The metric output is not normalized. Hence the evaluation scores are highly sensitive to the cardinality of input text sentences and input images. \\
\hline
\end{tabular}
\begin{tablenotes}
    \item[1] Here "might perform ambiguously" refers to the fact that since model-based metrics are biased towards the training data, it is hard to determine how well would they perform on unseen domains. For instance, if the model is trained on news summarization dataset, and the task is to evaluate medical report summaries, then the model performance cannot be determined without further experiments.
\end{tablenotes}
\end{threeparttable}
\end{table}

To sum up, only a handful of works have focused on the evaluation of multi-modal summaries. Even the proposed evaluation metrics have a lot of drawbacks. The evaluation metrics proposed by \citet{zhu2018msmo} and \citet{zhu3multimodal} require a large human evaluation score-based training data to learn the parameter weights. Since these metrics are highly dependent on the human-annotated dataset, the quality of this dataset can compromise the evaluation process if the training dataset is restrictive in domain coverage or is of poor quality. It also becomes difficult to generalize these metrics since they depend on the domain of training data. The evaluation technique proposed by \citet{modani2016summarizing}, although independent from gold summaries, is too naive, and has its own drawbacks. The evaluation metric is not normalized, and hence shows great variation when comparing the results of two input data instances with different sizes. 

Overall, the discussed strategies have their own pros and cons; however, there is a great scope for future improvement in the area of `evaluation techniques for multi-modal summaries' (refer to Section \ref{sec:future}).


\begin{table}[tp]  \centering
\caption{\textbf{Results of different methods for text and image output modalities.} This study is limited to works that contain text in the generated multi-modal summary\textsuperscript{$\dagger$}. Note that the comparison should be done with care as most of the proposed approaches use different datasets (the ``Dataset No.'' column corresponds to the ID column in Table \ref{tab:datasets}). Column `ME' indicates presence/absence of manual evaluation in the corresponding work. Here 'N.A' or Not Available is used to denote the unavailablity of images in the output or unavailabilty of scores for an evaluation metric. '(ABS)' denotes abstractive summarization type and '(EXT)' denotes extractive summarization type. \newline \small\textsuperscript{$\dagger$ For population based techniques \cite{jangra2020multimodal, jangra2021multimodal}, the best score across multiple solutions were reported in this work.}}\label{tab:results}

\renewcommand{\arraystretch}{2}
\resizebox{\textwidth}{!}{%
\begin{tabular}{|l|l|l|c|c|c|c|c|c|c|c|} \hline
\multirow{2}{*}{\textbf{Paper}} & \multirow{2}{*}{\textbf{Dataset No.}} & \multirow{2}{*}{\textbf{Domain}} & \multicolumn{4}{c|}{\textbf{Text score (ROUGE)}} & \multicolumn{3}{c|}{\textbf{Image score}} & \multirow{2}{*}{\textbf{ME}} \\ \cline{4-10}
 &  & & \textbf{R-1} & \textbf{R-2} & \textbf{R-L} & \textbf{R-SU4} & \textbf{Precision} & \textbf{Recall} & \textbf{MAP} & \\ \hline
\multirow{2}{*}{\citet{li2017multi} (EXT)} & \citet{li2017multi} (English) & News & 0.442 & 0.133 & N.A & 0.187 & N.A & N.A & N.A & \checkmark \\
 & \citet{li2017multi} (Chinese) & & 0.414 & 0.125 & N.A & 0.173 & N.A & N.A & N.A & \checkmark \\ \hline
\citet{li2018multi} (ABS) & \citet{li2018multi} & News & 0.472 & 0.248 & 0.444 & N.A & N.A & N.A & N.A & \\ \hline
\citet{zhu2018msmo} (ABS) & \citet{zhu2018msmo} & News & 0.408 & 0.1827 & 0.377 & N.A & 0.624 & N.A & N.A & \checkmark \\
\citet{zhu3multimodal} (ABS) & \citet{zhu2018msmo}& & 0.411 & 0.183 & 0.378 & N.A & 0.654 & N.A & N.A & \checkmark \\ \hline
\citet{chen2018extractive} (EXT) & \citet{chen2018extractive} & News & 0.271 & 0.125 & 0.156 & N.A & N.A & N.A & N.A & \\ \hline
\citet{chen2018abstractive} (ABS) & \citet{chen2018abstractive} & News & 0.326 & 0.120 & 0.238 & N.A & N.A & 0.4978 & N.A & \\ \hline
\citet{libovicky2018multimodal} (ABS) & \citet{sanabria2018how2}& Multi-domain & N.A & N.A & 0.549 & N.A & N.A & N.A & N.A & \checkmark \\ \hline
\citet{jangra2020text} (EXT) & \citet{jangra2020text} & News & 0.260 & 0.074 & 0.226 & N.A & 0.599 & 0.38 & N.A & \\
\citet{jangra2020multimodal} (EXT) & \citet{jangra2020text} & & 0.420 & 0.167 & 0.390 & N.A & 0.767 & 0.982 & N.A & \\ \hline
\citet{jangra2021multimodal} (EXT) & \citet{jangra2021multimodal}& News & 0.556 & 0.256 & 0.473 & N.A & 0.620 & 0.720 & N.A & \checkmark \\ \hline
\citet{xu2013cross} (EXT) & \citet{xu2013cross}& News & 0.369 & 0.097 & N.A & N.A & N.A & N.A & N.A & \\ \hline
\citet{bian2013multimedia} (EXT) & \citet{bian2013multimedia}& Social Media & 0.507 & 0.303 & N.A & 0.232 & N.A & N.A & N.A & \\ \hline
\citet{yan2012visualizing} (EXT) & \citet{yan2012visualizing}& News & 0.442 & 0.109 & 0.320 & N.A & N.A & N.A & N.A & \\ \hline
\multirow{2}{*}{\citet{bian2014multimedia} (EXT)} & \citet{bian2014multimedia} (social trends)& Social Media & 0.504 & 0.307 & N.A & 0.235 & N.A & N.A & N.A & \\
 & \citet{bian2014multimedia} (product events) & & 0.478 & 0.279 & N.A & 0.187 & N.A & N.A & N.A & \\ \hline
\multirow{2}{*}{\citet{fu2020multi} (EXT)} & \citet{fu2020multi} (DailyMail) & News & 0.417 & 0.186 & 0.317 & N.A & N.A & N.A & N.A & \checkmark \\
 & \citet{fu2020multi} (CNN) & & 0.278 & 0.088 & 0.187 & N.A & N.A & N.A & N.A & \checkmark \\\hline
\citet{li2020vmsmo} (ABS) & \citet{li2020vmsmo}& News & 0.251 & 0.096 & 0.232 & N.A & N.A & N.A & 0.654 & \checkmark\\\hline
\multirow{3}{*}{\citet{li2020aspect} (ABS)} & \citet{li2020aspect} (Home Appliances)& E-commerce & 0.344 & 0.125 & 0.224 & N.A & N.A & N.A & N.A & \checkmark \\
 & \citet{li2020aspect} (Clothing) & & 0.319 & 0.111 & 0.215 & N.A & N.A & N.A & N.A & \checkmark \\
 & \citet{li2020aspect} (Cases \& Bags) & & 0.338 & 0.125 & 0.224 & N.A & N.A & N.A & N.A & \checkmark \\\hline

\end{tabular}}
\end{table}

\section{Results and Discussion}
Since the MMS task is quite broad, covering multiple sub-problem statements, it is difficult to compare models due to the lack of a standard evaluation metric (refer to Section \ref{sec:data}). We are then restricted to presenting the results using uni-modal evaluation techniques like ROUGE scores \cite{lin-2004-rouge} for text summaries, and precision-recall scores for image summaries. In Section \ref{sec:seg}, we have described the diversity of works done so far, with some working on timeline generation \cite{tiwari2018multimodal,xu2013cross,sahuguet2013socially}, while others working on generic news summarization \cite{jangra2020multimodal,zhu2018msmo}; making it difficult to conduct a fair comparison of different architectures\footnote{Note that we only display the results that have text as the \textit{central modality} (refer to Section \ref{sec:mms}).}. Even comparing two models that have a very similar settings like \citet{zhu2018msmo} and  \citet{chen2018abstractive} (both are trained on large scale abstractive news summarization datasets), is not adequate because datasets \#5 and \#7 have different sizes of training data (refer to Table \ref{tab:datasets}). Other such example is of \citet{fu2020multi} and \citet{li2020vmsmo}, both these works intake text-video inputs; however \citet{fu2020multi} is trained on English dataset with ~2k instances, and \citet{li2020vmsmo} is trained on Chinese dataset with ~1,84k instances (refer to Table \ref{tab:datasets}). Nonetheless, we attempted to give the readers an overview of the potential of existing architectures. There are a few observations that can be made even with these constraints. We can observe that the abstractive summarization models go neck-to-neck with extractive summarization models, even though extractive summarization models have an advantage of keeping the basic grammatical syntax intact, illustrating the advancement in neural summarization models in the MMS task. An extensive study can be found in Table \ref{tab:results}.

There exist some works that share a common dataset to illustrate the efficacy of their proposed model architectures. For instance, \citet{zhu2018msmo} and \citet{zhu3multimodal} share a common dataset (dataset \#2). Both the works produce competitive results, with \citet{zhu3multimodal} outperforming \citet{zhu2018msmo} by small difference in all modalities. It can also be observed from the results of \citet{li2017multi} that the input language does not affect the quality of summary at all. Results for both English and Chinese datasets (refer to dataset \#3 in Table \ref{tab:datasets}) are close, and the difference can be accredited to non-overlapping content across the two datasets. We can also observe from the results by \citet{fu2020multi} that neural models require large datasets to perform better. The CNN part of dataset only comprises of 200 data instances, while the DailyMail part of dataset comprises of 1970 instances. The authors also suggest that the larger size of videos in CNN data leads to worse performance, even though the underlying learning strategies are the same. 

Some datasets are also an extension of existing ones; for instance, dataset \#19 (\cite{jangra2020text}) was extended from dataset \#21 *\cite{li2017multi}) by incorporating images and videos in the references, while dataset \#20 (\cite{jangra2021multimodal}) was extended from dataset \#19 (\cite{jangra2020text}) by introducing complementary and supplementary enhancements for the multi-modal references. Therefore all four works share the same reference summaries. Hence even though the other modalities differ, the works can be partially compared with each other for the text modality. From this, it can be deduced that the two-step approach proposed by \cite{jangra2021multimodal} that first generates the \textit{Global Coverage Text Format summary (GCTF)} using grey-wolf optimizer on a multi-objective optimization setup, and then enhances this by using other modalities outperforms all the prior works; illustrating the power of population based techniques. The submodular optimization \cite{li2017multi} is able to outperform the Genetic Algorithm technique \cite{jangra2020multimodal}, which is again a population based-technique by some margin, which we believe can be credited to both, the ability of sub-modular optimization as well as the trade-off for a the multi-modal summary generation framework. Since \citet{li2017multi} only-generates text, while \citet{jangra2020multimodal} generates a multi-modal output comprising of text, images, and videos; there might be some trade-off to improve quality of other modalities over text. Since \citet{jangra2020text} and \citet{jangra2020multimodal} both present their works on the same dataset (dataset \#19), and it is evident that the population-based genetic algorithm proposed in \citet{jangra2020multimodal} produces better summaries as compared to the single point optimization strategy using integer linear programming proposed in \citet{jangra2020text}, both in terms of text as well as image output. For the video output\footnote{Since dataset \#19 (\cite{jangra2020text}) and \#20 (\cite{jangra2021multimodal}) are the only datasets which contain text and video in the output, we have reported the results in text instead of making another column in Table \ref{tab:results}. It should also be noted that accuracy is used to evaluate the video summary, because both of these datasets restrict a single video in the output summary. Since dataset \#20 is extended from dataset \#19 they both share the same text and video outputs.} \citet{jangra2020multimodal} and \citet{jangra2020text} performed equally well with an accuracy of 44\%, while \citet{jangra2021multimodal} was able to obtain video accuracy of 64\% (in contrast to the average accuracy of 16\% for random selection over 10 attempts).

Out of the 17 works reported in Table \ref{tab:results}, 8 have performed some sort of manual evaluation along with automatic evaluation to produce a clearer picture about the performance of various summarization strategies. Through these experiments, prior works have statistically shown how the presence of multi-modal information can not only aid the uni-modal summarization process, but improve the overall user experience. \citet{li2020vmsmo} has shown that an output containing text and images increases user satisfaction by 12.4\% in juxtaposition to text summaries. \citet{jangra2021multimodal} also illustrate that having visual cues in a text summary helps improve the overall satisfaction by 22\%, makes the topic 19\% more fascinating, and helps users understand the topic better by 14.5\%. \citet{jangra2021multimodal} also empirically justify through manual annotations that a multi-modal summary should have both complementary and supplementary enhancements to improve the user experience.

\section{Future Work} \label{sec:future}
The MMS task is relatively new, and the work done so far has only scratched the surface of what this field has to offer. In this section we discuss the future scope of the MMS task, including some possible improvements in existing works, as well as some possible new directions.

\subsection{Scope of improvement} 
\noindent\textbf{Better fusion of multi-modal information:} Almost all the works discussed in this survey adopt a late-joint representation approach, where uni-modal information is extracted beforehand, and the information-sharing across multiple modalities takes place at a later stage. These works either use a pre-trained model on image captions \cite{Simonyan15} or train the multi-modal correspondence in a naïve way, using a neural multi-modal attention mechanism. However \citet{liu-etal-2020-multistage} have proposed a multi-stage fusion approach with a fusion forget gate
module for solving the task of multimodal summarization in videos. Their proposed approach tries to improve the interaction between multiple modalities to complete the missing information of each modality. Further they have also introduced a forget gate to suppress the flow of unecessary multimodal noise. Using this approach the model was able to outperform the \citet{palaskar2019multimodal} model  8.3 BLEU-4 points, 7.4 ROUGEL points and 3.9 METEOR points in the How2 \cite{sanabria2018how2} dataset. Although these techniques are able to capture the essence of semantic overlap across modalities, there is still room for improvement in fusion modeling.

\vspace{1.5mm}
\noindent\textbf{Better evaluation metrics (for multi-modal summaries):} Most of the existing works use uni-modal evaluation techniques like ROUGE scores \cite{lin-2004-rouge} for text, and precision-recall based metrics for images and videos. The multi-modal evaluation metrics proposed by \citet{zhu2018msmo} and  \citet{zhu3multimodal} have shown some promise, but they require a large set of human evaluation scores of generated summaries for training to determine the parameter values, making them unfit as a universal metric, especially when the summaries to be evaluated are from different domains than the data the models were trained on. These proposed evaluation metrics are also very specific as they work only for text-image based multi-modal summaries. Hence the community still lacks an evaluation metric that could judge the quality of a summary comprising of multiple summaries. Even the standard text summarization metrics have some inherent shortcomings, as illustrated by the survey performed by \citet{ter2020makes}. They illustrated that even though these metrics are able to cover up basic concepts like informativeness, fluency, succinctness and factuality; they still lack other important aspects like usefulness as discovered by the survey conducted on users who frequently use automatic summarization mechanisms. In order to improve the overall user satisfaction, similar techniques should be incorporated for the evaluation of multi-modal summarization systems as well.

\vspace{1.5em}
\noindent\textbf{More datasets} All the datasets proposed in the community till date are mostly centered towards the news domain, even though there are multiple potential applications in other domains like medical report summarization, tutorial summarization, simplification summarization, slogan generation etc. which could benefit from multi-modal information. There are also potential new research areas that can be explored, but due to the lack of dataset availability, the community is unable to pursue research in these fields. Some of these are: explainable MMS, sentiment lossless MMS, multi-lingual MMS, data-stream MMS, large-scale MMS of long documents etc.

\vspace{1.5mm}
\noindent\textbf{Complementary and Supplementary MMS:}
It is well established fact that multi-modal systems improve the user experience and help paint a clearer picture of topics or events discussed in input documents \cite{li2020vmsmo, jangra2021multimodal}. However, there does not exist any system that can generate the complementary and supplementary multi-modal summaries together. A large majority of research work today focus on developing supplementary multi-modal summaries \cite{zhu2018msmo, chen2018abstractive}. There also exists some works that generate complementary multi-modal summaries as well \cite{li2020vmsmo}.  \citet{jangra2021multimodal} also illustrated how an ideal multi-modal summary should comprise of both complementary and supplementary enhancements.

But the concepts of complementary and supplementary enhancements should not be limited to visual modalities over textual central modality as proposed in \citet{jangra2021multimodal}. For instance, summarizing articles with user opinion from the comments section can be a great application\footnote{This task can be considered multi-modal if we extend the notion of modality to something more generic; however, since the scope of this survey is limited to the distinction in modality being the form of representing information, we do not consider such works in great detail.}. Even though this is a text-only task, the concepts of complementary and supplementary enhancements can be extended to cover up comments that cover vivid perspectives, in both favor and against the information presented in the article. 

No abstractive complementary-supplementary MMS framework or application has been proposed in the community so far, and hence the exploration potential in this is quite vast.

\subsection{New directions}

\noindent\textbf{Manually generated dataset for evaluation of MMS evaluation metrics :} There is a need of some human annotated datasets to evaluate the performance of existing and upcoming evaluation metrics. There have been some works in text summarization that can be used to draw out some parallels; for instance, \citet{fabbri2021summeval} releases the SummEval dataset that gives out human annotation scores for 1600 article documents scored by 11 annotators in four key-characteristics of a summary - \textit{consistency}, \textit{coherence}, \textit{fluency} and \textit{relevance}. Similar work is also needed in the MMS, where, other than uni-modal aspects, the ability to judge the cross-modal information correspondence also should be taken into account.

\vspace{1.5mm}
\noindent\textbf{Explainable and Controlled MMS:} 
\citet{maynez2020faithfulness} showed that automated abstractive summarization models suffer from the problem of hallucinations and often generate fictional content. Explainable and Controlled MMS refers to the process of developing summarization systems where we do not  treat these automated systems as black boxes generating summaries; Rather we have the power to understand and control the output of these models so as to produce content of our desired type.
Even though existing MMS summarization frameworks have shown substantial improvement in the recent few years, it is still a mystery how each modality is handled and understood to obtain the final summaries. This calls for more explainable systems that also output some meta-data in tandem with the summaries to better understand the functioning of these models. Attention mechanism \cite{bahdanau2016neural} is one way to get better insights in the model working. In the context of text summarization, \citet{haonan2020exploring} proposed a select and generate strategy where elements are first extracted from a document based on informativeness, novelty, and relevance and then an abstractor generates an abstractive summary using the extracted elements. Their extractor module features an interaction matrix to explain the selection logic and by changing the thresholds of the model one can control the final summary quality.

In the multimodal context, \citet{shang2022duo} proposed DGExplain which exploits the cross-modal association between the news of multiple modalities and the user comments to detect misinformation. Explainable and controlled multimodal summarization systems can be built using this kind of explainable framework to detect and filter incorrect content and summarize the true facts. \citet{mukherjee2022topic} proposed a multi-tasking approach to generate topic-aware multimodal summaries. Their proposed model aims to embed topic awareness in both the visual and textual outputs. Thus, these kinds of models are stepping stones towards developing systems that are able to control the information flow from different modalities in input and output. 


\vspace{1.5mm}
\noindent\textbf{Application-oriented MMS:}
We can use different MMS techniques to leverage the output of various tasks like product description generation, product review summarization, multi-modal microblog summarization, education material summarization, medical report summarization and simplification of any multi-modal content.
For each of these tasks, earlier text-only \cite{chen2019towards,yu2016product,ali2020topic} or image-only \cite{somasundaram2017machine} summarization methods were majorly used. However, \citet{li2020aspect} showed that the quality of e-commerce product descriptions could be improved by incorporating visual information and textual descriptions of a product during the summarization process. \citet{delbrouck2021qiai} utilized the visual features from the x-rays
associated with the radiology reports to improve the medical report summarization quality. 

During any natural disaster, people post relevant content on microblogging websites, which concerned authorities could use for rescue operations. \citet{saini2021multimodal} proposed a multi-modal approach to summarize these posts utilizing both the textual and visual aspects of the post to improve the summary quality. Recently, educational content has been multi-modal, comprising video, audio and text. We believe that educational material summary quality can be significantly enhanced if information from all of these modalities is utilized during the summarization process \cite{khullar2020mast}. All of these recent works highlight the ability of MMS to combine the knowledge from various modalities to produce superior-quality summaries. Hence making it a more robust choice over the traditional uni-modality-based methods for multiple applications in future.

\vspace{1.5mm}
\noindent\textbf{Sentiment/Emotion Lossless MMS:} The point of a summary is to provide users with the information that they'd gain from reading the entire document; and an ideal summary would not only do that, but also elicit the same sentiments that the user would feel when reading the entire document. \citet{10.1001/jamaophthalmol.2019.2004} proposes an extractive text summarization framework that attempts to retain the sentiment of input in the generated summary. This task would be very relevant in few domains like story summarization, novel summarization etc. where the users tend to empathize with the content in the summary. \citet{Khan2021ExploitingBF} proposed a transformer based architecture to perform aspect based multimodal sentiment analysis. In future we can combine ideas from aspect based summarization systems \cite{li2020aspect} and multimodal aspect based sentiment recognition frameworks to gnerate sentiment aware MMS. When working with multi-modal data, this becomes even more challenging and interesting since various additional flavors of sentiment can be obtained from different modalities; and in some cases some modalities can fill the lack of sentiment in others. For instance, in a news article covering an earthquake, the text tends to be objective and devoid of subjective and sentiment-bearing expressions in order to remain professional, but the images and videos are able to convey these sentiments and emotions conveniently. Hence we believe that this kind of multi-modal summarization would help move current systems one step further in an attempt to obtain ideal summaries\footnote{Note that this problem would be mostly restrictive to single-document summarization tasks (with some exceptions), since multiple articles tend to cover different aspects of a topic, often leading to conflicting opinions, and hence conflicting sentiments and emotions. There the problem statement can be changed to providing an unbiased and sentiment-less summary to be faithful to the users.}.

\vspace{1.5mm}
\noindent\textbf{Multi-lingual MMS:} Multi-modal information has proven to be useful for multi-modal neural machine translation tasks \cite{specia2018multi,qian2018multimodal}, and it has been a highly debated question whether language affects visual perception, a universal form of perception, shared by all individuals \cite{vulchanova2019language}. The fact that this question remains open till this date speaks volumes about how multi-modal information can prove to be useful for multi-lingual summarization tasks, if harnessed properly.

\vspace{1.5mm}
\noindent\textbf{Data-stream MMS:} Data-stream summarization, also known as \textit{update summarization} or \textit{online summarization} or \textit{dynamic summarization} has been explored in great extent in the automatic text summarization community \cite{saggion2013automatic, gupta2016sentiment, zhan2009gather, hu2017opinion, tsai2020improving, liu2015incrests, takamura2011summarizing, shou2013sumblr}. Data-stream summarization is used in situations where the input information is not static, and accordingly the summarization system needs to dynamically keep the summary up-to-date with the latest information. It is a challenging problem as it requires the summary to retain the key-highlights from past events, while being consistent and fluent with the most recent events as well. Data stream summarization has been used for various applications like social media content summarization \cite{liu2015incrests, shou2013sumblr}, review summarization \cite{gupta2016sentiment, zhan2009gather, hu2017opinion, tsai2020improving}, etc.

With the world moving towards multi-modal information representation, there is a need to make these models robust and adaptive to multi-modal information. A few of these are discussed in the `Application-oriented MMS' part of this section. 

\vspace{1.5mm}
\noindent\textbf{Query-based MMS:} A lot of work has been done in query based text summarization \cite{rahman2019improvement,litvak2017query}, but there is no existing research on query-based summarization in a multi-modal setting. Since it has been shown that visual content can help improve the quality of experience \cite{zhu2018msmo}, we believe that query-based summarization setup, that has a user-interaction, could really be improved by introducing multi-modal form of information.

\vspace{1.5mm}
\noindent\textbf{MMS at scale:} Although some work has been done on generic datasets in terms of domain coverage \cite{jangra2020multimodal,zhu2018msmo,li2017multi,chen2018abstractive}, most of the existing works have been performed in a protective environment with some pre-defined notions of input and output formats. In order to produce a large-scale ready-to-use MMS framework, a more generic setup is required, that has better generalization and high adaptive capabilities.

\vspace{1.5mm}
\noindent\textbf{MMS with user interaction:} Inspired from query-chain summarization frameworks \cite{baumel2014query}, there is a possibility of a multi-modal summarization based on user interaction, which could help improve the overall user satisfaction.

\section{Conclusion} \label{sec:conc}
Due to the improving technology, it has become convenient for people to create and share information in multiple modalities, a feat that was not possible a decade ago. As a result of this advancement, the need for multi-modal summarization is increasing. We present a survey to help familiarize users with techniques and challenges present for the MMS task. In this manuscript, we formally define the task of multi-modal summarization, and we also provide an extensive categorization of existing works depending upon various input, output and technique related details. We then include a comprehensive description of datasets used to tackle the MMS task. Moreover, we also briefly describe various techniques used to solve the MMS task, along with the evaluation metrics used to judge the quality of summaries produced. Finally, we also provide a few possible directions that research in MMS can take. We hope that this survey paper will significantly promote research in multi-modal summarization.

\nocite{chen2010automatic} \nocite{evangelopoulos2008movie}
\nocite{evangelopoulos2013multimodal}
\nocite{zhu2020multimodal}
\nocite{kato2021multi}
\nocite{javed2022multimodal}
\nocite{li2018read}
\nocite{oskouie2014multimodal}
\nocite{zlatintsi2017cognimuse}
\nocite{khullar2020mast}
\bibliographystyle{ACM-Reference-Format}
\bibliography{main}


\begin{thebibliography}{198}


\ifx \showCODEN    \undefined \def \showCODEN     #1{\unskip}     \fi
\ifx \showDOI      \undefined \def \showDOI       #1{#1}\fi
\ifx \showISBNx    \undefined \def \showISBNx     #1{\unskip}     \fi
\ifx \showISBNxiii \undefined \def \showISBNxiii  #1{\unskip}     \fi
\ifx \showISSN     \undefined \def \showISSN      #1{\unskip}     \fi
\ifx \showLCCN     \undefined \def \showLCCN      #1{\unskip}     \fi
\ifx \shownote     \undefined \def \shownote      #1{#1}          \fi
\ifx \showarticletitle \undefined \def \showarticletitle #1{#1}   \fi
\ifx \showURL      \undefined \def \showURL       {\relax}        \fi
\providecommand\bibfield[2]{#2}
\providecommand\bibinfo[2]{#2}
\providecommand\natexlab[1]{#1}
\providecommand\showeprint[2][]{arXiv:#2}

\bibitem[\protect\citeauthoryear{Alguliev, Aliguliyev, and
  Hajirahimova}{Alguliev et~al\mbox{.}}{2010}]%
        {alguliev2010multi}
\bibfield{author}{\bibinfo{person}{Rasim Alguliev}, \bibinfo{person}{Ramiz
  Aliguliyev}, {and} \bibinfo{person}{Makrufa Hajirahimova}.}
  \bibinfo{year}{2010}\natexlab{}.
\newblock \showarticletitle{Multi-document summarization model based on integer
  linear programming}.
\newblock \bibinfo{journal}{\emph{Intelligent Control and Automation}}
  \bibinfo{volume}{1}, \bibinfo{number}{02} (\bibinfo{year}{2010}),
  \bibinfo{pages}{105}.
\newblock


\bibitem[\protect\citeauthoryear{Ali, Noorian, Bagheri, Ding, and
  Al-Obeidat}{Ali et~al\mbox{.}}{2020}]%
        {ali2020topic}
\bibfield{author}{\bibinfo{person}{Syed~Muhammad Ali}, \bibinfo{person}{Zeinab
  Noorian}, \bibinfo{person}{Ebrahim Bagheri}, \bibinfo{person}{Chen Ding},
  {and} \bibinfo{person}{Feras Al-Obeidat}.} \bibinfo{year}{2020}\natexlab{}.
\newblock \showarticletitle{Topic and sentiment aware microblog summarization
  for twitter}.
\newblock \bibinfo{journal}{\emph{Journal of Intelligent Information Systems}}
  \bibinfo{volume}{54}, \bibinfo{number}{1} (\bibinfo{year}{2020}),
  \bibinfo{pages}{129--156}.
\newblock


\bibitem[\protect\citeauthoryear{Arshad, Gallo, Nawaz, and Calefati}{Arshad
  et~al\mbox{.}}{2019}]%
        {arshad2019aiding}
\bibfield{author}{\bibinfo{person}{Omer Arshad}, \bibinfo{person}{Ignazio
  Gallo}, \bibinfo{person}{Shah Nawaz}, {and} \bibinfo{person}{Alessandro
  Calefati}.} \bibinfo{year}{2019}\natexlab{}.
\newblock \showarticletitle{Aiding intra-text representations with visual
  context for multimodal named entity recognition}. In
  \bibinfo{booktitle}{\emph{2019 International Conference on Document Analysis
  and Recognition (ICDAR)}}. IEEE, \bibinfo{pages}{337--342}.
\newblock


\bibitem[\protect\citeauthoryear{Atrey, Hossain, El~Saddik, and
  Kankanhalli}{Atrey et~al\mbox{.}}{2010}]%
        {atrey2010multimodal}
\bibfield{author}{\bibinfo{person}{Pradeep~K Atrey}, \bibinfo{person}{M~Anwar
  Hossain}, \bibinfo{person}{Abdulmotaleb El~Saddik}, {and}
  \bibinfo{person}{Mohan~S Kankanhalli}.} \bibinfo{year}{2010}\natexlab{}.
\newblock \showarticletitle{Multimodal fusion for multimedia analysis: a
  survey}.
\newblock \bibinfo{journal}{\emph{Multimedia systems}} \bibinfo{volume}{16},
  \bibinfo{number}{6} (\bibinfo{year}{2010}), \bibinfo{pages}{345--379}.
\newblock


\bibitem[\protect\citeauthoryear{Bahdanau, Cho, and Bengio}{Bahdanau
  et~al\mbox{.}}{2016}]%
        {bahdanau2016neural}
\bibfield{author}{\bibinfo{person}{Dzmitry Bahdanau},
  \bibinfo{person}{Kyunghyun Cho}, {and} \bibinfo{person}{Yoshua Bengio}.}
  \bibinfo{year}{2016}\natexlab{}.
\newblock \bibinfo{title}{Neural Machine Translation by Jointly Learning to
  Align and Translate}.
\newblock
\newblock
\showeprint[arxiv]{1409.0473}~[cs.CL]


\bibitem[\protect\citeauthoryear{Baltru{\v{s}}aitis, Ahuja, and
  Morency}{Baltru{\v{s}}aitis et~al\mbox{.}}{2018}]%
        {baltruvsaitis2018multimodal}
\bibfield{author}{\bibinfo{person}{Tadas Baltru{\v{s}}aitis},
  \bibinfo{person}{Chaitanya Ahuja}, {and} \bibinfo{person}{Louis-Philippe
  Morency}.} \bibinfo{year}{2018}\natexlab{}.
\newblock \showarticletitle{Multimodal machine learning: A survey and
  taxonomy}.
\newblock \bibinfo{journal}{\emph{IEEE transactions on pattern analysis and
  machine intelligence}} \bibinfo{volume}{41}, \bibinfo{number}{2}
  (\bibinfo{year}{2018}), \bibinfo{pages}{423--443}.
\newblock


\bibitem[\protect\citeauthoryear{Barbieri, Ballesteros, Ronzano, and
  Saggion}{Barbieri et~al\mbox{.}}{2018}]%
        {barbieri2018multimodal}
\bibfield{author}{\bibinfo{person}{Francesco Barbieri}, \bibinfo{person}{Miguel
  Ballesteros}, \bibinfo{person}{Francesco Ronzano}, {and}
  \bibinfo{person}{Horacio Saggion}.} \bibinfo{year}{2018}\natexlab{}.
\newblock \showarticletitle{Multimodal emoji prediction}.
\newblock \bibinfo{journal}{\emph{arXiv preprint arXiv:1803.02392}}
  (\bibinfo{year}{2018}).
\newblock


\bibitem[\protect\citeauthoryear{Basavarajaiah and Sharma}{Basavarajaiah and
  Sharma}{2019}]%
        {basavarajaiah2019survey}
\bibfield{author}{\bibinfo{person}{Madhushree Basavarajaiah} {and}
  \bibinfo{person}{Priyanka Sharma}.} \bibinfo{year}{2019}\natexlab{}.
\newblock \showarticletitle{Survey of Compressed Domain Video Summarization
  Techniques}.
\newblock \bibinfo{journal}{\emph{ACM Computing Surveys (CSUR)}}
  \bibinfo{volume}{52}, \bibinfo{number}{6} (\bibinfo{year}{2019}),
  \bibinfo{pages}{1--29}.
\newblock


\bibitem[\protect\citeauthoryear{Baumel, Cohen, and Elhadad}{Baumel
  et~al\mbox{.}}{2014}]%
        {baumel2014query}
\bibfield{author}{\bibinfo{person}{Tal Baumel}, \bibinfo{person}{Raphael
  Cohen}, {and} \bibinfo{person}{Michael Elhadad}.}
  \bibinfo{year}{2014}\natexlab{}.
\newblock \showarticletitle{Query-chain focused summarization}. In
  \bibinfo{booktitle}{\emph{Proceedings of the 52nd Annual Meeting of the
  Association for Computational Linguistics (Volume 1: Long Papers)}}.
  \bibinfo{pages}{913--922}.
\newblock


\bibitem[\protect\citeauthoryear{Bian, Yang, and Chua}{Bian
  et~al\mbox{.}}{2013}]%
        {bian2013multimedia}
\bibfield{author}{\bibinfo{person}{Jingwen Bian}, \bibinfo{person}{Yang Yang},
  {and} \bibinfo{person}{Tat-Seng Chua}.} \bibinfo{year}{2013}\natexlab{}.
\newblock \showarticletitle{Multimedia summarization for trending topics in
  microblogs}. In \bibinfo{booktitle}{\emph{Proceedings of the 22nd ACM
  international conference on Information \& Knowledge Management}}.
  \bibinfo{pages}{1807--1812}.
\newblock


\bibitem[\protect\citeauthoryear{Bian, Yang, Zhang, and Chua}{Bian
  et~al\mbox{.}}{2014}]%
        {bian2014multimedia}
\bibfield{author}{\bibinfo{person}{Jingwen Bian}, \bibinfo{person}{Yang Yang},
  \bibinfo{person}{Hanwang Zhang}, {and} \bibinfo{person}{Tat-Seng Chua}.}
  \bibinfo{year}{2014}\natexlab{}.
\newblock \showarticletitle{Multimedia summarization for social events in
  microblog stream}.
\newblock \bibinfo{journal}{\emph{IEEE Transactions on multimedia}}
  \bibinfo{volume}{17}, \bibinfo{number}{2} (\bibinfo{year}{2014}),
  \bibinfo{pages}{216--228}.
\newblock


\bibitem[\protect\citeauthoryear{Blei, Ng, and Jordan}{Blei
  et~al\mbox{.}}{2003}]%
        {blei2003latent}
\bibfield{author}{\bibinfo{person}{David~M Blei}, \bibinfo{person}{Andrew~Y
  Ng}, {and} \bibinfo{person}{Michael~I Jordan}.}
  \bibinfo{year}{2003}\natexlab{}.
\newblock \showarticletitle{Latent dirichlet allocation}.
\newblock \bibinfo{journal}{\emph{Journal of machine Learning research}}
  \bibinfo{volume}{3}, \bibinfo{number}{Jan} (\bibinfo{year}{2003}),
  \bibinfo{pages}{993--1022}.
\newblock


\bibitem[\protect\citeauthoryear{Botschen, Gurevych, Klie, Mousselly-Sergieh,
  and Roth}{Botschen et~al\mbox{.}}{2018}]%
        {botschen2018multimodal}
\bibfield{author}{\bibinfo{person}{Teresa Botschen}, \bibinfo{person}{Iryna
  Gurevych}, \bibinfo{person}{Jan-Christoph Klie}, \bibinfo{person}{Hatem
  Mousselly-Sergieh}, {and} \bibinfo{person}{Stefan Roth}.}
  \bibinfo{year}{2018}\natexlab{}.
\newblock \showarticletitle{Multimodal frame identification with multilingual
  evaluation}. In \bibinfo{booktitle}{\emph{Proceedings of the 2018 Conference
  of the North American Chapter of the Association for Computational
  Linguistics: Human Language Technologies, Volume 1 (Long Papers)}}.
  \bibinfo{pages}{1481--1491}.
\newblock


\bibitem[\protect\citeauthoryear{Caglayan, Madhyastha, Specia, and
  Barrault}{Caglayan et~al\mbox{.}}{2019}]%
        {caglayan2019probing}
\bibfield{author}{\bibinfo{person}{Ozan Caglayan}, \bibinfo{person}{Pranava
  Madhyastha}, \bibinfo{person}{Lucia Specia}, {and} \bibinfo{person}{Lo{\"\i}c
  Barrault}.} \bibinfo{year}{2019}\natexlab{}.
\newblock \showarticletitle{Probing the need for visual context in multimodal
  machine translation}.
\newblock \bibinfo{journal}{\emph{arXiv preprint arXiv:1903.08678}}
  (\bibinfo{year}{2019}).
\newblock


\bibitem[\protect\citeauthoryear{Chen, De~Vleeschouwer, Barrob{\'e}s, Escalada,
  and Conejero}{Chen et~al\mbox{.}}{2010}]%
        {chen2010automatic}
\bibfield{author}{\bibinfo{person}{Fan Chen}, \bibinfo{person}{Christophe
  De~Vleeschouwer}, \bibinfo{person}{H~Duxans Barrob{\'e}s},
  \bibinfo{person}{J~Gregorio Escalada}, {and} \bibinfo{person}{David
  Conejero}.} \bibinfo{year}{2010}\natexlab{}.
\newblock \showarticletitle{Automatic summarization of audio-visual soccer
  feeds}. In \bibinfo{booktitle}{\emph{2010 IEEE International Conference on
  Multimedia and Expo}}. IEEE, \bibinfo{pages}{837--842}.
\newblock


\bibitem[\protect\citeauthoryear{Chen and Zhuge}{Chen and Zhuge}{2018a}]%
        {chen2018abstractive}
\bibfield{author}{\bibinfo{person}{Jingqiang Chen} {and} \bibinfo{person}{Hai
  Zhuge}.} \bibinfo{year}{2018}\natexlab{a}.
\newblock \showarticletitle{Abstractive Text-Image Summarization Using
  Multi-Modal Attentional Hierarchical RNN}. In
  \bibinfo{booktitle}{\emph{Proceedings of the 2018 Conference on Empirical
  Methods in Natural Language Processing}}. \bibinfo{pages}{4046--4056}.
\newblock


\bibitem[\protect\citeauthoryear{Chen and Zhuge}{Chen and Zhuge}{2018b}]%
        {chen2018extractive}
\bibfield{author}{\bibinfo{person}{Jingqiang Chen} {and} \bibinfo{person}{Hai
  Zhuge}.} \bibinfo{year}{2018}\natexlab{b}.
\newblock \showarticletitle{Extractive Text-Image Summarization Using
  Multi-Modal RNN}. In \bibinfo{booktitle}{\emph{2018 14th International
  Conference on Semantics, Knowledge and Grids (SKG)}}. IEEE,
  \bibinfo{pages}{245--248}.
\newblock


\bibitem[\protect\citeauthoryear{Chen and Zhuge}{Chen and Zhuge}{2019}]%
        {chen2019news}
\bibfield{author}{\bibinfo{person}{Jingqiang Chen} {and} \bibinfo{person}{Hai
  Zhuge}.} \bibinfo{year}{2019}\natexlab{}.
\newblock \showarticletitle{News Image Captioning Based on Text Summarization
  Using Image as Query}. In \bibinfo{booktitle}{\emph{2019 15th International
  Conference on Semantics, Knowledge and Grids (SKG)}}. IEEE,
  \bibinfo{pages}{123--126}.
\newblock


\bibitem[\protect\citeauthoryear{Chen and Zhuge}{Chen and Zhuge}{2020}]%
        {chen2020news}
\bibfield{author}{\bibinfo{person}{Jingqiang Chen} {and} \bibinfo{person}{Hai
  Zhuge}.} \bibinfo{year}{2020}\natexlab{}.
\newblock \showarticletitle{A news image captioning approach based on
  multimodal pointer-generator network}.
\newblock \bibinfo{journal}{\emph{Concurrency and Computation: Practice and
  Experience}} (\bibinfo{year}{2020}), \bibinfo{pages}{e5721}.
\newblock


\bibitem[\protect\citeauthoryear{Chen, Lin, Zhang, Yang, Zhou, and Tang}{Chen
  et~al\mbox{.}}{2019}]%
        {chen2019towards}
\bibfield{author}{\bibinfo{person}{Qibin Chen}, \bibinfo{person}{Junyang Lin},
  \bibinfo{person}{Yichang Zhang}, \bibinfo{person}{Hongxia Yang},
  \bibinfo{person}{Jingren Zhou}, {and} \bibinfo{person}{Jie Tang}.}
  \bibinfo{year}{2019}\natexlab{}.
\newblock \showarticletitle{Towards knowledge-based personalized product
  description generation in e-commerce}. In
  \bibinfo{booktitle}{\emph{Proceedings of the 25th ACM SIGKDD International
  Conference on Knowledge Discovery \& Data Mining}}.
  \bibinfo{pages}{3040--3050}.
\newblock


\bibitem[\protect\citeauthoryear{Chen and Bansal}{Chen and Bansal}{2018}]%
        {chen2018fast}
\bibfield{author}{\bibinfo{person}{Yen-Chun Chen} {and} \bibinfo{person}{Mohit
  Bansal}.} \bibinfo{year}{2018}\natexlab{}.
\newblock \showarticletitle{Fast Abstractive Summarization with
  Reinforce-Selected Sentence Rewriting}. In
  \bibinfo{booktitle}{\emph{Proceedings of the 56th Annual Meeting of the
  Association for Computational Linguistics (Volume 1: Long Papers)}}.
  \bibinfo{pages}{675--686}.
\newblock


\bibitem[\protect\citeauthoryear{Chen, Li, Yu, Kholy, Ahmed, Gan, Cheng, and
  Liu}{Chen et~al\mbox{.}}{2020}]%
        {chen2020uniter}
\bibfield{author}{\bibinfo{person}{Yen-Chun Chen}, \bibinfo{person}{Linjie Li},
  \bibinfo{person}{Licheng Yu}, \bibinfo{person}{Ahmed~El Kholy},
  \bibinfo{person}{Faisal Ahmed}, \bibinfo{person}{Zhe Gan},
  \bibinfo{person}{Yu Cheng}, {and} \bibinfo{person}{Jingjing Liu}.}
  \bibinfo{year}{2020}\natexlab{}.
\newblock \bibinfo{title}{UNITER: UNiversal Image-TExt Representation
  Learning}.
\newblock
\newblock
\showeprint[arxiv]{1909.11740}~[cs.CV]


\bibitem[\protect\citeauthoryear{Cho, van Merri{\"e}nboer, Gulcehre, Bahdanau,
  Bougares, Schwenk, and Bengio}{Cho et~al\mbox{.}}{2014}]%
        {cho2014learning}
\bibfield{author}{\bibinfo{person}{Kyunghyun Cho}, \bibinfo{person}{Bart van
  Merri{\"e}nboer}, \bibinfo{person}{Caglar Gulcehre}, \bibinfo{person}{Dzmitry
  Bahdanau}, \bibinfo{person}{Fethi Bougares}, \bibinfo{person}{Holger
  Schwenk}, {and} \bibinfo{person}{Yoshua Bengio}.}
  \bibinfo{year}{2014}\natexlab{}.
\newblock \showarticletitle{Learning Phrase Representations using RNN
  Encoder--Decoder for Statistical Machine Translation}. In
  \bibinfo{booktitle}{\emph{Proceedings of the 2014 Conference on Empirical
  Methods in Natural Language Processing (EMNLP)}}.
  \bibinfo{pages}{1724--1734}.
\newblock


\bibitem[\protect\citeauthoryear{Clark, Celikyilmaz, and Smith}{Clark
  et~al\mbox{.}}{2019}]%
        {clark-etal-2019-sentence}
\bibfield{author}{\bibinfo{person}{Elizabeth Clark}, \bibinfo{person}{Asli
  Celikyilmaz}, {and} \bibinfo{person}{Noah~A. Smith}.}
  \bibinfo{year}{2019}\natexlab{}.
\newblock \showarticletitle{Sentence Mover{'}s Similarity: Automatic Evaluation
  for Multi-Sentence Texts}. In \bibinfo{booktitle}{\emph{Proceedings of the
  57th Annual Meeting of the Association for Computational Linguistics}}.
  \bibinfo{publisher}{Association for Computational Linguistics},
  \bibinfo{address}{Florence, Italy}.
\newblock
\urldef\tempurl%
\url{https://doi.org/10.18653/v1/P19-1264}
\showDOI{\tempurl}


\bibitem[\protect\citeauthoryear{Coman, Nechaev, and Zara}{Coman
  et~al\mbox{.}}{2018}]%
        {coman2018predicting}
\bibfield{author}{\bibinfo{person}{Andrei~Catalin Coman},
  \bibinfo{person}{Yaroslav Nechaev}, {and} \bibinfo{person}{Giacomo Zara}.}
  \bibinfo{year}{2018}\natexlab{}.
\newblock \showarticletitle{Predicting emoji exploiting multimodal data: FBK
  participation in ITAmoji task}.
\newblock \bibinfo{journal}{\emph{EVALITA Evaluation of NLP and Speech Tools
  for Italian}}  \bibinfo{volume}{12} (\bibinfo{year}{2018}),
  \bibinfo{pages}{135}.
\newblock


\bibitem[\protect\citeauthoryear{Cui, Radosavljevic, Chou, Lin, Nguyen, Huang,
  Schneider, and Djuric}{Cui et~al\mbox{.}}{2019}]%
        {cui2019multimodal}
\bibfield{author}{\bibinfo{person}{Henggang Cui}, \bibinfo{person}{Vladan
  Radosavljevic}, \bibinfo{person}{Fang-Chieh Chou}, \bibinfo{person}{Tsung-Han
  Lin}, \bibinfo{person}{Thi Nguyen}, \bibinfo{person}{Tzu-Kuo Huang},
  \bibinfo{person}{Jeff Schneider}, {and} \bibinfo{person}{Nemanja Djuric}.}
  \bibinfo{year}{2019}\natexlab{}.
\newblock \showarticletitle{Multimodal trajectory predictions for autonomous
  driving using deep convolutional networks}. In \bibinfo{booktitle}{\emph{2019
  International Conference on Robotics and Automation (ICRA)}}. IEEE,
  \bibinfo{pages}{2090--2096}.
\newblock


\bibitem[\protect\citeauthoryear{Delbrouck, Zhang, and Rubin}{Delbrouck
  et~al\mbox{.}}{2021}]%
        {delbrouck2021qiai}
\bibfield{author}{\bibinfo{person}{Jean-Benoit Delbrouck},
  \bibinfo{person}{Cassie Zhang}, {and} \bibinfo{person}{Daniel Rubin}.}
  \bibinfo{year}{2021}\natexlab{}.
\newblock \showarticletitle{QIAI at MEDIQA 2021: Multimodal Radiology Report
  Summarization}. In \bibinfo{booktitle}{\emph{Proceedings of the 20th Workshop
  on Biomedical Language Processing}}. \bibinfo{pages}{285--290}.
\newblock


\bibitem[\protect\citeauthoryear{Deng, Dong, Socher, Li, Li, and Fei-Fei}{Deng
  et~al\mbox{.}}{2009}]%
        {deng2009imagenet}
\bibfield{author}{\bibinfo{person}{Jia Deng}, \bibinfo{person}{Wei Dong},
  \bibinfo{person}{Richard Socher}, \bibinfo{person}{Li-Jia Li},
  \bibinfo{person}{Kai Li}, {and} \bibinfo{person}{Li Fei-Fei}.}
  \bibinfo{year}{2009}\natexlab{}.
\newblock \showarticletitle{Imagenet: A large-scale hierarchical image
  database}. In \bibinfo{booktitle}{\emph{2009 IEEE conference on computer
  vision and pattern recognition}}. Ieee, \bibinfo{pages}{248--255}.
\newblock


\bibitem[\protect\citeauthoryear{Devlin, Chang, Lee, and Toutanova}{Devlin
  et~al\mbox{.}}{2019}]%
        {devlin2018bert}
\bibfield{author}{\bibinfo{person}{Jacob Devlin}, \bibinfo{person}{Ming-Wei
  Chang}, \bibinfo{person}{Kenton Lee}, {and} \bibinfo{person}{Kristina
  Toutanova}.} \bibinfo{year}{2019}\natexlab{}.
\newblock \showarticletitle{{BERT}: Pre-training of Deep Bidirectional
  Transformers for Language Understanding}. In
  \bibinfo{booktitle}{\emph{Proceedings of the 2019 Conference of the North
  {A}merican Chapter of the Association for Computational Linguistics: Human
  Language Technologies, Volume 1 (Long and Short Papers)}}.
  \bibinfo{publisher}{Association for Computational Linguistics},
  \bibinfo{address}{Minneapolis, Minnesota}, \bibinfo{pages}{4171--4186}.
\newblock
\urldef\tempurl%
\url{https://doi.org/10.18653/v1/N19-1423}
\showDOI{\tempurl}


\bibitem[\protect\citeauthoryear{Dosovitskiy, Beyer, Kolesnikov, Weissenborn,
  Zhai, Unterthiner, Dehghani, Minderer, Heigold, Gelly,
  et~al\mbox{.}}{Dosovitskiy et~al\mbox{.}}{2020}]%
        {dosovitskiy2020image}
\bibfield{author}{\bibinfo{person}{Alexey Dosovitskiy}, \bibinfo{person}{Lucas
  Beyer}, \bibinfo{person}{Alexander Kolesnikov}, \bibinfo{person}{Dirk
  Weissenborn}, \bibinfo{person}{Xiaohua Zhai}, \bibinfo{person}{Thomas
  Unterthiner}, \bibinfo{person}{Mostafa Dehghani}, \bibinfo{person}{Matthias
  Minderer}, \bibinfo{person}{Georg Heigold}, \bibinfo{person}{Sylvain Gelly},
  {et~al\mbox{.}}} \bibinfo{year}{2020}\natexlab{}.
\newblock \showarticletitle{An Image is Worth 16x16 Words: Transformers for
  Image Recognition at Scale}. In \bibinfo{booktitle}{\emph{International
  Conference on Learning Representations}}.
\newblock


\bibitem[\protect\citeauthoryear{Elliott}{Elliott}{2018}]%
        {elliott2018adversarial}
\bibfield{author}{\bibinfo{person}{Desmond Elliott}.}
  \bibinfo{year}{2018}\natexlab{}.
\newblock \showarticletitle{Adversarial evaluation of multimodal machine
  translation}. In \bibinfo{booktitle}{\emph{Proceedings of the 2018 Conference
  on Empirical Methods in Natural Language Processing}}.
  \bibinfo{pages}{2974--2978}.
\newblock


\bibitem[\protect\citeauthoryear{Elliott, Frank, Barrault, Bougares, and
  Specia}{Elliott et~al\mbox{.}}{2017}]%
        {elliott2017findings}
\bibfield{author}{\bibinfo{person}{Desmond Elliott}, \bibinfo{person}{Stella
  Frank}, \bibinfo{person}{Lo{\"\i}c Barrault}, \bibinfo{person}{Fethi
  Bougares}, {and} \bibinfo{person}{Lucia Specia}.}
  \bibinfo{year}{2017}\natexlab{}.
\newblock \showarticletitle{Findings of the second shared task on multimodal
  machine translation and multilingual image description}.
\newblock \bibinfo{journal}{\emph{arXiv preprint arXiv:1710.07177}}
  (\bibinfo{year}{2017}).
\newblock


\bibitem[\protect\citeauthoryear{Erkan and Radev}{Erkan and Radev}{2004}]%
        {erkan2004lexrank}
\bibfield{author}{\bibinfo{person}{G{\"u}nes Erkan} {and}
  \bibinfo{person}{Dragomir~R Radev}.} \bibinfo{year}{2004}\natexlab{}.
\newblock \showarticletitle{Lexrank: Graph-based lexical centrality as salience
  in text summarization}.
\newblock \bibinfo{journal}{\emph{Jour. of artif. intel. res.}}
  \bibinfo{volume}{22} (\bibinfo{year}{2004}), \bibinfo{pages}{457--479}.
\newblock


\bibitem[\protect\citeauthoryear{Ermakova, Cossu, and Mothe}{Ermakova
  et~al\mbox{.}}{2019}]%
        {ermakova2019survey}
\bibfield{author}{\bibinfo{person}{Liana Ermakova},
  \bibinfo{person}{Jean~Val{\`e}re Cossu}, {and} \bibinfo{person}{Josiane
  Mothe}.} \bibinfo{year}{2019}\natexlab{}.
\newblock \showarticletitle{A survey on evaluation of summarization methods}.
\newblock \bibinfo{journal}{\emph{Information Processing \& Management}}
  \bibinfo{volume}{56}, \bibinfo{number}{5} (\bibinfo{year}{2019}),
  \bibinfo{pages}{1794--1814}.
\newblock


\bibitem[\protect\citeauthoryear{Erol, Lee, and Hull}{Erol
  et~al\mbox{.}}{2003}]%
        {erol2003multimodal}
\bibfield{author}{\bibinfo{person}{Berna Erol}, \bibinfo{person}{D-S Lee},
  {and} \bibinfo{person}{Jonathan Hull}.} \bibinfo{year}{2003}\natexlab{}.
\newblock \showarticletitle{Multimodal summarization of meeting recordings}. In
  \bibinfo{booktitle}{\emph{2003 International Conference on Multimedia and
  Expo. ICME'03. Proceedings (Cat. No. 03TH8698)}}, Vol.~\bibinfo{volume}{3}.
  IEEE, \bibinfo{pages}{III--25}.
\newblock


\bibitem[\protect\citeauthoryear{Eskandar, Sadollah, Bahreininejad, and
  Hamdi}{Eskandar et~al\mbox{.}}{2012}]%
        {eskandar2012water}
\bibfield{author}{\bibinfo{person}{Hadi Eskandar}, \bibinfo{person}{Ali
  Sadollah}, \bibinfo{person}{Ardeshir Bahreininejad}, {and}
  \bibinfo{person}{Mohd Hamdi}.} \bibinfo{year}{2012}\natexlab{}.
\newblock \showarticletitle{Water cycle algorithm--A novel metaheuristic
  optimization method for solving constrained engineering optimization
  problems}.
\newblock \bibinfo{journal}{\emph{Computers \& Structures}}
  \bibinfo{volume}{110} (\bibinfo{year}{2012}), \bibinfo{pages}{151--166}.
\newblock


\bibitem[\protect\citeauthoryear{Evangelopoulos, Rapantzikos, Potamianos,
  Maragos, Zlatintsi, and Avrithis}{Evangelopoulos et~al\mbox{.}}{2008}]%
        {evangelopoulos2008movie}
\bibfield{author}{\bibinfo{person}{Georgios Evangelopoulos},
  \bibinfo{person}{Konstantinos Rapantzikos}, \bibinfo{person}{Alexandros
  Potamianos}, \bibinfo{person}{Petros Maragos}, \bibinfo{person}{A Zlatintsi},
  {and} \bibinfo{person}{Yannis Avrithis}.} \bibinfo{year}{2008}\natexlab{}.
\newblock \showarticletitle{Movie summarization based on audiovisual saliency
  detection}. In \bibinfo{booktitle}{\emph{2008 15th IEEE International
  Conference on Image Processing}}. IEEE, \bibinfo{pages}{2528--2531}.
\newblock


\bibitem[\protect\citeauthoryear{Evangelopoulos, Zlatintsi, Potamianos,
  Maragos, Rapantzikos, Skoumas, and Avrithis}{Evangelopoulos
  et~al\mbox{.}}{2013}]%
        {evangelopoulos2013multimodal}
\bibfield{author}{\bibinfo{person}{Georgios Evangelopoulos},
  \bibinfo{person}{Athanasia Zlatintsi}, \bibinfo{person}{Alexandros
  Potamianos}, \bibinfo{person}{Petros Maragos}, \bibinfo{person}{Konstantinos
  Rapantzikos}, \bibinfo{person}{Georgios Skoumas}, {and}
  \bibinfo{person}{Yannis Avrithis}.} \bibinfo{year}{2013}\natexlab{}.
\newblock \showarticletitle{Multimodal saliency and fusion for movie
  summarization based on aural, visual, and textual attention}.
\newblock \bibinfo{journal}{\emph{IEEE Transactions on Multimedia}}
  \bibinfo{volume}{15}, \bibinfo{number}{7} (\bibinfo{year}{2013}),
  \bibinfo{pages}{1553--1568}.
\newblock


\bibitem[\protect\citeauthoryear{Evangelopoulos, Zlatintsi, Skoumas,
  Rapantzikos, Potamianos, Maragos, and Avrithis}{Evangelopoulos
  et~al\mbox{.}}{2009}]%
        {evangelopoulos2009video}
\bibfield{author}{\bibinfo{person}{Georgios Evangelopoulos},
  \bibinfo{person}{Athanasia Zlatintsi}, \bibinfo{person}{Georgios Skoumas},
  \bibinfo{person}{Konstantinos Rapantzikos}, \bibinfo{person}{Alexandros
  Potamianos}, \bibinfo{person}{Petros Maragos}, {and} \bibinfo{person}{Yannis
  Avrithis}.} \bibinfo{year}{2009}\natexlab{}.
\newblock \showarticletitle{Video event detection and summarization using
  audio, visual and text saliency}. In \bibinfo{booktitle}{\emph{2009 IEEE
  International Conference on Acoustics, Speech and Signal Processing}}. IEEE,
  \bibinfo{pages}{3553--3556}.
\newblock


\bibitem[\protect\citeauthoryear{Fabbri, Kryscinski, McCann, Socher, and
  Radev}{Fabbri et~al\mbox{.}}{2021}]%
        {fabbri2021summeval}
\bibfield{author}{\bibinfo{person}{A.~R. Fabbri}, \bibinfo{person}{Wojciech
  Kryscinski}, \bibinfo{person}{Bryan McCann}, \bibinfo{person}{R. Socher},
  {and} \bibinfo{person}{Dragomir Radev}.} \bibinfo{year}{2021}\natexlab{}.
\newblock \showarticletitle{SummEval: Re-evaluating Summarization Evaluation}.
\newblock \bibinfo{journal}{\emph{Transactions of the Association for
  Computational Linguistics}}  \bibinfo{volume}{9} (\bibinfo{year}{2021}),
  \bibinfo{pages}{391--409}.
\newblock


\bibitem[\protect\citeauthoryear{Feng, Yang, Cer, Arivazhagan, and Wang}{Feng
  et~al\mbox{.}}{2020}]%
        {feng2020languageagnostic}
\bibfield{author}{\bibinfo{person}{Fangxiaoyu Feng}, \bibinfo{person}{Yinfei
  Yang}, \bibinfo{person}{Daniel Cer}, \bibinfo{person}{Naveen Arivazhagan},
  {and} \bibinfo{person}{Wei Wang}.} \bibinfo{year}{2020}\natexlab{}.
\newblock \bibinfo{title}{Language-agnostic BERT Sentence Embedding}.
\newblock
\newblock
\showeprint[arxiv]{2007.01852}~[cs.CL]


\bibitem[\protect\citeauthoryear{Fierrez-Aguilar, Ortega-Garcia,
  Gonzalez-Rodriguez, and Bigun}{Fierrez-Aguilar et~al\mbox{.}}{2005}]%
        {fierrez2005discriminative}
\bibfield{author}{\bibinfo{person}{Julian Fierrez-Aguilar},
  \bibinfo{person}{Javier Ortega-Garcia}, \bibinfo{person}{Joaquin
  Gonzalez-Rodriguez}, {and} \bibinfo{person}{Josef Bigun}.}
  \bibinfo{year}{2005}\natexlab{}.
\newblock \showarticletitle{Discriminative multimodal biometric authentication
  based on quality measures}.
\newblock \bibinfo{journal}{\emph{Pattern recognition}} \bibinfo{volume}{38},
  \bibinfo{number}{5} (\bibinfo{year}{2005}), \bibinfo{pages}{777--779}.
\newblock


\bibitem[\protect\citeauthoryear{Fu, Wang, and Yang}{Fu et~al\mbox{.}}{2020}]%
        {fu2020multi}
\bibfield{author}{\bibinfo{person}{Xiyan Fu}, \bibinfo{person}{Jun Wang}, {and}
  \bibinfo{person}{Zhenglu Yang}.} \bibinfo{year}{2020}\natexlab{}.
\newblock \showarticletitle{Multi-modal Summarization for Video-containing
  Documents}.
\newblock \bibinfo{journal}{\emph{arXiv preprint arXiv:2009.08018}}
  (\bibinfo{year}{2020}).
\newblock


\bibitem[\protect\citeauthoryear{Galanis, Lampouras, and
  Androutsopoulos}{Galanis et~al\mbox{.}}{2012}]%
        {galanis2012extractive}
\bibfield{author}{\bibinfo{person}{Dimitrios Galanis},
  \bibinfo{person}{Gerasimos Lampouras}, {and} \bibinfo{person}{Ion
  Androutsopoulos}.} \bibinfo{year}{2012}\natexlab{}.
\newblock \showarticletitle{Extractive multi-document summarization with
  integer linear programming and support vector regression}. In
  \bibinfo{booktitle}{\emph{Proceedings of COLING 2012}}.
  \bibinfo{pages}{911--926}.
\newblock


\bibitem[\protect\citeauthoryear{Gambhir and Gupta}{Gambhir and Gupta}{2017}]%
        {gambhir2017recent}
\bibfield{author}{\bibinfo{person}{Mahak Gambhir} {and} \bibinfo{person}{Vishal
  Gupta}.} \bibinfo{year}{2017}\natexlab{}.
\newblock \showarticletitle{Recent automatic text summarization techniques: a
  survey}.
\newblock \bibinfo{journal}{\emph{Artificial Intelligence Review}}
  \bibinfo{volume}{47}, \bibinfo{number}{1} (\bibinfo{year}{2017}),
  \bibinfo{pages}{1--66}.
\newblock


\bibitem[\protect\citeauthoryear{Gao, Zhao, and Eger}{Gao
  et~al\mbox{.}}{2020}]%
        {gao2020supert}
\bibfield{author}{\bibinfo{person}{Yang Gao}, \bibinfo{person}{Wei Zhao}, {and}
  \bibinfo{person}{Steffen Eger}.} \bibinfo{year}{2020}\natexlab{}.
\newblock \showarticletitle{{SUPERT}: Towards New Frontiers in Unsupervised
  Evaluation Metrics for Multi-Document Summarization}. In
  \bibinfo{booktitle}{\emph{Proceedings of the 58th Annual Meeting of the
  Association for Computational Linguistics}}. \bibinfo{publisher}{Association
  for Computational Linguistics}, \bibinfo{address}{Online}.
\newblock
\urldef\tempurl%
\url{https://doi.org/10.18653/v1/2020.acl-main.124}
\showDOI{\tempurl}


\bibitem[\protect\citeauthoryear{Gulshan, Rajan, Widner, Wu, Wubbels, Rhodes,
  Whitehouse, Coram, Corrado, Ramasamy, Raman, Peng, and Webster}{Gulshan
  et~al\mbox{.}}{2019}]%
        {10.1001/jamaophthalmol.2019.2004}
\bibfield{author}{\bibinfo{person}{Varun Gulshan}, \bibinfo{person}{Renu~P.
  Rajan}, \bibinfo{person}{Kasumi Widner}, \bibinfo{person}{Derek Wu},
  \bibinfo{person}{Peter Wubbels}, \bibinfo{person}{Tyler Rhodes},
  \bibinfo{person}{Kira Whitehouse}, \bibinfo{person}{Marc Coram},
  \bibinfo{person}{Greg Corrado}, \bibinfo{person}{Kim Ramasamy},
  \bibinfo{person}{Rajiv Raman}, \bibinfo{person}{Lily Peng}, {and}
  \bibinfo{person}{Dale~R. Webster}.} \bibinfo{year}{2019}\natexlab{}.
\newblock \showarticletitle{{Performance of a Deep-Learning Algorithm vs Manual
  Grading for Detecting Diabetic Retinopathy in India}}.
\newblock \bibinfo{journal}{\emph{JAMA Ophthalmology}} \bibinfo{volume}{137},
  \bibinfo{number}{9} (\bibinfo{date}{09} \bibinfo{year}{2019}),
  \bibinfo{pages}{987--993}.
\newblock
\showISSN{2168-6165}
\urldef\tempurl%
\url{https://doi.org/10.1001/jamaophthalmol.2019.2004}
\showDOI{\tempurl}


\bibitem[\protect\citeauthoryear{Gupta, Tiwari, and Robert}{Gupta
  et~al\mbox{.}}{2016}]%
        {gupta2016sentiment}
\bibfield{author}{\bibinfo{person}{Pankaj Gupta}, \bibinfo{person}{Ritu
  Tiwari}, {and} \bibinfo{person}{Nirmal Robert}.}
  \bibinfo{year}{2016}\natexlab{}.
\newblock \showarticletitle{Sentiment analysis and text summarization of online
  reviews: A survey}. In \bibinfo{booktitle}{\emph{2016 International
  Conference on Communication and Signal Processing (ICCSP)}}. IEEE,
  \bibinfo{pages}{0241--0245}.
\newblock


\bibitem[\protect\citeauthoryear{Gupta and Lehal}{Gupta and Lehal}{2010}]%
        {gupta2010survey}
\bibfield{author}{\bibinfo{person}{Vishal Gupta} {and}
  \bibinfo{person}{Gurpreet~Singh Lehal}.} \bibinfo{year}{2010}\natexlab{}.
\newblock \showarticletitle{A survey of text summarization extractive
  techniques}.
\newblock \bibinfo{journal}{\emph{Journal of emerging technologies in web
  intelligence}} \bibinfo{volume}{2}, \bibinfo{number}{3}
  (\bibinfo{year}{2010}), \bibinfo{pages}{258--268}.
\newblock


\bibitem[\protect\citeauthoryear{Haonan, Yang, Yu, Lapata, and Heyan}{Haonan
  et~al\mbox{.}}{2020}]%
        {haonan2020exploring}
\bibfield{author}{\bibinfo{person}{Wang Haonan}, \bibinfo{person}{Gao Yang},
  \bibinfo{person}{Bai Yu}, \bibinfo{person}{Mirella Lapata}, {and}
  \bibinfo{person}{Huang Heyan}.} \bibinfo{year}{2020}\natexlab{}.
\newblock \showarticletitle{Exploring Explainable Selection to Control
  Abstractive Summarization}.
\newblock \bibinfo{journal}{\emph{arXiv preprint arXiv:2004.11779}}
  (\bibinfo{year}{2020}).
\newblock


\bibitem[\protect\citeauthoryear{Hara, Kataoka, and Satoh}{Hara
  et~al\mbox{.}}{2018}]%
        {hara2018can}
\bibfield{author}{\bibinfo{person}{Kensho Hara}, \bibinfo{person}{Hirokatsu
  Kataoka}, {and} \bibinfo{person}{Yutaka Satoh}.}
  \bibinfo{year}{2018}\natexlab{}.
\newblock \showarticletitle{Can spatiotemporal 3d cnns retrace the history of
  2d cnns and imagenet?}. In \bibinfo{booktitle}{\emph{Proceedings of the IEEE
  conference on Computer Vision and Pattern Recognition}}.
  \bibinfo{pages}{6546--6555}.
\newblock


\bibitem[\protect\citeauthoryear{He, Zhang, Ren, and Sun}{He
  et~al\mbox{.}}{2016}]%
        {he2016deep}
\bibfield{author}{\bibinfo{person}{Kaiming He}, \bibinfo{person}{Xiangyu
  Zhang}, \bibinfo{person}{Shaoqing Ren}, {and} \bibinfo{person}{Jian Sun}.}
  \bibinfo{year}{2016}\natexlab{}.
\newblock \showarticletitle{Deep residual learning for image recognition}. In
  \bibinfo{booktitle}{\emph{Proceedings of the IEEE conference on computer
  vision and pattern recognition}}. \bibinfo{pages}{770--778}.
\newblock


\bibitem[\protect\citeauthoryear{Hochreiter and Schmidhuber}{Hochreiter and
  Schmidhuber}{1997}]%
        {hochreiter1997long}
\bibfield{author}{\bibinfo{person}{Sepp Hochreiter} {and}
  \bibinfo{person}{J{\"u}rgen Schmidhuber}.} \bibinfo{year}{1997}\natexlab{}.
\newblock \showarticletitle{Long short-term memory}.
\newblock \bibinfo{journal}{\emph{Neural computation}} \bibinfo{volume}{9},
  \bibinfo{number}{8} (\bibinfo{year}{1997}), \bibinfo{pages}{1735--1780}.
\newblock


\bibitem[\protect\citeauthoryear{Hodosh, Young, and Hockenmaier}{Hodosh
  et~al\mbox{.}}{2013}]%
        {hodosh2013framing}
\bibfield{author}{\bibinfo{person}{Micah Hodosh}, \bibinfo{person}{Peter
  Young}, {and} \bibinfo{person}{Julia Hockenmaier}.}
  \bibinfo{year}{2013}\natexlab{}.
\newblock \showarticletitle{Framing image description as a ranking task: Data,
  models and evaluation metrics}.
\newblock \bibinfo{journal}{\emph{Journal of Artificial Intelligence Research}}
   \bibinfo{volume}{47} (\bibinfo{year}{2013}), \bibinfo{pages}{853--899}.
\newblock


\bibitem[\protect\citeauthoryear{Hori, Hori, Lee, Zhang, Harsham, Hershey,
  Marks, and Sumi}{Hori et~al\mbox{.}}{2017}]%
        {hori2017attention}
\bibfield{author}{\bibinfo{person}{Chiori Hori}, \bibinfo{person}{Takaaki
  Hori}, \bibinfo{person}{Teng-Yok Lee}, \bibinfo{person}{Ziming Zhang},
  \bibinfo{person}{Bret Harsham}, \bibinfo{person}{John~R Hershey},
  \bibinfo{person}{Tim~K Marks}, {and} \bibinfo{person}{Kazuhiko Sumi}.}
  \bibinfo{year}{2017}\natexlab{}.
\newblock \showarticletitle{Attention-based multimodal fusion for video
  description}. In \bibinfo{booktitle}{\emph{Proceedings of the IEEE
  international conference on computer vision}}. \bibinfo{pages}{4193--4202}.
\newblock


\bibitem[\protect\citeauthoryear{Hori, Hori, Wichern, Wang, Lee, Cherian, and
  Marks}{Hori et~al\mbox{.}}{2018}]%
        {hori2018multimodal}
\bibfield{author}{\bibinfo{person}{Chiori Hori}, \bibinfo{person}{Takaaki
  Hori}, \bibinfo{person}{Gordon Wichern}, \bibinfo{person}{Jue Wang},
  \bibinfo{person}{Teng-yok Lee}, \bibinfo{person}{Anoop Cherian}, {and}
  \bibinfo{person}{Tim~K Marks}.} \bibinfo{year}{2018}\natexlab{}.
\newblock \showarticletitle{Multimodal Attention for Fusion of Audio and
  Spatiotemporal Features for Video Description.}. In
  \bibinfo{booktitle}{\emph{CVPR Workshops}}. \bibinfo{pages}{2528--2531}.
\newblock


\bibitem[\protect\citeauthoryear{Hu, Chen, and Chou}{Hu et~al\mbox{.}}{2017}]%
        {hu2017opinion}
\bibfield{author}{\bibinfo{person}{Ya-Han Hu}, \bibinfo{person}{Yen-Liang
  Chen}, {and} \bibinfo{person}{Hui-Ling Chou}.}
  \bibinfo{year}{2017}\natexlab{}.
\newblock \showarticletitle{Opinion mining from online hotel reviews--a text
  summarization approach}.
\newblock \bibinfo{journal}{\emph{Information Processing \& Management}}
  \bibinfo{volume}{53}, \bibinfo{number}{2} (\bibinfo{year}{2017}),
  \bibinfo{pages}{436--449}.
\newblock


\bibitem[\protect\citeauthoryear{Huang, Liu, Shiang, Oh, and Dyer}{Huang
  et~al\mbox{.}}{2016}]%
        {huang2016attention}
\bibfield{author}{\bibinfo{person}{Po-Yao Huang}, \bibinfo{person}{Frederick
  Liu}, \bibinfo{person}{Sz-Rung Shiang}, \bibinfo{person}{Jean Oh}, {and}
  \bibinfo{person}{Chris Dyer}.} \bibinfo{year}{2016}\natexlab{}.
\newblock \showarticletitle{Attention-based multimodal neural machine
  translation}. In \bibinfo{booktitle}{\emph{Proceedings of the First
  Conference on Machine Translation: Volume 2, Shared Task Papers}}.
  \bibinfo{pages}{639--645}.
\newblock


\bibitem[\protect\citeauthoryear{Huang, Liu, Belongie, and Kautz}{Huang
  et~al\mbox{.}}{2018}]%
        {huang2018multimodal}
\bibfield{author}{\bibinfo{person}{Xun Huang}, \bibinfo{person}{Ming-Yu Liu},
  \bibinfo{person}{Serge Belongie}, {and} \bibinfo{person}{Jan Kautz}.}
  \bibinfo{year}{2018}\natexlab{}.
\newblock \showarticletitle{Multimodal unsupervised image-to-image
  translation}. In \bibinfo{booktitle}{\emph{Proceedings of the European
  Conference on Computer Vision (ECCV)}}. \bibinfo{pages}{172--189}.
\newblock


\bibitem[\protect\citeauthoryear{Huang, Zeng, Liu, Fu, and Fu}{Huang
  et~al\mbox{.}}{2020}]%
        {huang2020pixel}
\bibfield{author}{\bibinfo{person}{Zhicheng Huang}, \bibinfo{person}{Zhaoyang
  Zeng}, \bibinfo{person}{Bei Liu}, \bibinfo{person}{Dongmei Fu}, {and}
  \bibinfo{person}{Jianlong Fu}.} \bibinfo{year}{2020}\natexlab{}.
\newblock \showarticletitle{Pixel-bert: Aligning image pixels with text by deep
  multi-modal transformers}.
\newblock \bibinfo{journal}{\emph{arXiv preprint arXiv:2004.00849}}
  (\bibinfo{year}{2020}).
\newblock


\bibitem[\protect\citeauthoryear{Hussain, Muhammad, Ding, Lloret, Baik, and
  de~Albuquerque}{Hussain et~al\mbox{.}}{2020}]%
        {hussain2020comprehensive}
\bibfield{author}{\bibinfo{person}{Tanveer Hussain}, \bibinfo{person}{Khan
  Muhammad}, \bibinfo{person}{Weiping Ding}, \bibinfo{person}{Jaime Lloret},
  \bibinfo{person}{Sung~Wook Baik}, {and} \bibinfo{person}{Victor Hugo~C de
  Albuquerque}.} \bibinfo{year}{2020}\natexlab{}.
\newblock \showarticletitle{A comprehensive survey of multi-view video
  summarization}.
\newblock \bibinfo{journal}{\emph{Pattern Recognition}}  \bibinfo{volume}{109}
  (\bibinfo{year}{2020}), \bibinfo{pages}{107567}.
\newblock


\bibitem[\protect\citeauthoryear{Indovina, Uludag, Snelick, Mink, and
  Jain}{Indovina et~al\mbox{.}}{2003}]%
        {indovina2003multimodal}
\bibfield{author}{\bibinfo{person}{M Indovina}, \bibinfo{person}{U Uludag},
  \bibinfo{person}{R Snelick}, \bibinfo{person}{A Mink}, {and}
  \bibinfo{person}{A Jain}.} \bibinfo{year}{2003}\natexlab{}.
\newblock \showarticletitle{Multimodal biometric authentication methods: a COTS
  approach}. In \bibinfo{booktitle}{\emph{Proc. of Workshop on Multimodal User
  Authentication}}. Citeseer, \bibinfo{pages}{99--106}.
\newblock


\bibitem[\protect\citeauthoryear{Jaimes and Sebe}{Jaimes and Sebe}{2007}]%
        {jaimes2007multimodal}
\bibfield{author}{\bibinfo{person}{Alejandro Jaimes} {and}
  \bibinfo{person}{Nicu Sebe}.} \bibinfo{year}{2007}\natexlab{}.
\newblock \showarticletitle{Multimodal human--computer interaction: A survey}.
\newblock \bibinfo{journal}{\emph{Computer vision and image understanding}}
  \bibinfo{volume}{108}, \bibinfo{number}{1-2} (\bibinfo{year}{2007}),
  \bibinfo{pages}{116--134}.
\newblock


\bibitem[\protect\citeauthoryear{Jain, Jangra, Saha, and Jatowt}{Jain
  et~al\mbox{.}}{2022a}]%
        {jain2022survey}
\bibfield{author}{\bibinfo{person}{Raghav Jain}, \bibinfo{person}{Anubhav
  Jangra}, \bibinfo{person}{Sriparna Saha}, {and} \bibinfo{person}{Adam
  Jatowt}.} \bibinfo{year}{2022}\natexlab{a}.
\newblock \showarticletitle{A Survey on Medical Document Summarization}.
\newblock \bibinfo{journal}{\emph{arXiv preprint arXiv:2212.01669}}
  (\bibinfo{year}{2022}).
\newblock


\bibitem[\protect\citeauthoryear{Jain, Mavi, Jangra, and Saha}{Jain
  et~al\mbox{.}}{2022b}]%
        {10.1007/978-3-030-99736-6_21}
\bibfield{author}{\bibinfo{person}{Raghav Jain}, \bibinfo{person}{Vaibhav
  Mavi}, \bibinfo{person}{Anubhav Jangra}, {and} \bibinfo{person}{Sriparna
  Saha}.} \bibinfo{year}{2022}\natexlab{b}.
\newblock \showarticletitle{WIDAR - Weighted Input Document Augmented ROUGE}.
  In \bibinfo{booktitle}{\emph{Advances in Information Retrieval: 44th European
  Conference on IR Research, ECIR 2022, Stavanger, Norway, April 10–14, 2022,
  Proceedings, Part I}} (Stavanger, Norway).
  \bibinfo{publisher}{Springer-Verlag}, \bibinfo{address}{Berlin, Heidelberg},
  \bibinfo{pages}{304–321}.
\newblock
\showISBNx{978-3-030-99735-9}
\urldef\tempurl%
\url{https://doi.org/10.1007/978-3-030-99736-6_21}
\showDOI{\tempurl}


\bibitem[\protect\citeauthoryear{Jangra, Jain, Mavi, Saha, and
  Bhattacharyya}{Jangra et~al\mbox{.}}{2020a}]%
        {jangra2020semantic}
\bibfield{author}{\bibinfo{person}{Anubhav Jangra}, \bibinfo{person}{Raghav
  Jain}, \bibinfo{person}{Vaibhav Mavi}, \bibinfo{person}{Sriparna Saha}, {and}
  \bibinfo{person}{Pushpak Bhattacharyya}.} \bibinfo{year}{2020}\natexlab{a}.
\newblock \showarticletitle{Semantic Extractor-Paraphraser based Abstractive
  Summarization}. In \bibinfo{booktitle}{\emph{Proceedings of the 17th
  International Conference on Natural Language Processing (ICON)}}.
  \bibinfo{pages}{191--199}.
\newblock


\bibitem[\protect\citeauthoryear{Jangra, Jatowt, Hasanuzzaman, and Saha}{Jangra
  et~al\mbox{.}}{2020b}]%
        {jangra2020text}
\bibfield{author}{\bibinfo{person}{Anubhav Jangra}, \bibinfo{person}{Adam
  Jatowt}, \bibinfo{person}{Mohammad Hasanuzzaman}, {and}
  \bibinfo{person}{Sriparna Saha}.} \bibinfo{year}{2020}\natexlab{b}.
\newblock \showarticletitle{Text-Image-Video Summary Generation Using Joint
  Integer Linear Programming}. In \bibinfo{booktitle}{\emph{European Conference
  on Information Retrieval}}. Springer, \bibinfo{pages}{190--198}.
\newblock


\bibitem[\protect\citeauthoryear{Jangra, Saha, Jatowt, and Hasanuzzaman}{Jangra
  et~al\mbox{.}}{2020c}]%
        {jangra2020multimodal}
\bibfield{author}{\bibinfo{person}{Anubhav Jangra}, \bibinfo{person}{Sriparna
  Saha}, \bibinfo{person}{Adam Jatowt}, {and} \bibinfo{person}{Mohammad
  Hasanuzzaman}.} \bibinfo{year}{2020}\natexlab{c}.
\newblock \showarticletitle{Multi-Modal Summary Generation Using
  Multi-Objective Optimization} \emph{(\bibinfo{series}{SIGIR '20})}.
  \bibinfo{publisher}{Association for Computing Machinery},
  \bibinfo{address}{New York, NY, USA}, \bibinfo{pages}{1745–1748}.
\newblock
\showISBNx{9781450380164}
\urldef\tempurl%
\url{https://doi.org/10.1145/3397271.3401232}
\showDOI{\tempurl}


\bibitem[\protect\citeauthoryear{Jangra, Saha, Jatowt, and Hasanuzzaman}{Jangra
  et~al\mbox{.}}{2021}]%
        {jangra2021multimodal}
\bibfield{author}{\bibinfo{person}{Anubhav Jangra}, \bibinfo{person}{Sriparna
  Saha}, \bibinfo{person}{Adam Jatowt}, {and} \bibinfo{person}{Mohammed
  Hasanuzzaman}.} \bibinfo{year}{2021}\natexlab{}.
\newblock \bibinfo{booktitle}{\emph{Multi-Modal Supplementary-Complementary
  Summarization Using Multi-Objective Optimization}}.
\newblock \bibinfo{publisher}{Association for Computing Machinery},
  \bibinfo{address}{New York, NY, USA}, \bibinfo{pages}{818–828}.
\newblock
\showISBNx{9781450380379}
\urldef\tempurl%
\url{https://doi.org/10.1145/3404835.3462877}
\showURL{%
\tempurl}


\bibitem[\protect\citeauthoryear{Javed, Sufyan~Beg, and Akhtar}{Javed
  et~al\mbox{.}}{2022}]%
        {javed2022multimodal}
\bibfield{author}{\bibinfo{person}{Hira Javed}, \bibinfo{person}{MM
  Sufyan~Beg}, {and} \bibinfo{person}{Nadeem Akhtar}.}
  \bibinfo{year}{2022}\natexlab{}.
\newblock \showarticletitle{Multimodal Summarization: A Concise Review}. In
  \bibinfo{booktitle}{\emph{Proceedings of the International Conference on
  Computational Intelligence and Sustainable Technologies}}. Springer,
  \bibinfo{pages}{613--623}.
\newblock


\bibitem[\protect\citeauthoryear{Jha, Dias, Lechervy, Moreno, Jangra, Pais, and
  Saha}{Jha et~al\mbox{.}}{2022}]%
        {jha2022combining}
\bibfield{author}{\bibinfo{person}{Prince Jha}, \bibinfo{person}{Ga{\"e}l
  Dias}, \bibinfo{person}{Alexis Lechervy}, \bibinfo{person}{Jose~G Moreno},
  \bibinfo{person}{Anubhav Jangra}, \bibinfo{person}{Sebasti{\~a}o Pais}, {and}
  \bibinfo{person}{Sriparna Saha}.} \bibinfo{year}{2022}\natexlab{}.
\newblock \showarticletitle{Combining Vision and Language Representations for
  Patch-based Identification of Lexico-Semantic Relations}. In
  \bibinfo{booktitle}{\emph{Proceedings of the 30th ACM International
  Conference on Multimedia}}. \bibinfo{pages}{4406--4415}.
\newblock


\bibitem[\protect\citeauthoryear{Karpathy, Joulin, and Fei-Fei}{Karpathy
  et~al\mbox{.}}{2014}]%
        {karpathy2014deep}
\bibfield{author}{\bibinfo{person}{Andrej Karpathy}, \bibinfo{person}{Armand
  Joulin}, {and} \bibinfo{person}{Li~F Fei-Fei}.}
  \bibinfo{year}{2014}\natexlab{}.
\newblock \showarticletitle{Deep fragment embeddings for bidirectional image
  sentence mapping}. In \bibinfo{booktitle}{\emph{Advances in neural
  information processing systems}}. \bibinfo{pages}{1889--1897}.
\newblock


\bibitem[\protect\citeauthoryear{Kato}{Kato}{2021}]%
        {kato2021multi}
\bibfield{author}{\bibinfo{person}{Tsuneaki Kato}.}
  \bibinfo{year}{2021}\natexlab{}.
\newblock \showarticletitle{Multi-modal Summarization}.
\newblock In \bibinfo{booktitle}{\emph{Evaluating Information Retrieval and
  Access Tasks}}. \bibinfo{publisher}{Springer, Singapore},
  \bibinfo{pages}{71--82}.
\newblock


\bibitem[\protect\citeauthoryear{Kay, Carreira, Simonyan, Zhang, Hillier,
  Vijayanarasimhan, Viola, Green, Back, Natsev, et~al\mbox{.}}{Kay
  et~al\mbox{.}}{2017}]%
        {kay2017kinetics}
\bibfield{author}{\bibinfo{person}{Will Kay}, \bibinfo{person}{Joao Carreira},
  \bibinfo{person}{Karen Simonyan}, \bibinfo{person}{Brian Zhang},
  \bibinfo{person}{Chloe Hillier}, \bibinfo{person}{Sudheendra
  Vijayanarasimhan}, \bibinfo{person}{Fabio Viola}, \bibinfo{person}{Tim
  Green}, \bibinfo{person}{Trevor Back}, \bibinfo{person}{Paul Natsev},
  {et~al\mbox{.}}} \bibinfo{year}{2017}\natexlab{}.
\newblock \showarticletitle{The kinetics human action video dataset}.
\newblock \bibinfo{journal}{\emph{arXiv preprint arXiv:1705.06950}}
  (\bibinfo{year}{2017}).
\newblock


\bibitem[\protect\citeauthoryear{Khan, Naseer, Hayat, Zamir, Khan, and
  Shah}{Khan et~al\mbox{.}}{2021}]%
        {khan2021transformers}
\bibfield{author}{\bibinfo{person}{Salman Khan}, \bibinfo{person}{Muzammal
  Naseer}, \bibinfo{person}{Munawar Hayat}, \bibinfo{person}{Syed~Waqas Zamir},
  \bibinfo{person}{Fahad~Shahbaz Khan}, {and} \bibinfo{person}{Mubarak Shah}.}
  \bibinfo{year}{2021}\natexlab{}.
\newblock \showarticletitle{Transformers in vision: A survey}.
\newblock \bibinfo{journal}{\emph{arXiv preprint arXiv:2101.01169}}
  (\bibinfo{year}{2021}).
\newblock


\bibitem[\protect\citeauthoryear{Khan and Fu}{Khan and Fu}{2021}]%
        {Khan2021ExploitingBF}
\bibfield{author}{\bibinfo{person}{Zaid Khan} {and}
  \bibinfo{person}{Yun~Raymond Fu}.} \bibinfo{year}{2021}\natexlab{}.
\newblock \showarticletitle{Exploiting BERT for Multimodal Target Sentiment
  Classification through Input Space Translation}.
\newblock \bibinfo{journal}{\emph{Proceedings of the 29th ACM International
  Conference on Multimedia}} (\bibinfo{year}{2021}).
\newblock


\bibitem[\protect\citeauthoryear{Khullar and Arora}{Khullar and Arora}{2020}]%
        {khullar2020mast}
\bibfield{author}{\bibinfo{person}{Aman Khullar} {and} \bibinfo{person}{Udit
  Arora}.} \bibinfo{year}{2020}\natexlab{}.
\newblock \showarticletitle{MAST: Multimodal abstractive summarization with
  trimodal hierarchical attention}.
\newblock \bibinfo{journal}{\emph{arXiv preprint arXiv:2010.08021}}
  (\bibinfo{year}{2020}).
\newblock


\bibitem[\protect\citeauthoryear{Kim, Lee, Kwak, Heo, Kim, Ha, and Zhang}{Kim
  et~al\mbox{.}}{2016a}]%
        {kim2016multimodal}
\bibfield{author}{\bibinfo{person}{Jin-Hwa Kim}, \bibinfo{person}{Sang-Woo
  Lee}, \bibinfo{person}{Donghyun Kwak}, \bibinfo{person}{Min-Oh Heo},
  \bibinfo{person}{Jeonghee Kim}, \bibinfo{person}{Jung-Woo Ha}, {and}
  \bibinfo{person}{Byoung-Tak Zhang}.} \bibinfo{year}{2016}\natexlab{a}.
\newblock \showarticletitle{Multimodal residual learning for visual qa}.
\newblock \bibinfo{journal}{\emph{Advances in neural information processing
  systems}}  \bibinfo{volume}{29} (\bibinfo{year}{2016}),
  \bibinfo{pages}{361--369}.
\newblock


\bibitem[\protect\citeauthoryear{Kim, On, Lim, Kim, Ha, and Zhang}{Kim
  et~al\mbox{.}}{2016b}]%
        {kim2016hadamard}
\bibfield{author}{\bibinfo{person}{Jin-Hwa Kim}, \bibinfo{person}{Kyoung-Woon
  On}, \bibinfo{person}{Woosang Lim}, \bibinfo{person}{Jeonghee Kim},
  \bibinfo{person}{Jung-Woo Ha}, {and} \bibinfo{person}{Byoung-Tak Zhang}.}
  \bibinfo{year}{2016}\natexlab{b}.
\newblock \showarticletitle{Hadamard product for low-rank bilinear pooling}.
\newblock \bibinfo{journal}{\emph{arXiv preprint arXiv:1610.04325}}
  (\bibinfo{year}{2016}).
\newblock


\bibitem[\protect\citeauthoryear{Kim, Salter, DeLuccia, Son, Amer, and
  Tamrakar}{Kim et~al\mbox{.}}{2018}]%
        {kim2018smilee}
\bibfield{author}{\bibinfo{person}{Sujeong Kim}, \bibinfo{person}{David
  Salter}, \bibinfo{person}{Luke DeLuccia}, \bibinfo{person}{Kilho Son},
  \bibinfo{person}{Mohamed~R Amer}, {and} \bibinfo{person}{Amir Tamrakar}.}
  \bibinfo{year}{2018}\natexlab{}.
\newblock \showarticletitle{SMILEE: Symmetric multi-modal interactions with
  language-gesture enabled (AI) embodiment}. In
  \bibinfo{booktitle}{\emph{Proceedings of the 2018 Conference of the North
  American Chapter of the Association for Computational Linguistics:
  Demonstrations}}. \bibinfo{pages}{86--90}.
\newblock


\bibitem[\protect\citeauthoryear{Kini and Pai}{Kini and Pai}{2019}]%
        {kini2019survey}
\bibfield{author}{\bibinfo{person}{Mahesh Kini} {and} \bibinfo{person}{Karthik
  Pai}.} \bibinfo{year}{2019}\natexlab{}.
\newblock \showarticletitle{A Survey on Video Summarization Techniques}. In
  \bibinfo{booktitle}{\emph{2019 Innovations in Power and Advanced Computing
  Technologies (i-PACT)}}, Vol.~\bibinfo{volume}{1}. IEEE,
  \bibinfo{pages}{1--5}.
\newblock


\bibitem[\protect\citeauthoryear{Kirchner, Tabie, and Seeland}{Kirchner
  et~al\mbox{.}}{2014}]%
        {kirchner2014multimodal}
\bibfield{author}{\bibinfo{person}{Elsa~Andrea Kirchner}, \bibinfo{person}{Marc
  Tabie}, {and} \bibinfo{person}{Anett Seeland}.}
  \bibinfo{year}{2014}\natexlab{}.
\newblock \showarticletitle{Multimodal movement prediction-towards an
  individual assistance of patients}.
\newblock \bibinfo{journal}{\emph{PloS one}} \bibinfo{volume}{9},
  \bibinfo{number}{1} (\bibinfo{year}{2014}), \bibinfo{pages}{e85060}.
\newblock


\bibitem[\protect\citeauthoryear{Klein, Lev, Sadeh, and Wolf}{Klein
  et~al\mbox{.}}{2014}]%
        {klein2014fisher}
\bibfield{author}{\bibinfo{person}{Benjamin Klein}, \bibinfo{person}{Guy Lev},
  \bibinfo{person}{Gil Sadeh}, {and} \bibinfo{person}{Lior Wolf}.}
  \bibinfo{year}{2014}\natexlab{}.
\newblock \showarticletitle{Fisher vectors derived from hybrid
  gaussian-laplacian mixture models for image annotation}.
\newblock \bibinfo{journal}{\emph{arXiv preprint arXiv:1411.7399}}
  (\bibinfo{year}{2014}).
\newblock


\bibitem[\protect\citeauthoryear{Kostoulas, Chanel, Muszynski, Lombardo, and
  Pun}{Kostoulas et~al\mbox{.}}{2017}]%
        {kostoulas2017films}
\bibfield{author}{\bibinfo{person}{Theodoros Kostoulas},
  \bibinfo{person}{Guillaume Chanel}, \bibinfo{person}{Michal Muszynski},
  \bibinfo{person}{Patrizia Lombardo}, {and} \bibinfo{person}{Thierry Pun}.}
  \bibinfo{year}{2017}\natexlab{}.
\newblock \showarticletitle{Films, affective computing and aesthetic
  experience: Identifying emotional and aesthetic highlights from multimodal
  signals in a social setting}.
\newblock \bibinfo{journal}{\emph{Frontiers in ICT}}  \bibinfo{volume}{4}
  (\bibinfo{year}{2017}), \bibinfo{pages}{11}.
\newblock


\bibitem[\protect\citeauthoryear{Leviathan and Matias}{Leviathan and
  Matias}{2018}]%
        {leviathan2018google}
\bibfield{author}{\bibinfo{person}{Yaniv Leviathan} {and}
  \bibinfo{person}{Yossi Matias}.} \bibinfo{year}{2018}\natexlab{}.
\newblock \showarticletitle{Google Duplex: An AI system for accomplishing
  real-world tasks over the phone}.
\newblock  (\bibinfo{year}{2018}).
\newblock


\bibitem[\protect\citeauthoryear{Li, Yuan, Xu, Wu, He, and Zhou}{Li
  et~al\mbox{.}}{2020c}]%
        {li2020aspect}
\bibfield{author}{\bibinfo{person}{Haoran Li}, \bibinfo{person}{Peng Yuan},
  \bibinfo{person}{Song Xu}, \bibinfo{person}{Youzheng Wu},
  \bibinfo{person}{Xiaodong He}, {and} \bibinfo{person}{Bowen Zhou}.}
  \bibinfo{year}{2020}\natexlab{c}.
\newblock \showarticletitle{Aspect-Aware Multimodal Summarization for Chinese
  E-Commerce Products.}. In \bibinfo{booktitle}{\emph{AAAI}}.
  \bibinfo{pages}{8188--8195}.
\newblock


\bibitem[\protect\citeauthoryear{Li, Zhu, Liu, Zhang, and Zong}{Li
  et~al\mbox{.}}{2018a}]%
        {li2018multi}
\bibfield{author}{\bibinfo{person}{Haoran Li}, \bibinfo{person}{Junnan Zhu},
  \bibinfo{person}{Tianshang Liu}, \bibinfo{person}{Jiajun Zhang}, {and}
  \bibinfo{person}{Chengqing Zong}.} \bibinfo{year}{2018}\natexlab{a}.
\newblock \showarticletitle{Multi-modal Sentence Summarization with Modality
  Attention and Image Filtering}. In \bibinfo{booktitle}{\emph{Proceedings of
  the Twenty-Seventh International Joint Conference on Artificial Intelligence,
  {IJCAI-18}}}. \bibinfo{publisher}{International Joint Conferences on
  Artificial Intelligence Organization}, \bibinfo{pages}{4152--4158}.
\newblock
\urldef\tempurl%
\url{https://doi.org/10.24963/ijcai.2018/577}
\showDOI{\tempurl}


\bibitem[\protect\citeauthoryear{Li, Zhu, Ma, Zhang, and Zong}{Li
  et~al\mbox{.}}{2017}]%
        {li2017multi}
\bibfield{author}{\bibinfo{person}{Haoran Li}, \bibinfo{person}{Junnan Zhu},
  \bibinfo{person}{Cong Ma}, \bibinfo{person}{Jiajun Zhang}, {and}
  \bibinfo{person}{Chengqing Zong}.} \bibinfo{year}{2017}\natexlab{}.
\newblock \showarticletitle{Multi-modal summarization for asynchronous
  collection of text, image, audio and video}. In
  \bibinfo{booktitle}{\emph{Proceedings of the 2017 Conference on Empirical
  Methods in Natural Language Processing}}. \bibinfo{pages}{1092--1102}.
\newblock


\bibitem[\protect\citeauthoryear{Li, Zhu, Ma, Zhang, and Zong}{Li
  et~al\mbox{.}}{2018b}]%
        {li2018read}
\bibfield{author}{\bibinfo{person}{Haoran Li}, \bibinfo{person}{Junnan Zhu},
  \bibinfo{person}{Cong Ma}, \bibinfo{person}{Jiajun Zhang}, {and}
  \bibinfo{person}{Chengqing Zong}.} \bibinfo{year}{2018}\natexlab{b}.
\newblock \showarticletitle{Read, watch, listen, and summarize: Multi-modal
  summarization for asynchronous text, image, audio and video}.
\newblock \bibinfo{journal}{\emph{IEEE Transactions on Knowledge and Data
  Engineering}} \bibinfo{volume}{31}, \bibinfo{number}{5}
  (\bibinfo{year}{2018}), \bibinfo{pages}{996--1009}.
\newblock


\bibitem[\protect\citeauthoryear{Li, Yang, Smyth, and Dong}{Li
  et~al\mbox{.}}{2020b}]%
        {li2020maec}
\bibfield{author}{\bibinfo{person}{Jiazheng Li}, \bibinfo{person}{Linyi Yang},
  \bibinfo{person}{Barry Smyth}, {and} \bibinfo{person}{Ruihai Dong}.}
  \bibinfo{year}{2020}\natexlab{b}.
\newblock \showarticletitle{MAEC: A Multimodal Aligned Earnings Conference Call
  Dataset for Financial Risk Prediction}. In
  \bibinfo{booktitle}{\emph{Proceedings of the 29th ACM International
  Conference on Information \& Knowledge Management}}.
  \bibinfo{pages}{3063--3070}.
\newblock


\bibitem[\protect\citeauthoryear{Li, Yatskar, Yin, Hsieh, and Chang}{Li
  et~al\mbox{.}}{2019}]%
        {li2019visualbert}
\bibfield{author}{\bibinfo{person}{Liunian~Harold Li}, \bibinfo{person}{Mark
  Yatskar}, \bibinfo{person}{Da Yin}, \bibinfo{person}{Cho-Jui Hsieh}, {and}
  \bibinfo{person}{Kai-Wei Chang}.} \bibinfo{year}{2019}\natexlab{}.
\newblock \showarticletitle{Visualbert: A simple and performant baseline for
  vision and language}.
\newblock \bibinfo{journal}{\emph{arXiv preprint arXiv:1908.03557}}
  (\bibinfo{year}{2019}).
\newblock


\bibitem[\protect\citeauthoryear{Li, Chen, Gao, Chan, Zhao, and Yan}{Li
  et~al\mbox{.}}{2020a}]%
        {li2020vmsmo}
\bibfield{author}{\bibinfo{person}{Mingzhe Li}, \bibinfo{person}{Xiuying Chen},
  \bibinfo{person}{Shen Gao}, \bibinfo{person}{Zhangming Chan},
  \bibinfo{person}{Dongyan Zhao}, {and} \bibinfo{person}{Rui Yan}.}
  \bibinfo{year}{2020}\natexlab{a}.
\newblock \showarticletitle{VMSMO: Learning to Generate Multimodal Summary for
  Video-based News Articles}.
\newblock \bibinfo{journal}{\emph{arXiv preprint arXiv:2010.05406}}
  (\bibinfo{year}{2020}).
\newblock


\bibitem[\protect\citeauthoryear{Libovick{\`y}, Palaskar, Gella, and
  Metze}{Libovick{\`y} et~al\mbox{.}}{2018}]%
        {libovicky2018multimodal}
\bibfield{author}{\bibinfo{person}{Jindrich Libovick{\`y}},
  \bibinfo{person}{Shruti Palaskar}, \bibinfo{person}{Spandana Gella}, {and}
  \bibinfo{person}{Florian Metze}.} \bibinfo{year}{2018}\natexlab{}.
\newblock \showarticletitle{Multimodal Abstractive Summarization for
  Open-Domain Videos}. In \bibinfo{booktitle}{\emph{Proceedings of the Workshop
  on Visually Grounded Interaction and Language (ViGIL). NIPS}}.
\newblock


\bibitem[\protect\citeauthoryear{Lin}{Lin}{2004}]%
        {lin-2004-rouge}
\bibfield{author}{\bibinfo{person}{Chin-Yew Lin}.}
  \bibinfo{year}{2004}\natexlab{}.
\newblock \showarticletitle{{ROUGE}: A Package for Automatic Evaluation of
  Summaries}. In \bibinfo{booktitle}{\emph{Text Summarization Branches Out}}.
  \bibinfo{publisher}{Association for Computational Linguistics},
  \bibinfo{address}{Barcelona, Spain}, \bibinfo{pages}{74--81}.
\newblock
\urldef\tempurl%
\url{https://www.aclweb.org/anthology/W04-1013}
\showURL{%
\tempurl}


\bibitem[\protect\citeauthoryear{Lin and Bilmes}{Lin and Bilmes}{2010}]%
        {lin2010multi}
\bibfield{author}{\bibinfo{person}{Hui Lin} {and} \bibinfo{person}{Jeff
  Bilmes}.} \bibinfo{year}{2010}\natexlab{}.
\newblock \showarticletitle{Multi-document summarization via budgeted
  maximization of submodular functions}. In \bibinfo{booktitle}{\emph{Human
  Language Technologies: The 2010 Annual Conference of the North American
  Chapter of the Association for Computational Linguistics}}.
  \bibinfo{pages}{912--920}.
\newblock


\bibitem[\protect\citeauthoryear{Lin, Bertasius, Wang, Chang, Parikh, and
  Torresani}{Lin et~al\mbox{.}}{2021}]%
        {lin2021vx2text}
\bibfield{author}{\bibinfo{person}{Xudong Lin}, \bibinfo{person}{Gedas
  Bertasius}, \bibinfo{person}{Jue Wang}, \bibinfo{person}{Shih-Fu Chang},
  \bibinfo{person}{Devi Parikh}, {and} \bibinfo{person}{Lorenzo Torresani}.}
  \bibinfo{year}{2021}\natexlab{}.
\newblock \showarticletitle{VX2TEXT: End-to-End Learning of Video-Based Text
  Generation From Multimodal Inputs}. In \bibinfo{booktitle}{\emph{Proceedings
  of the IEEE/CVF Conference on Computer Vision and Pattern Recognition}}.
  \bibinfo{pages}{7005--7015}.
\newblock


\bibitem[\protect\citeauthoryear{Litvak and Vanetik}{Litvak and
  Vanetik}{2017}]%
        {litvak2017query}
\bibfield{author}{\bibinfo{person}{Marina Litvak} {and}
  \bibinfo{person}{Natalia Vanetik}.} \bibinfo{year}{2017}\natexlab{}.
\newblock \showarticletitle{Query-based summarization using MDL principle}. In
  \bibinfo{booktitle}{\emph{Proceedings of the multiling 2017 workshop on
  summarization and summary evaluation across source types and genres}}.
  \bibinfo{pages}{22--31}.
\newblock


\bibitem[\protect\citeauthoryear{Liu, Chen, and Tseng}{Liu
  et~al\mbox{.}}{2015}]%
        {liu2015incrests}
\bibfield{author}{\bibinfo{person}{Cheng-Ying Liu}, \bibinfo{person}{Ming-Syan
  Chen}, {and} \bibinfo{person}{Chi-Yao Tseng}.}
  \bibinfo{year}{2015}\natexlab{}.
\newblock \showarticletitle{Incrests: Towards real-time incremental short text
  summarization on comment streams from social network services}.
\newblock \bibinfo{journal}{\emph{IEEE Transactions on Knowledge and Data
  Engineering}} \bibinfo{volume}{27}, \bibinfo{number}{11}
  (\bibinfo{year}{2015}), \bibinfo{pages}{2986--3000}.
\newblock


\bibitem[\protect\citeauthoryear{Liu, Li, Xu, and Natarajan}{Liu
  et~al\mbox{.}}{2018}]%
        {liu2018learn}
\bibfield{author}{\bibinfo{person}{Kuan Liu}, \bibinfo{person}{Yanen Li},
  \bibinfo{person}{Ning Xu}, {and} \bibinfo{person}{Prem Natarajan}.}
  \bibinfo{year}{2018}\natexlab{}.
\newblock \showarticletitle{Learn to combine modalities in multimodal deep
  learning}.
\newblock \bibinfo{journal}{\emph{arXiv preprint arXiv:1805.11730}}
  (\bibinfo{year}{2018}).
\newblock


\bibitem[\protect\citeauthoryear{Liu, Sun, Yu, Zhang, and Xu}{Liu
  et~al\mbox{.}}{2020}]%
        {liu-etal-2020-multistage}
\bibfield{author}{\bibinfo{person}{Nayu Liu}, \bibinfo{person}{Xian Sun},
  \bibinfo{person}{Hongfeng Yu}, \bibinfo{person}{Wenkai Zhang}, {and}
  \bibinfo{person}{Guangluan Xu}.} \bibinfo{year}{2020}\natexlab{}.
\newblock \showarticletitle{Multistage Fusion with Forget Gate for Multimodal
  Summarization in Open-Domain Videos}. In
  \bibinfo{booktitle}{\emph{Proceedings of the 2020 Conference on Empirical
  Methods in Natural Language Processing (EMNLP)}}.
  \bibinfo{publisher}{Association for Computational Linguistics},
  \bibinfo{address}{Online}, \bibinfo{pages}{1834--1845}.
\newblock
\urldef\tempurl%
\url{https://doi.org/10.18653/v1/2020.emnlp-main.144}
\showDOI{\tempurl}


\bibitem[\protect\citeauthoryear{Liu and Jiang}{Liu and Jiang}{2020}]%
        {liu2020aesthetic}
\bibfield{author}{\bibinfo{person}{Xin Liu} {and} \bibinfo{person}{Yujia
  Jiang}.} \bibinfo{year}{2020}\natexlab{}.
\newblock \showarticletitle{Aesthetic assessment of website design based on
  multimodal fusion}.
\newblock \bibinfo{journal}{\emph{Future Generation Computer Systems}}
  (\bibinfo{year}{2020}).
\newblock


\bibitem[\protect\citeauthoryear{Liu, Liu, Radev, and Neubig}{Liu
  et~al\mbox{.}}{2022}]%
        {liu2022brio}
\bibfield{author}{\bibinfo{person}{Yixin Liu}, \bibinfo{person}{Pengfei Liu},
  \bibinfo{person}{Dragomir Radev}, {and} \bibinfo{person}{Graham Neubig}.}
  \bibinfo{year}{2022}\natexlab{}.
\newblock \showarticletitle{BRIO: Bringing Order to Abstractive Summarization}.
  In \bibinfo{booktitle}{\emph{Proceedings of the 60th Annual Meeting of the
  Association for Computational Linguistics (Volume 1: Long Papers)}}.
  \bibinfo{pages}{2890--2903}.
\newblock


\bibitem[\protect\citeauthoryear{Lu, Batra, Parikh, and Lee}{Lu
  et~al\mbox{.}}{2019}]%
        {lu2019vilbert}
\bibfield{author}{\bibinfo{person}{Jiasen Lu}, \bibinfo{person}{Dhruv Batra},
  \bibinfo{person}{Devi Parikh}, {and} \bibinfo{person}{Stefan Lee}.}
  \bibinfo{year}{2019}\natexlab{}.
\newblock \showarticletitle{Vilbert: Pretraining task-agnostic visiolinguistic
  representations for vision-and-language tasks}. In
  \bibinfo{booktitle}{\emph{Advances in Neural Information Processing
  Systems}}. \bibinfo{pages}{13--23}.
\newblock


\bibitem[\protect\citeauthoryear{Luhn}{Luhn}{1958}]%
        {luhn1958automatic}
\bibfield{author}{\bibinfo{person}{Hans~Peter Luhn}.}
  \bibinfo{year}{1958}\natexlab{}.
\newblock \showarticletitle{The automatic creation of literature abstracts}.
\newblock \bibinfo{journal}{\emph{IBM Journal of research and development}}
  \bibinfo{volume}{2}, \bibinfo{number}{2} (\bibinfo{year}{1958}),
  \bibinfo{pages}{159--165}.
\newblock


\bibitem[\protect\citeauthoryear{Ma, Zhang, Guo, Wang, and Sheng}{Ma
  et~al\mbox{.}}{2020}]%
        {ma2020multidocument}
\bibfield{author}{\bibinfo{person}{Congbo Ma}, \bibinfo{person}{Wei~Emma
  Zhang}, \bibinfo{person}{Mingyu Guo}, \bibinfo{person}{Hu Wang}, {and}
  \bibinfo{person}{Quan~Z. Sheng}.} \bibinfo{year}{2020}\natexlab{}.
\newblock \bibinfo{title}{Multi-document Summarization via Deep Learning
  Techniques: A Survey}.
\newblock
\newblock
\showeprint[arxiv]{2011.04843}~[cs.CL]


\bibitem[\protect\citeauthoryear{Maynez, Narayan, Bohnet, and McDonald}{Maynez
  et~al\mbox{.}}{2020}]%
        {maynez2020faithfulness}
\bibfield{author}{\bibinfo{person}{Joshua Maynez}, \bibinfo{person}{Shashi
  Narayan}, \bibinfo{person}{Bernd Bohnet}, {and} \bibinfo{person}{Ryan
  McDonald}.} \bibinfo{year}{2020}\natexlab{}.
\newblock \showarticletitle{On faithfulness and factuality in abstractive
  summarization}.
\newblock \bibinfo{journal}{\emph{arXiv preprint arXiv:2005.00661}}
  (\bibinfo{year}{2020}).
\newblock


\bibitem[\protect\citeauthoryear{Miao, Cao, Li, and Guan}{Miao
  et~al\mbox{.}}{2020}]%
        {miao2020multi}
\bibfield{author}{\bibinfo{person}{Lianhai Miao}, \bibinfo{person}{Da Cao},
  \bibinfo{person}{Juntao Li}, {and} \bibinfo{person}{Weili Guan}.}
  \bibinfo{year}{2020}\natexlab{}.
\newblock \showarticletitle{Multi-modal product title compression}.
\newblock \bibinfo{journal}{\emph{Information Processing \& Management}}
  \bibinfo{volume}{57}, \bibinfo{number}{1} (\bibinfo{year}{2020}),
  \bibinfo{pages}{102123}.
\newblock


\bibitem[\protect\citeauthoryear{Mihalcea}{Mihalcea}{2004}]%
        {mihalcea2004graph}
\bibfield{author}{\bibinfo{person}{Rada Mihalcea}.}
  \bibinfo{year}{2004}\natexlab{}.
\newblock \showarticletitle{Graph-based ranking algorithms for sentence
  extraction, applied to text summarization}. In
  \bibinfo{booktitle}{\emph{Proceedings of the ACL Interactive Poster and
  Demonstration Sessions}}. \bibinfo{pages}{170--173}.
\newblock


\bibitem[\protect\citeauthoryear{Mihalcea and Tarau}{Mihalcea and
  Tarau}{2004}]%
        {mihalcea2004textrank}
\bibfield{author}{\bibinfo{person}{Rada Mihalcea} {and} \bibinfo{person}{Paul
  Tarau}.} \bibinfo{year}{2004}\natexlab{}.
\newblock \showarticletitle{Textrank: Bringing order into text}. In
  \bibinfo{booktitle}{\emph{Proceedings of the 2004 conference on empirical
  methods in natural language processing}}. \bibinfo{pages}{404--411}.
\newblock


\bibitem[\protect\citeauthoryear{Mikolov, Sutskever, Chen, Corrado, and
  Dean}{Mikolov et~al\mbox{.}}{2013}]%
        {mikolov2013distributed}
\bibfield{author}{\bibinfo{person}{Tomas Mikolov}, \bibinfo{person}{Ilya
  Sutskever}, \bibinfo{person}{Kai Chen}, \bibinfo{person}{Greg~S Corrado},
  {and} \bibinfo{person}{Jeff Dean}.} \bibinfo{year}{2013}\natexlab{}.
\newblock \showarticletitle{Distributed representations of words and phrases
  and their compositionality}. In \bibinfo{booktitle}{\emph{Advances in neural
  information processing systems}}. \bibinfo{pages}{3111--3119}.
\newblock


\bibitem[\protect\citeauthoryear{Mirjalili, Mirjalili, and Lewis}{Mirjalili
  et~al\mbox{.}}{2014}]%
        {mirjalili2014grey}
\bibfield{author}{\bibinfo{person}{Seyedali Mirjalili},
  \bibinfo{person}{Seyed~Mohammad Mirjalili}, {and} \bibinfo{person}{Andrew
  Lewis}.} \bibinfo{year}{2014}\natexlab{}.
\newblock \showarticletitle{Grey wolf optimizer}.
\newblock \bibinfo{journal}{\emph{Advances in engineering software}}
  \bibinfo{volume}{69} (\bibinfo{year}{2014}), \bibinfo{pages}{46--61}.
\newblock


\bibitem[\protect\citeauthoryear{Modani, Khabiri, Srinivasan, and
  Caverlee}{Modani et~al\mbox{.}}{2015}]%
        {modani2015creating}
\bibfield{author}{\bibinfo{person}{Natwar Modani}, \bibinfo{person}{Elham
  Khabiri}, \bibinfo{person}{Harini Srinivasan}, {and} \bibinfo{person}{James
  Caverlee}.} \bibinfo{year}{2015}\natexlab{}.
\newblock \showarticletitle{Creating diverse product review summaries: a graph
  approach}. In \bibinfo{booktitle}{\emph{International Conference on Web
  Information Systems Engineering}}. Springer, \bibinfo{pages}{169--184}.
\newblock


\bibitem[\protect\citeauthoryear{Modani, Maneriker, Hiranandani, Sinha,
  Subramanian, Gupta, et~al\mbox{.}}{Modani et~al\mbox{.}}{2016}]%
        {modani2016summarizing}
\bibfield{author}{\bibinfo{person}{Natwar Modani}, \bibinfo{person}{Pranav
  Maneriker}, \bibinfo{person}{Gaurush Hiranandani}, \bibinfo{person}{Atanu~R
  Sinha}, \bibinfo{person}{Vaishnavi Subramanian}, \bibinfo{person}{Shivani
  Gupta}, {et~al\mbox{.}}} \bibinfo{year}{2016}\natexlab{}.
\newblock \showarticletitle{Summarizing multimedia content}. In
  \bibinfo{booktitle}{\emph{International Conference on Web Information Systems
  Engineering}}. Springer, \bibinfo{pages}{340--348}.
\newblock


\bibitem[\protect\citeauthoryear{Money and Agius}{Money and Agius}{2008}]%
        {money2008video}
\bibfield{author}{\bibinfo{person}{Arthur~G Money} {and} \bibinfo{person}{Harry
  Agius}.} \bibinfo{year}{2008}\natexlab{}.
\newblock \showarticletitle{Video summarisation: A conceptual framework and
  survey of the state of the art}.
\newblock \bibinfo{journal}{\emph{Journal of visual communication and image
  representation}} \bibinfo{volume}{19}, \bibinfo{number}{2}
  (\bibinfo{year}{2008}), \bibinfo{pages}{121--143}.
\newblock


\bibitem[\protect\citeauthoryear{Moon, Neves, and Carvalho}{Moon
  et~al\mbox{.}}{2018a}]%
        {moon2018multimodalb}
\bibfield{author}{\bibinfo{person}{Seungwhan Moon}, \bibinfo{person}{Leonardo
  Neves}, {and} \bibinfo{person}{Vitor Carvalho}.}
  \bibinfo{year}{2018}\natexlab{a}.
\newblock \showarticletitle{Multimodal named entity disambiguation for noisy
  social media posts}. In \bibinfo{booktitle}{\emph{Proceedings of the 56th
  Annual Meeting of the Association for Computational Linguistics (Volume 1:
  Long Papers)}}. \bibinfo{pages}{2000--2008}.
\newblock


\bibitem[\protect\citeauthoryear{Moon, Neves, and Carvalho}{Moon
  et~al\mbox{.}}{2018b}]%
        {moon2018multimodal}
\bibfield{author}{\bibinfo{person}{Seungwhan Moon}, \bibinfo{person}{Leonardo
  Neves}, {and} \bibinfo{person}{Vitor Carvalho}.}
  \bibinfo{year}{2018}\natexlab{b}.
\newblock \showarticletitle{Multimodal named entity recognition for short
  social media posts}.
\newblock \bibinfo{journal}{\emph{arXiv preprint arXiv:1802.07862}}
  (\bibinfo{year}{2018}).
\newblock


\bibitem[\protect\citeauthoryear{Morency, Mihalcea, and Doshi}{Morency
  et~al\mbox{.}}{2011}]%
        {morency2011towards}
\bibfield{author}{\bibinfo{person}{Louis-Philippe Morency},
  \bibinfo{person}{Rada Mihalcea}, {and} \bibinfo{person}{Payal Doshi}.}
  \bibinfo{year}{2011}\natexlab{}.
\newblock \showarticletitle{Towards multimodal sentiment analysis: Harvesting
  opinions from the web}. In \bibinfo{booktitle}{\emph{Proceedings of the 13th
  international conference on multimodal interfaces}}.
  \bibinfo{pages}{169--176}.
\newblock


\bibitem[\protect\citeauthoryear{Mukherjee, Jangra, Saha, and Jatowt}{Mukherjee
  et~al\mbox{.}}{2022}]%
        {mukherjee2022topic}
\bibfield{author}{\bibinfo{person}{Sourajit Mukherjee},
  \bibinfo{person}{Anubhav Jangra}, \bibinfo{person}{Sriparna Saha}, {and}
  \bibinfo{person}{Adam Jatowt}.} \bibinfo{year}{2022}\natexlab{}.
\newblock \showarticletitle{Topic-aware Multimodal Summarization}. In
  \bibinfo{booktitle}{\emph{Findings of the Association for Computational
  Linguistics: AACL-IJCNLP 2022}}. \bibinfo{pages}{387--398}.
\newblock


\bibitem[\protect\citeauthoryear{Nemhauser, Wolsey, and Fisher}{Nemhauser
  et~al\mbox{.}}{1978}]%
        {nemhauser1978analysis}
\bibfield{author}{\bibinfo{person}{George~L Nemhauser},
  \bibinfo{person}{Laurence~A Wolsey}, {and} \bibinfo{person}{Marshall~L
  Fisher}.} \bibinfo{year}{1978}\natexlab{}.
\newblock \showarticletitle{An analysis of approximations for maximizing
  submodular set functions—I}.
\newblock \bibinfo{journal}{\emph{Mathematical programming}}
  \bibinfo{volume}{14}, \bibinfo{number}{1} (\bibinfo{year}{1978}),
  \bibinfo{pages}{265--294}.
\newblock


\bibitem[\protect\citeauthoryear{Nenkova and McKeown}{Nenkova and
  McKeown}{2012}]%
        {nenkova2012survey}
\bibfield{author}{\bibinfo{person}{Ani Nenkova} {and} \bibinfo{person}{Kathleen
  McKeown}.} \bibinfo{year}{2012}\natexlab{}.
\newblock \showarticletitle{A survey of text summarization techniques}.
\newblock In \bibinfo{booktitle}{\emph{Mining text data}}.
  \bibinfo{publisher}{Springer}, \bibinfo{pages}{43--76}.
\newblock


\bibitem[\protect\citeauthoryear{Ng and Abrecht}{Ng and Abrecht}{2015}]%
        {ng-abrecht-2015-better}
\bibfield{author}{\bibinfo{person}{Jun-Ping Ng} {and} \bibinfo{person}{Viktoria
  Abrecht}.} \bibinfo{year}{2015}\natexlab{}.
\newblock \showarticletitle{Better Summarization Evaluation with Word
  Embeddings for {ROUGE}}. In \bibinfo{booktitle}{\emph{Proceedings of the 2015
  Conference on Empirical Methods in Natural Language Processing}}.
  \bibinfo{publisher}{Association for Computational Linguistics},
  \bibinfo{address}{Lisbon, Portugal}, \bibinfo{pages}{1925--1930}.
\newblock
\urldef\tempurl%
\url{https://doi.org/10.18653/v1/D15-1222}
\showDOI{\tempurl}


\bibitem[\protect\citeauthoryear{Oskouie, Alipour, and
  Eftekhari-Moghadam}{Oskouie et~al\mbox{.}}{2014}]%
        {oskouie2014multimodal}
\bibfield{author}{\bibinfo{person}{Payam Oskouie}, \bibinfo{person}{Sara
  Alipour}, {and} \bibinfo{person}{Amir-Masoud Eftekhari-Moghadam}.}
  \bibinfo{year}{2014}\natexlab{}.
\newblock \showarticletitle{Multimodal feature extraction and fusion for
  semantic mining of soccer video: a survey}.
\newblock \bibinfo{journal}{\emph{Artificial Intelligence Review}}
  \bibinfo{volume}{42}, \bibinfo{number}{2} (\bibinfo{year}{2014}),
  \bibinfo{pages}{173--210}.
\newblock


\bibitem[\protect\citeauthoryear{Pakhira, Bandyopadhyay, and Maulik}{Pakhira
  et~al\mbox{.}}{2004}]%
        {pakhira2004validity}
\bibfield{author}{\bibinfo{person}{Malay~K Pakhira},
  \bibinfo{person}{Sanghamitra Bandyopadhyay}, {and} \bibinfo{person}{Ujjwal
  Maulik}.} \bibinfo{year}{2004}\natexlab{}.
\newblock \showarticletitle{Validity index for crisp and fuzzy clusters}.
\newblock \bibinfo{journal}{\emph{Pattern recognition}} \bibinfo{volume}{37},
  \bibinfo{number}{3} (\bibinfo{year}{2004}), \bibinfo{pages}{487--501}.
\newblock


\bibitem[\protect\citeauthoryear{Palaskar, Libovick{\`y}, Gella, and
  Metze}{Palaskar et~al\mbox{.}}{2019}]%
        {palaskar2019multimodal}
\bibfield{author}{\bibinfo{person}{Shruti Palaskar}, \bibinfo{person}{Jindrich
  Libovick{\`y}}, \bibinfo{person}{Spandana Gella}, {and}
  \bibinfo{person}{Florian Metze}.} \bibinfo{year}{2019}\natexlab{}.
\newblock \showarticletitle{Multimodal abstractive summarization for how2
  videos}.
\newblock \bibinfo{journal}{\emph{arXiv preprint arXiv:1906.07901}}
  (\bibinfo{year}{2019}).
\newblock


\bibitem[\protect\citeauthoryear{Parmar, Vaswani, Uszkoreit, Kaiser, Shazeer,
  Ku, and Tran}{Parmar et~al\mbox{.}}{2018}]%
        {parmar2018image}
\bibfield{author}{\bibinfo{person}{Niki Parmar}, \bibinfo{person}{Ashish
  Vaswani}, \bibinfo{person}{Jakob Uszkoreit}, \bibinfo{person}{Lukasz Kaiser},
  \bibinfo{person}{Noam Shazeer}, \bibinfo{person}{Alexander Ku}, {and}
  \bibinfo{person}{Dustin Tran}.} \bibinfo{year}{2018}\natexlab{}.
\newblock \showarticletitle{Image transformer}. In
  \bibinfo{booktitle}{\emph{International Conference on Machine Learning}}.
  PMLR, \bibinfo{pages}{4055--4064}.
\newblock


\bibitem[\protect\citeauthoryear{Patrick, Huang, Asano, Metze, Hauptmann,
  Henriques, and Vedaldi}{Patrick et~al\mbox{.}}{2020}]%
        {patrick2020support}
\bibfield{author}{\bibinfo{person}{Mandela Patrick}, \bibinfo{person}{Po-Yao
  Huang}, \bibinfo{person}{Yuki Asano}, \bibinfo{person}{Florian Metze},
  \bibinfo{person}{Alexander Hauptmann}, \bibinfo{person}{Joao Henriques},
  {and} \bibinfo{person}{Andrea Vedaldi}.} \bibinfo{year}{2020}\natexlab{}.
\newblock \showarticletitle{Support-set bottlenecks for video-text
  representation learning}.
\newblock \bibinfo{journal}{\emph{arXiv preprint arXiv:2010.02824}}
  (\bibinfo{year}{2020}).
\newblock


\bibitem[\protect\citeauthoryear{Pearson}{Pearson}{1901}]%
        {pearson1901liii}
\bibfield{author}{\bibinfo{person}{Karl Pearson}.}
  \bibinfo{year}{1901}\natexlab{}.
\newblock \showarticletitle{LIII. On lines and planes of closest fit to systems
  of points in space}.
\newblock \bibinfo{journal}{\emph{The London, Edinburgh, and Dublin
  Philosophical Magazine and Journal of Science}} \bibinfo{volume}{2},
  \bibinfo{number}{11} (\bibinfo{year}{1901}), \bibinfo{pages}{559--572}.
\newblock


\bibitem[\protect\citeauthoryear{Pennington, Socher, and Manning}{Pennington
  et~al\mbox{.}}{2014}]%
        {pennington2014glove}
\bibfield{author}{\bibinfo{person}{Jeffrey Pennington},
  \bibinfo{person}{Richard Socher}, {and} \bibinfo{person}{Christopher~D
  Manning}.} \bibinfo{year}{2014}\natexlab{}.
\newblock \showarticletitle{Glove: Global vectors for word representation}. In
  \bibinfo{booktitle}{\emph{Proceedings of the 2014 conference on empirical
  methods in natural language processing (EMNLP)}}.
  \bibinfo{pages}{1532--1543}.
\newblock


\bibitem[\protect\citeauthoryear{Peyrard, Botschen, and Gurevych}{Peyrard
  et~al\mbox{.}}{2017}]%
        {peyrard-etal-2017-learning}
\bibfield{author}{\bibinfo{person}{Maxime Peyrard}, \bibinfo{person}{Teresa
  Botschen}, {and} \bibinfo{person}{Iryna Gurevych}.}
  \bibinfo{year}{2017}\natexlab{}.
\newblock \showarticletitle{Learning to Score System Summaries for Better
  Content Selection Evaluation.}. In \bibinfo{booktitle}{\emph{Proceedings of
  the Workshop on New Frontiers in Summarization}}.
  \bibinfo{publisher}{Association for Computational Linguistics},
  \bibinfo{address}{Copenhagen, Denmark}.
\newblock
\urldef\tempurl%
\url{https://doi.org/10.18653/v1/W17-4510}
\showDOI{\tempurl}


\bibitem[\protect\citeauthoryear{Qian, Li, Ren, and Jiang}{Qian
  et~al\mbox{.}}{2019}]%
        {qian2019social}
\bibfield{author}{\bibinfo{person}{Xueming Qian}, \bibinfo{person}{Mingdi Li},
  \bibinfo{person}{Yayun Ren}, {and} \bibinfo{person}{Shuhui Jiang}.}
  \bibinfo{year}{2019}\natexlab{}.
\newblock \showarticletitle{Social media based event summarization by
  user--text--image co-clustering}.
\newblock \bibinfo{journal}{\emph{Knowledge-Based Systems}}
  \bibinfo{volume}{164} (\bibinfo{year}{2019}), \bibinfo{pages}{107--121}.
\newblock


\bibitem[\protect\citeauthoryear{Qian, Xue, Yang, Tang, Hou, and Mei}{Qian
  et~al\mbox{.}}{2014}]%
        {qian2014landmark}
\bibfield{author}{\bibinfo{person}{Xueming Qian}, \bibinfo{person}{Yao Xue},
  \bibinfo{person}{Xiyu Yang}, \bibinfo{person}{Yuan~Yan Tang},
  \bibinfo{person}{Xingsong Hou}, {and} \bibinfo{person}{Tao Mei}.}
  \bibinfo{year}{2014}\natexlab{}.
\newblock \showarticletitle{Landmark summarization with diverse viewpoints}.
\newblock \bibinfo{journal}{\emph{IEEE Transactions on Circuits and Systems for
  Video Technology}} \bibinfo{volume}{25}, \bibinfo{number}{11}
  (\bibinfo{year}{2014}), \bibinfo{pages}{1857--1869}.
\newblock


\bibitem[\protect\citeauthoryear{Qian, Zhong, and Zhou}{Qian
  et~al\mbox{.}}{2018}]%
        {qian2018multimodal}
\bibfield{author}{\bibinfo{person}{Xin Qian}, \bibinfo{person}{Ziyi Zhong},
  {and} \bibinfo{person}{Jieli Zhou}.} \bibinfo{year}{2018}\natexlab{}.
\newblock \showarticletitle{Multimodal machine translation with reinforcement
  learning}.
\newblock \bibinfo{journal}{\emph{arXiv preprint arXiv:1805.02356}}
  (\bibinfo{year}{2018}).
\newblock


\bibitem[\protect\citeauthoryear{Rahman and Borah}{Rahman and Borah}{2019}]%
        {rahman2019improvement}
\bibfield{author}{\bibinfo{person}{Nazreena Rahman} {and}
  \bibinfo{person}{Bhogeswar Borah}.} \bibinfo{year}{2019}\natexlab{}.
\newblock \showarticletitle{Improvement of query-based text summarization using
  word sense disambiguation}.
\newblock \bibinfo{journal}{\emph{Complex \& Intelligent Systems}}
  (\bibinfo{year}{2019}), \bibinfo{pages}{1--11}.
\newblock


\bibitem[\protect\citeauthoryear{Ramachandram and Taylor}{Ramachandram and
  Taylor}{2017}]%
        {ramachandram2017deep}
\bibfield{author}{\bibinfo{person}{Dhanesh Ramachandram} {and}
  \bibinfo{person}{Graham~W Taylor}.} \bibinfo{year}{2017}\natexlab{}.
\newblock \showarticletitle{Deep multimodal learning: A survey on recent
  advances and trends}.
\newblock \bibinfo{journal}{\emph{IEEE Signal Processing Magazine}}
  \bibinfo{volume}{34}, \bibinfo{number}{6} (\bibinfo{year}{2017}),
  \bibinfo{pages}{96--108}.
\newblock


\bibitem[\protect\citeauthoryear{Ramanishka, Das, Park, Venugopalan, Hendricks,
  Rohrbach, and Saenko}{Ramanishka et~al\mbox{.}}{2016}]%
        {ramanishka2016multimodal}
\bibfield{author}{\bibinfo{person}{Vasili Ramanishka}, \bibinfo{person}{Abir
  Das}, \bibinfo{person}{Dong~Huk Park}, \bibinfo{person}{Subhashini
  Venugopalan}, \bibinfo{person}{Lisa~Anne Hendricks}, \bibinfo{person}{Marcus
  Rohrbach}, {and} \bibinfo{person}{Kate Saenko}.}
  \bibinfo{year}{2016}\natexlab{}.
\newblock \showarticletitle{Multimodal video description}. In
  \bibinfo{booktitle}{\emph{Proceedings of the 24th ACM international
  conference on Multimedia}}. \bibinfo{pages}{1092--1096}.
\newblock


\bibitem[\protect\citeauthoryear{Ramesh, Pavlov, Goh, Gray, Voss, Radford,
  Chen, and Sutskever}{Ramesh et~al\mbox{.}}{2021}]%
        {ramesh2021zeroshot}
\bibfield{author}{\bibinfo{person}{Aditya Ramesh}, \bibinfo{person}{Mikhail
  Pavlov}, \bibinfo{person}{Gabriel Goh}, \bibinfo{person}{Scott Gray},
  \bibinfo{person}{Chelsea Voss}, \bibinfo{person}{Alec Radford},
  \bibinfo{person}{Mark Chen}, {and} \bibinfo{person}{Ilya Sutskever}.}
  \bibinfo{year}{2021}\natexlab{}.
\newblock \bibinfo{title}{Zero-Shot Text-to-Image Generation}.
\newblock
\newblock
\showeprint[arxiv]{2102.12092}~[cs.CV]


\bibitem[\protect\citeauthoryear{Rashtchian, Young, Hodosh, and
  Hockenmaier}{Rashtchian et~al\mbox{.}}{2010}]%
        {rashtchian2010collecting}
\bibfield{author}{\bibinfo{person}{Cyrus Rashtchian}, \bibinfo{person}{Peter
  Young}, \bibinfo{person}{Micah Hodosh}, {and} \bibinfo{person}{Julia
  Hockenmaier}.} \bibinfo{year}{2010}\natexlab{}.
\newblock \showarticletitle{Collecting image annotations using amazon’s
  mechanical turk}. In \bibinfo{booktitle}{\emph{Proceedings of the NAACL HLT
  2010 Workshop on Creating Speech and Language Data with Amazon’s Mechanical
  Turk}}. \bibinfo{pages}{139--147}.
\newblock


\bibitem[\protect\citeauthoryear{Rosas, Mihalcea, and Morency}{Rosas
  et~al\mbox{.}}{2013}]%
        {rosas2013multimodal}
\bibfield{author}{\bibinfo{person}{Ver{\'o}nica~P{\'e}rez Rosas},
  \bibinfo{person}{Rada Mihalcea}, {and} \bibinfo{person}{Louis-Philippe
  Morency}.} \bibinfo{year}{2013}\natexlab{}.
\newblock \showarticletitle{Multimodal sentiment analysis of spanish online
  videos}.
\newblock \bibinfo{journal}{\emph{IEEE Intelligent Systems}}
  \bibinfo{volume}{28}, \bibinfo{number}{3} (\bibinfo{year}{2013}),
  \bibinfo{pages}{38--45}.
\newblock


\bibitem[\protect\citeauthoryear{Ryden, Park, Ulriksen, and Miller}{Ryden
  et~al\mbox{.}}{2004}]%
        {ryden2004multimodal}
\bibfield{author}{\bibinfo{person}{Nils Ryden}, \bibinfo{person}{Choon~B Park},
  \bibinfo{person}{Peter Ulriksen}, {and} \bibinfo{person}{Richard~D Miller}.}
  \bibinfo{year}{2004}\natexlab{}.
\newblock \showarticletitle{Multimodal approach to seismic pavement testing}.
\newblock \bibinfo{journal}{\emph{Journal of geotechnical and geoenvironmental
  engineering}} \bibinfo{volume}{130}, \bibinfo{number}{6}
  (\bibinfo{year}{2004}), \bibinfo{pages}{636--645}.
\newblock


\bibitem[\protect\citeauthoryear{Saggion and Poibeau}{Saggion and
  Poibeau}{2013}]%
        {saggion2013automatic}
\bibfield{author}{\bibinfo{person}{Horacio Saggion} {and}
  \bibinfo{person}{Thierry Poibeau}.} \bibinfo{year}{2013}\natexlab{}.
\newblock \showarticletitle{Automatic text summarization: Past, present and
  future}.
\newblock In \bibinfo{booktitle}{\emph{Multi-source, multilingual information
  extraction and summarization}}. \bibinfo{publisher}{Springer},
  \bibinfo{pages}{3--21}.
\newblock


\bibitem[\protect\citeauthoryear{Sahuguet and Huet}{Sahuguet and Huet}{2013}]%
        {sahuguet2013socially}
\bibfield{author}{\bibinfo{person}{Mathilde Sahuguet} {and}
  \bibinfo{person}{Benoit Huet}.} \bibinfo{year}{2013}\natexlab{}.
\newblock \showarticletitle{Socially motivated multimedia topic timeline
  summarization}. In \bibinfo{booktitle}{\emph{Proceedings of the 2nd
  international workshop on Socially-aware multimedia}}.
  \bibinfo{pages}{19--24}.
\newblock


\bibitem[\protect\citeauthoryear{Saini, Saha, Bhattacharyya, Mrinal, and
  Mishra}{Saini et~al\mbox{.}}{2021}]%
        {saini2021multimodal}
\bibfield{author}{\bibinfo{person}{Naveen Saini}, \bibinfo{person}{Sriparna
  Saha}, \bibinfo{person}{Pushpak Bhattacharyya}, \bibinfo{person}{Shubhankar
  Mrinal}, {and} \bibinfo{person}{Santosh~Kumar Mishra}.}
  \bibinfo{year}{2021}\natexlab{}.
\newblock \showarticletitle{On multimodal microblog summarization}.
\newblock \bibinfo{journal}{\emph{IEEE Transactions on Computational Social
  Systems}} \bibinfo{volume}{9}, \bibinfo{number}{5} (\bibinfo{year}{2021}),
  \bibinfo{pages}{1317--1329}.
\newblock


\bibitem[\protect\citeauthoryear{Saini, Saha, Chakraborty, and
  Bhattacharyya}{Saini et~al\mbox{.}}{2019a}]%
        {saini2019extractive2}
\bibfield{author}{\bibinfo{person}{Naveen Saini}, \bibinfo{person}{Sriparna
  Saha}, \bibinfo{person}{Dhiraj Chakraborty}, {and} \bibinfo{person}{Pushpak
  Bhattacharyya}.} \bibinfo{year}{2019}\natexlab{a}.
\newblock \showarticletitle{Extractive single document summarization using
  binary differential evolution: Optimization of different sentence quality
  measures}.
\newblock \bibinfo{journal}{\emph{PloS one}} \bibinfo{volume}{14},
  \bibinfo{number}{11} (\bibinfo{year}{2019}), \bibinfo{pages}{e0223477}.
\newblock


\bibitem[\protect\citeauthoryear{Saini, Saha, Jangra, and Bhattacharyya}{Saini
  et~al\mbox{.}}{2019b}]%
        {saini2019extractive}
\bibfield{author}{\bibinfo{person}{Naveen Saini}, \bibinfo{person}{Sriparna
  Saha}, \bibinfo{person}{Anubhav Jangra}, {and} \bibinfo{person}{Pushpak
  Bhattacharyya}.} \bibinfo{year}{2019}\natexlab{b}.
\newblock \showarticletitle{Extractive single document summarization using
  multi-objective optimization: Exploring self-organized differential
  evolution, grey wolf optimizer and water cycle algorithm}.
\newblock \bibinfo{journal}{\emph{Knowledge-Based Systems}}
  \bibinfo{volume}{164} (\bibinfo{year}{2019}), \bibinfo{pages}{45--67}.
\newblock


\bibitem[\protect\citeauthoryear{Salton}{Salton}{1989}]%
        {salton1989automatic}
\bibfield{author}{\bibinfo{person}{Gerard Salton}.}
  \bibinfo{year}{1989}\natexlab{}.
\newblock \showarticletitle{Automatic text processing: The transformation,
  analysis, and retrieval of}.
\newblock \bibinfo{journal}{\emph{Reading: Addison-Wesley}}
  \bibinfo{volume}{169} (\bibinfo{year}{1989}).
\newblock


\bibitem[\protect\citeauthoryear{Sanabria, Precioso, and Menguy}{Sanabria
  et~al\mbox{.}}{2019}]%
        {sanabria2019deep}
\bibfield{author}{\bibinfo{person}{Melissa Sanabria},
  \bibinfo{person}{Fr{\'e}d{\'e}ric Precioso}, {and} \bibinfo{person}{Thomas
  Menguy}.} \bibinfo{year}{2019}\natexlab{}.
\newblock \showarticletitle{A Deep Architecture for Multimodal Summarization of
  Soccer Games}. In \bibinfo{booktitle}{\emph{Proceedings Proceedings of the
  2nd International Workshop on Multimedia Content Analysis in Sports}}.
  \bibinfo{pages}{16--24}.
\newblock


\bibitem[\protect\citeauthoryear{Sanabria, Caglayan, Palaskar, Elliott,
  Barrault, Specia, and Metze}{Sanabria et~al\mbox{.}}{2018}]%
        {sanabria2018how2}
\bibfield{author}{\bibinfo{person}{Ramon Sanabria}, \bibinfo{person}{Ozan
  Caglayan}, \bibinfo{person}{Shruti Palaskar}, \bibinfo{person}{Desmond
  Elliott}, \bibinfo{person}{Lo{\"\i}c Barrault}, \bibinfo{person}{Lucia
  Specia}, {and} \bibinfo{person}{Florian Metze}.}
  \bibinfo{year}{2018}\natexlab{}.
\newblock \showarticletitle{How2: a large-scale dataset for multimodal language
  understanding}.
\newblock \bibinfo{journal}{\emph{arXiv preprint arXiv:1811.00347}}
  (\bibinfo{year}{2018}).
\newblock


\bibitem[\protect\citeauthoryear{Sawhney, Mathur, Mangal, Khanna, Shah, and
  Zimmermann}{Sawhney et~al\mbox{.}}{2020}]%
        {sawhney2020multimodal}
\bibfield{author}{\bibinfo{person}{Ramit Sawhney}, \bibinfo{person}{Puneet
  Mathur}, \bibinfo{person}{Ayush Mangal}, \bibinfo{person}{Piyush Khanna},
  \bibinfo{person}{Rajiv~Ratn Shah}, {and} \bibinfo{person}{Roger Zimmermann}.}
  \bibinfo{year}{2020}\natexlab{}.
\newblock \showarticletitle{Multimodal multi-task financial risk forecasting}.
  In \bibinfo{booktitle}{\emph{Proceedings of the 28th ACM International
  Conference on Multimedia}}. \bibinfo{pages}{456--465}.
\newblock


\bibitem[\protect\citeauthoryear{Sebastian and Puthiyidam}{Sebastian and
  Puthiyidam}{2015}]%
        {sebastian2015survey}
\bibfield{author}{\bibinfo{person}{Tinumol Sebastian} {and}
  \bibinfo{person}{Jiby~J Puthiyidam}.} \bibinfo{year}{2015}\natexlab{}.
\newblock \showarticletitle{A survey on video summarization techniques}.
\newblock \bibinfo{journal}{\emph{Int. J. Comput. Appl}} \bibinfo{volume}{132},
  \bibinfo{number}{13} (\bibinfo{year}{2015}), \bibinfo{pages}{30--32}.
\newblock


\bibitem[\protect\citeauthoryear{Sebe, Cohen, Gevers, and Huang}{Sebe
  et~al\mbox{.}}{2005}]%
        {sebe2005multimodal}
\bibfield{author}{\bibinfo{person}{Nicu Sebe}, \bibinfo{person}{Ira Cohen},
  \bibinfo{person}{Theo Gevers}, {and} \bibinfo{person}{Thomas~S Huang}.}
  \bibinfo{year}{2005}\natexlab{}.
\newblock \showarticletitle{Multimodal approaches for emotion recognition: a
  survey}. In \bibinfo{booktitle}{\emph{Internet Imaging VI}},
  Vol.~\bibinfo{volume}{5670}. International Society for Optics and Photonics,
  \bibinfo{pages}{56--67}.
\newblock


\bibitem[\protect\citeauthoryear{See, Liu, and Manning}{See
  et~al\mbox{.}}{2017}]%
        {DBLP:journals/corr/SeeLM17}
\bibfield{author}{\bibinfo{person}{Abigail See}, \bibinfo{person}{Peter~J.
  Liu}, {and} \bibinfo{person}{Christopher~D. Manning}.}
  \bibinfo{year}{2017}\natexlab{}.
\newblock \showarticletitle{Get To The Point: Summarization with
  Pointer-Generator Networks}.
\newblock \bibinfo{journal}{\emph{CoRR}}  \bibinfo{volume}{abs/1704.04368}
  (\bibinfo{year}{2017}).
\newblock
\showeprint[arxiv]{1704.04368}
\urldef\tempurl%
\url{http://arxiv.org/abs/1704.04368}
\showURL{%
\tempurl}


\bibitem[\protect\citeauthoryear{Shang, Kou, Zhang, and Wang}{Shang
  et~al\mbox{.}}{2022}]%
        {shang2022duo}
\bibfield{author}{\bibinfo{person}{Lanyu Shang}, \bibinfo{person}{Ziyi Kou},
  \bibinfo{person}{Yang Zhang}, {and} \bibinfo{person}{Dong Wang}.}
  \bibinfo{year}{2022}\natexlab{}.
\newblock \showarticletitle{A Duo-generative Approach to Explainable Multimodal
  COVID-19 Misinformation Detection}. In \bibinfo{booktitle}{\emph{Proceedings
  of the ACM Web Conference 2022}}. \bibinfo{pages}{3623--3631}.
\newblock


\bibitem[\protect\citeauthoryear{Shou, Wang, Chen, and Chen}{Shou
  et~al\mbox{.}}{2013}]%
        {shou2013sumblr}
\bibfield{author}{\bibinfo{person}{Lidan Shou}, \bibinfo{person}{Zhenhua Wang},
  \bibinfo{person}{Ke Chen}, {and} \bibinfo{person}{Gang Chen}.}
  \bibinfo{year}{2013}\natexlab{}.
\newblock \showarticletitle{Sumblr: continuous summarization of evolving tweet
  streams}. In \bibinfo{booktitle}{\emph{Proceedings of the 36th international
  ACM SIGIR conference on Research and development in information retrieval}}.
  \bibinfo{pages}{533--542}.
\newblock


\bibitem[\protect\citeauthoryear{Simonyan and Zisserman}{Simonyan and
  Zisserman}{2015}]%
        {Simonyan15}
\bibfield{author}{\bibinfo{person}{Karen Simonyan} {and}
  \bibinfo{person}{Andrew Zisserman}.} \bibinfo{year}{2015}\natexlab{}.
\newblock \showarticletitle{Very Deep Convolutional Networks for Large-Scale
  Image Recognition}. In \bibinfo{booktitle}{\emph{International Conference on
  Learning Representations}}.
\newblock


\bibitem[\protect\citeauthoryear{Singh, Nasery, Mehta, Agarwal, Lamba, and
  Srinivasan}{Singh et~al\mbox{.}}{2021}]%
        {singh2021mimoqa}
\bibfield{author}{\bibinfo{person}{Hrituraj Singh}, \bibinfo{person}{Anshul
  Nasery}, \bibinfo{person}{Denil Mehta}, \bibinfo{person}{Aishwarya Agarwal},
  \bibinfo{person}{Jatin Lamba}, {and} \bibinfo{person}{Balaji~Vasan
  Srinivasan}.} \bibinfo{year}{2021}\natexlab{}.
\newblock \showarticletitle{Mimoqa: Multimodal input multimodal output question
  answering}. In \bibinfo{booktitle}{\emph{Proceedings of the 2021 Conference
  of the North American Chapter of the Association for Computational
  Linguistics: Human Language Technologies}}. \bibinfo{pages}{5317--5332}.
\newblock


\bibitem[\protect\citeauthoryear{Sipos, Shivaswamy, and Joachims}{Sipos
  et~al\mbox{.}}{2012}]%
        {sipos2012large}
\bibfield{author}{\bibinfo{person}{Ruben Sipos}, \bibinfo{person}{Pannaga
  Shivaswamy}, {and} \bibinfo{person}{Thorsten Joachims}.}
  \bibinfo{year}{2012}\natexlab{}.
\newblock \showarticletitle{Large-margin learning of submodular summarization
  models}. In \bibinfo{booktitle}{\emph{Proceedings of the 13th Conference of
  the European Chapter of the Association for Computational Linguistics}}.
  \bibinfo{pages}{224--233}.
\newblock


\bibitem[\protect\citeauthoryear{Snelick, Uludag, Mink, Indovina, and
  Jain}{Snelick et~al\mbox{.}}{2005}]%
        {snelick2005large}
\bibfield{author}{\bibinfo{person}{Robert Snelick}, \bibinfo{person}{Umut
  Uludag}, \bibinfo{person}{Alan Mink}, \bibinfo{person}{Mike Indovina}, {and}
  \bibinfo{person}{Anil Jain}.} \bibinfo{year}{2005}\natexlab{}.
\newblock \showarticletitle{Large-scale evaluation of multimodal biometric
  authentication using state-of-the-art systems}.
\newblock \bibinfo{journal}{\emph{IEEE transactions on pattern analysis and
  machine intelligence}} \bibinfo{volume}{27}, \bibinfo{number}{3}
  (\bibinfo{year}{2005}), \bibinfo{pages}{450--455}.
\newblock


\bibitem[\protect\citeauthoryear{Soleymani, Garcia, Jou, Schuller, Chang, and
  Pantic}{Soleymani et~al\mbox{.}}{2017}]%
        {soleymani2017survey}
\bibfield{author}{\bibinfo{person}{Mohammad Soleymani}, \bibinfo{person}{David
  Garcia}, \bibinfo{person}{Brendan Jou}, \bibinfo{person}{Bj{\"o}rn Schuller},
  \bibinfo{person}{Shih-Fu Chang}, {and} \bibinfo{person}{Maja Pantic}.}
  \bibinfo{year}{2017}\natexlab{}.
\newblock \showarticletitle{A survey of multimodal sentiment analysis}.
\newblock \bibinfo{journal}{\emph{Image and Vision Computing}}
  \bibinfo{volume}{65} (\bibinfo{year}{2017}), \bibinfo{pages}{3--14}.
\newblock


\bibitem[\protect\citeauthoryear{Somasundaram and Alli}{Somasundaram and
  Alli}{2017}]%
        {somasundaram2017machine}
\bibfield{author}{\bibinfo{person}{SK Somasundaram} {and} \bibinfo{person}{P
  Alli}.} \bibinfo{year}{2017}\natexlab{}.
\newblock \showarticletitle{A machine learning ensemble classifier for early
  prediction of diabetic retinopathy}.
\newblock \bibinfo{journal}{\emph{Journal of Medical Systems}}
  \bibinfo{volume}{41}, \bibinfo{number}{12} (\bibinfo{year}{2017}),
  \bibinfo{pages}{1--12}.
\newblock


\bibitem[\protect\citeauthoryear{Specia}{Specia}{2018}]%
        {specia2018multi}
\bibfield{author}{\bibinfo{person}{Lucia Specia}.}
  \bibinfo{year}{2018}\natexlab{}.
\newblock \showarticletitle{Multi-modal Context Modelling for Machine
  Translation}.
\newblock  (\bibinfo{year}{2018}).
\newblock


\bibitem[\protect\citeauthoryear{Suman, Reddy, Saha, and Bhattacharyya}{Suman
  et~al\mbox{.}}{2020}]%
        {suman2020pay}
\bibfield{author}{\bibinfo{person}{Chanchal Suman},
  \bibinfo{person}{Saichethan~Miriyala Reddy}, \bibinfo{person}{Sriparna Saha},
  {and} \bibinfo{person}{Pushpak Bhattacharyya}.}
  \bibinfo{year}{2020}\natexlab{}.
\newblock \showarticletitle{Why pay more? A simple and efficient named entity
  recognition system for tweets}.
\newblock \bibinfo{journal}{\emph{Expert Systems with Applications}}
  (\bibinfo{year}{2020}), \bibinfo{pages}{114101}.
\newblock


\bibitem[\protect\citeauthoryear{Sun, Myers, Vondrick, Murphy, and Schmid}{Sun
  et~al\mbox{.}}{2019}]%
        {sun2019videobert}
\bibfield{author}{\bibinfo{person}{Chen Sun}, \bibinfo{person}{Austin Myers},
  \bibinfo{person}{Carl Vondrick}, \bibinfo{person}{Kevin Murphy}, {and}
  \bibinfo{person}{Cordelia Schmid}.} \bibinfo{year}{2019}\natexlab{}.
\newblock \showarticletitle{Videobert: A joint model for video and language
  representation learning}. In \bibinfo{booktitle}{\emph{Proceedings of the
  IEEE International Conference on Computer Vision}}.
  \bibinfo{pages}{7464--7473}.
\newblock


\bibitem[\protect\citeauthoryear{Szegedy, Liu, Jia, Sermanet, Reed, Anguelov,
  Erhan, Vanhoucke, and Rabinovich}{Szegedy et~al\mbox{.}}{2015}]%
        {szegedy2015going}
\bibfield{author}{\bibinfo{person}{Christian Szegedy}, \bibinfo{person}{Wei
  Liu}, \bibinfo{person}{Yangqing Jia}, \bibinfo{person}{Pierre Sermanet},
  \bibinfo{person}{Scott Reed}, \bibinfo{person}{Dragomir Anguelov},
  \bibinfo{person}{Dumitru Erhan}, \bibinfo{person}{Vincent Vanhoucke}, {and}
  \bibinfo{person}{Andrew Rabinovich}.} \bibinfo{year}{2015}\natexlab{}.
\newblock \showarticletitle{Going deeper with convolutions}. In
  \bibinfo{booktitle}{\emph{Proceedings of the IEEE conference on computer
  vision and pattern recognition}}. \bibinfo{pages}{1--9}.
\newblock


\bibitem[\protect\citeauthoryear{Takamura, Yokono, and Okumura}{Takamura
  et~al\mbox{.}}{2011}]%
        {takamura2011summarizing}
\bibfield{author}{\bibinfo{person}{Hiroya Takamura}, \bibinfo{person}{Hikaru
  Yokono}, {and} \bibinfo{person}{Manabu Okumura}.}
  \bibinfo{year}{2011}\natexlab{}.
\newblock \showarticletitle{Summarizing a document stream}. In
  \bibinfo{booktitle}{\emph{European conference on information retrieval}}.
  Springer, \bibinfo{pages}{177--188}.
\newblock


\bibitem[\protect\citeauthoryear{Tas and Kiyani}{Tas and Kiyani}{2007}]%
        {tas2007survey}
\bibfield{author}{\bibinfo{person}{Oguzhan Tas} {and} \bibinfo{person}{Farzad
  Kiyani}.} \bibinfo{year}{2007}\natexlab{}.
\newblock \showarticletitle{A survey automatic text summarization}.
\newblock \bibinfo{journal}{\emph{PressAcademia Procedia}} \bibinfo{volume}{5},
  \bibinfo{number}{1} (\bibinfo{year}{2007}), \bibinfo{pages}{205--213}.
\newblock


\bibitem[\protect\citeauthoryear{ter Hoeve, Kiseleva, and de~Rijke}{ter Hoeve
  et~al\mbox{.}}{2020}]%
        {ter2020makes}
\bibfield{author}{\bibinfo{person}{Maartje ter Hoeve}, \bibinfo{person}{Julia
  Kiseleva}, {and} \bibinfo{person}{Maarten de Rijke}.}
  \bibinfo{year}{2020}\natexlab{}.
\newblock \showarticletitle{What Makes a Good Summary? Reconsidering the Focus
  of Automatic Summarization}.
\newblock \bibinfo{journal}{\emph{arXiv preprint arXiv:2012.07619}}
  (\bibinfo{year}{2020}).
\newblock


\bibitem[\protect\citeauthoryear{Tiwari, Weth, and Kankanhalli}{Tiwari
  et~al\mbox{.}}{2018}]%
        {tiwari2018multimodal}
\bibfield{author}{\bibinfo{person}{Akanksha Tiwari}, \bibinfo{person}{Christian
  Von~Der Weth}, {and} \bibinfo{person}{Mohan~S Kankanhalli}.}
  \bibinfo{year}{2018}\natexlab{}.
\newblock \showarticletitle{Multimodal Multiplatform Social Media Event
  Summarization}.
\newblock \bibinfo{journal}{\emph{ACM Transactions on Multimedia Computing,
  Communications, and Applications (TOMM)}} \bibinfo{volume}{14},
  \bibinfo{number}{2s} (\bibinfo{year}{2018}), \bibinfo{pages}{1--23}.
\newblock


\bibitem[\protect\citeauthoryear{Tjondronegoro, Tao, Sasongko, and
  Lau}{Tjondronegoro et~al\mbox{.}}{2011}]%
        {tjondronegoro2011multi}
\bibfield{author}{\bibinfo{person}{Dian Tjondronegoro},
  \bibinfo{person}{Xiaohui Tao}, \bibinfo{person}{Johannes Sasongko}, {and}
  \bibinfo{person}{Cher~Han Lau}.} \bibinfo{year}{2011}\natexlab{}.
\newblock \showarticletitle{Multi-modal summarization of key events and top
  players in sports tournament videos}. In
  \bibinfo{booktitle}{\emph{Applications of Computer Vision (WACV), 2011 IEEE
  Workshop on}}. IEEE, \bibinfo{pages}{471--478}.
\newblock


\bibitem[\protect\citeauthoryear{Touvron, Cord, Sablayrolles, Synnaeve, and
  J{\'e}gou}{Touvron et~al\mbox{.}}{2021}]%
        {touvron2021going}
\bibfield{author}{\bibinfo{person}{Hugo Touvron}, \bibinfo{person}{Matthieu
  Cord}, \bibinfo{person}{Alexandre Sablayrolles}, \bibinfo{person}{Gabriel
  Synnaeve}, {and} \bibinfo{person}{Herv{\'e} J{\'e}gou}.}
  \bibinfo{year}{2021}\natexlab{}.
\newblock \showarticletitle{Going deeper with image transformers}.
\newblock \bibinfo{journal}{\emph{arXiv preprint arXiv:2103.17239}}
  (\bibinfo{year}{2021}).
\newblock


\bibitem[\protect\citeauthoryear{Tsai, Chen, Hu, and Chen}{Tsai
  et~al\mbox{.}}{2020}]%
        {tsai2020improving}
\bibfield{author}{\bibinfo{person}{Chih-Fong Tsai}, \bibinfo{person}{Kuanchin
  Chen}, \bibinfo{person}{Ya-Han Hu}, {and} \bibinfo{person}{Wei-Kai Chen}.}
  \bibinfo{year}{2020}\natexlab{}.
\newblock \showarticletitle{Improving text summarization of online hotel
  reviews with review helpfulness and sentiment}.
\newblock \bibinfo{journal}{\emph{Tourism Management}}  \bibinfo{volume}{80}
  (\bibinfo{year}{2020}), \bibinfo{pages}{104122}.
\newblock


\bibitem[\protect\citeauthoryear{UzZaman, Bigham, and Allen}{UzZaman
  et~al\mbox{.}}{2011}]%
        {uzzaman2011multimodal}
\bibfield{author}{\bibinfo{person}{Naushad UzZaman}, \bibinfo{person}{Jeffrey~P
  Bigham}, {and} \bibinfo{person}{James~F Allen}.}
  \bibinfo{year}{2011}\natexlab{}.
\newblock \showarticletitle{Multimodal summarization of complex sentences}. In
  \bibinfo{booktitle}{\emph{Proceedings of the 16th international conference on
  Intelligent user interfaces}}. ACM, \bibinfo{pages}{43--52}.
\newblock


\bibitem[\protect\citeauthoryear{Vasilyev and Bohannon}{Vasilyev and
  Bohannon}{2021}]%
        {Lita2005BLANCLE}
\bibfield{author}{\bibinfo{person}{Oleg Vasilyev} {and} \bibinfo{person}{John
  Bohannon}.} \bibinfo{year}{2021}\natexlab{}.
\newblock \showarticletitle{Is Human Scoring the Best Criteria for Summary
  Evaluation?}. In \bibinfo{booktitle}{\emph{Findings of the Association for
  Computational Linguistics: ACL-IJCNLP 2021}}. \bibinfo{publisher}{Association
  for Computational Linguistics}, \bibinfo{address}{Online}.
\newblock
\urldef\tempurl%
\url{https://doi.org/10.18653/v1/2021.findings-acl.192}
\showDOI{\tempurl}


\bibitem[\protect\citeauthoryear{Vaswani, Shazeer, Parmar, Uszkoreit, Jones,
  Gomez, Kaiser, and Polosukhin}{Vaswani et~al\mbox{.}}{2017}]%
        {vaswani2017attention}
\bibfield{author}{\bibinfo{person}{Ashish Vaswani}, \bibinfo{person}{Noam
  Shazeer}, \bibinfo{person}{Niki Parmar}, \bibinfo{person}{Jakob Uszkoreit},
  \bibinfo{person}{Llion Jones}, \bibinfo{person}{Aidan~N Gomez},
  \bibinfo{person}{{\L}ukasz Kaiser}, {and} \bibinfo{person}{Illia
  Polosukhin}.} \bibinfo{year}{2017}\natexlab{}.
\newblock \showarticletitle{Attention is all you need}. In
  \bibinfo{booktitle}{\emph{Advances in neural information processing
  systems}}. \bibinfo{pages}{5998--6008}.
\newblock


\bibitem[\protect\citeauthoryear{Verma, Jangra, Saha, Jatowt, and Roy}{Verma
  et~al\mbox{.}}{2022}]%
        {verma2022maked}
\bibfield{author}{\bibinfo{person}{Yash Verma}, \bibinfo{person}{Anubhav
  Jangra}, \bibinfo{person}{Sriparna Saha}, \bibinfo{person}{Adam Jatowt},
  {and} \bibinfo{person}{Dwaipayan Roy}.} \bibinfo{year}{2022}\natexlab{}.
\newblock \showarticletitle{MAKED: Multi-lingual Automatic Keyword Extraction
  Dataset}. In \bibinfo{booktitle}{\emph{Proceedings of the Thirteenth Language
  Resources and Evaluation Conference}}. \bibinfo{pages}{6170--6179}.
\newblock


\bibitem[\protect\citeauthoryear{Vulchanova, Vulchanov, Fritz, and
  Milburn}{Vulchanova et~al\mbox{.}}{2019}]%
        {vulchanova2019language}
\bibfield{author}{\bibinfo{person}{Mila Vulchanova}, \bibinfo{person}{Valentin
  Vulchanov}, \bibinfo{person}{Isabella Fritz}, {and} \bibinfo{person}{Evelyn~A
  Milburn}.} \bibinfo{year}{2019}\natexlab{}.
\newblock \showarticletitle{Language and perception: introduction to the
  special issue speakers and listeners in the visual world}.
\newblock \bibinfo{journal}{\emph{Journal of Cultural Cognitive Science}}
  (\bibinfo{year}{2019}), \bibinfo{pages}{1--10}.
\newblock


\bibitem[\protect\citeauthoryear{Wang, Li, and Lazebnik}{Wang
  et~al\mbox{.}}{2016}]%
        {wang2016learning}
\bibfield{author}{\bibinfo{person}{Liwei Wang}, \bibinfo{person}{Yin Li}, {and}
  \bibinfo{person}{Svetlana Lazebnik}.} \bibinfo{year}{2016}\natexlab{}.
\newblock \showarticletitle{Learning deep structure-preserving image-text
  embeddings}. In \bibinfo{booktitle}{\emph{Proceedings of the IEEE conference
  on computer vision and pattern recognition}}. \bibinfo{pages}{5005--5013}.
\newblock


\bibitem[\protect\citeauthoryear{Wang, Farhadi, Rao, and Brunton}{Wang
  et~al\mbox{.}}{2018}]%
        {wang2018ajile}
\bibfield{author}{\bibinfo{person}{Nancy~XR Wang}, \bibinfo{person}{Ali
  Farhadi}, \bibinfo{person}{Rajesh~PN Rao}, {and} \bibinfo{person}{Bingni~W
  Brunton}.} \bibinfo{year}{2018}\natexlab{}.
\newblock \showarticletitle{AJILE movement prediction: Multimodal deep learning
  for natural human neural recordings and video}. In
  \bibinfo{booktitle}{\emph{Thirty-Second AAAI Conference on Artificial
  Intelligence}}.
\newblock


\bibitem[\protect\citeauthoryear{Xu, Kong, and Zhang}{Xu et~al\mbox{.}}{2013}]%
        {xu2013cross}
\bibfield{author}{\bibinfo{person}{Shize Xu}, \bibinfo{person}{Liang Kong},
  {and} \bibinfo{person}{Yan Zhang}.} \bibinfo{year}{2013}\natexlab{}.
\newblock \showarticletitle{A cross-media evolutionary timeline generation
  framework based on iterative recommendation}. In
  \bibinfo{booktitle}{\emph{Proceedings of the 3rd ACM conference on
  International conference on multimedia retrieval}}. \bibinfo{pages}{73--80}.
\newblock


\bibitem[\protect\citeauthoryear{Yadav and Vishwakarma}{Yadav and
  Vishwakarma}{2020}]%
        {yadav2020deep}
\bibfield{author}{\bibinfo{person}{Ashima Yadav} {and}
  \bibinfo{person}{Dinesh~Kumar Vishwakarma}.} \bibinfo{year}{2020}\natexlab{}.
\newblock \showarticletitle{A Deep Multi-Level Attentive network for Multimodal
  Sentiment Analysis}.
\newblock \bibinfo{journal}{\emph{arXiv preprint arXiv:2012.08256}}
  (\bibinfo{year}{2020}).
\newblock


\bibitem[\protect\citeauthoryear{Yan, Wan, Lapata, Zhao, Cheng, and Li}{Yan
  et~al\mbox{.}}{2012}]%
        {yan2012visualizing}
\bibfield{author}{\bibinfo{person}{Rui Yan}, \bibinfo{person}{Xiaojun Wan},
  \bibinfo{person}{Mirella Lapata}, \bibinfo{person}{Wayne~Xin Zhao},
  \bibinfo{person}{Pu-Jen Cheng}, {and} \bibinfo{person}{Xiaoming Li}.}
  \bibinfo{year}{2012}\natexlab{}.
\newblock \showarticletitle{Visualizing timelines: Evolutionary summarization
  via iterative reinforcement between text and image streams}. In
  \bibinfo{booktitle}{\emph{Proceedings of the 21st ACM international
  conference on Information and knowledge management}}.
  \bibinfo{pages}{275--284}.
\newblock


\bibitem[\protect\citeauthoryear{Yao, Wan, and Xiao}{Yao et~al\mbox{.}}{2017}]%
        {yao2017recent}
\bibfield{author}{\bibinfo{person}{Jin-ge Yao}, \bibinfo{person}{Xiaojun Wan},
  {and} \bibinfo{person}{Jianguo Xiao}.} \bibinfo{year}{2017}\natexlab{}.
\newblock \showarticletitle{Recent advances in document summarization}.
\newblock \bibinfo{journal}{\emph{Knowledge and Information Systems}}
  \bibinfo{volume}{53}, \bibinfo{number}{2} (\bibinfo{year}{2017}),
  \bibinfo{pages}{297--336}.
\newblock


\bibitem[\protect\citeauthoryear{Young, Lai, Hodosh, and Hockenmaier}{Young
  et~al\mbox{.}}{2014}]%
        {young2014image}
\bibfield{author}{\bibinfo{person}{Peter Young}, \bibinfo{person}{Alice Lai},
  \bibinfo{person}{Micah Hodosh}, {and} \bibinfo{person}{Julia Hockenmaier}.}
  \bibinfo{year}{2014}\natexlab{}.
\newblock \showarticletitle{From image descriptions to visual denotations: New
  similarity metrics for semantic inference over event descriptions}.
\newblock \bibinfo{journal}{\emph{Transactions of the Association for
  Computational Linguistics}}  \bibinfo{volume}{2} (\bibinfo{year}{2014}),
  \bibinfo{pages}{67--78}.
\newblock


\bibitem[\protect\citeauthoryear{Yu, Jiang, Yang, and Xia}{Yu
  et~al\mbox{.}}{2020}]%
        {yu2020improving}
\bibfield{author}{\bibinfo{person}{Jianfei Yu}, \bibinfo{person}{Jing Jiang},
  \bibinfo{person}{Li Yang}, {and} \bibinfo{person}{Rui Xia}.}
  \bibinfo{year}{2020}\natexlab{}.
\newblock \showarticletitle{Improving multimodal named entity recognition via
  entity span detection with unified multimodal transformer}. Association for
  Computational Linguistics.
\newblock


\bibitem[\protect\citeauthoryear{Yu, Huang, Shi, and Zhu}{Yu
  et~al\mbox{.}}{2016}]%
        {yu2016product}
\bibfield{author}{\bibinfo{person}{Naitong Yu}, \bibinfo{person}{Minlie Huang},
  \bibinfo{person}{Yuanyuan Shi}, {and} \bibinfo{person}{Xiaoyan Zhu}.}
  \bibinfo{year}{2016}\natexlab{}.
\newblock \showarticletitle{Product review summarization by exploiting phrase
  properties}. In \bibinfo{booktitle}{\emph{Proceedings of COLING 2016, the
  26th International Conference on Computational Linguistics: Technical
  Papers}}. \bibinfo{pages}{1113--1124}.
\newblock


\bibitem[\protect\citeauthoryear{Zadeh, Chen, Poria, Cambria, and
  Morency}{Zadeh et~al\mbox{.}}{2017}]%
        {zadeh2017tensor}
\bibfield{author}{\bibinfo{person}{Amir Zadeh}, \bibinfo{person}{Minghai Chen},
  \bibinfo{person}{Soujanya Poria}, \bibinfo{person}{Erik Cambria}, {and}
  \bibinfo{person}{Louis-Philippe Morency}.} \bibinfo{year}{2017}\natexlab{}.
\newblock \showarticletitle{Tensor fusion network for multimodal sentiment
  analysis}.
\newblock \bibinfo{journal}{\emph{arXiv preprint arXiv:1707.07250}}
  (\bibinfo{year}{2017}).
\newblock


\bibitem[\protect\citeauthoryear{Zahavy, Magnani, Krishnan, and Mannor}{Zahavy
  et~al\mbox{.}}{2016}]%
        {zahavy2016picture}
\bibfield{author}{\bibinfo{person}{Tom Zahavy}, \bibinfo{person}{Alessandro
  Magnani}, \bibinfo{person}{Abhinandan Krishnan}, {and} \bibinfo{person}{Shie
  Mannor}.} \bibinfo{year}{2016}\natexlab{}.
\newblock \showarticletitle{Is a picture worth a thousand words? A Deep
  Multi-Modal Fusion Architecture for Product Classification in e-commerce}.
\newblock \bibinfo{journal}{\emph{arXiv preprint arXiv:1611.09534}}
  (\bibinfo{year}{2016}).
\newblock


\bibitem[\protect\citeauthoryear{Zhan, Loh, and Liu}{Zhan
  et~al\mbox{.}}{2009}]%
        {zhan2009gather}
\bibfield{author}{\bibinfo{person}{Jiaming Zhan}, \bibinfo{person}{Han~Tong
  Loh}, {and} \bibinfo{person}{Ying Liu}.} \bibinfo{year}{2009}\natexlab{}.
\newblock \showarticletitle{Gather customer concerns from online product
  reviews--A text summarization approach}.
\newblock \bibinfo{journal}{\emph{Expert Systems with Applications}}
  \bibinfo{volume}{36}, \bibinfo{number}{2} (\bibinfo{year}{2009}),
  \bibinfo{pages}{2107--2115}.
\newblock


\bibitem[\protect\citeauthoryear{Zhang, Gao, Zhang, Zhang, Tian, and
  Zimmermann}{Zhang et~al\mbox{.}}{2014}]%
        {zhang2014perception}
\bibfield{author}{\bibinfo{person}{Luming Zhang}, \bibinfo{person}{Yue Gao},
  \bibinfo{person}{Chao Zhang}, \bibinfo{person}{Hanwang Zhang},
  \bibinfo{person}{Qi Tian}, {and} \bibinfo{person}{Roger Zimmermann}.}
  \bibinfo{year}{2014}\natexlab{}.
\newblock \showarticletitle{Perception-guided multimodal feature fusion for
  photo aesthetics assessment}. In \bibinfo{booktitle}{\emph{Proceedings of the
  22nd ACM international conference on Multimedia}}. \bibinfo{pages}{237--246}.
\newblock


\bibitem[\protect\citeauthoryear{Zhang, Fu, Liu, and Huang}{Zhang
  et~al\mbox{.}}{2018}]%
        {zhang2018adaptive}
\bibfield{author}{\bibinfo{person}{Qi Zhang}, \bibinfo{person}{Jinlan Fu},
  \bibinfo{person}{Xiaoyu Liu}, {and} \bibinfo{person}{Xuanjing Huang}.}
  \bibinfo{year}{2018}\natexlab{}.
\newblock \showarticletitle{Adaptive Co-attention Network for Named Entity
  Recognition in Tweets.}. In \bibinfo{booktitle}{\emph{AAAI}}.
  \bibinfo{pages}{5674--5681}.
\newblock


\bibitem[\protect\citeauthoryear{Zhang*, Kishore*, Wu*, Weinberger, and
  Artzi}{Zhang* et~al\mbox{.}}{2020}]%
        {Zhang*2020BERTScore:}
\bibfield{author}{\bibinfo{person}{Tianyi Zhang*}, \bibinfo{person}{Varsha
  Kishore*}, \bibinfo{person}{Felix Wu*}, \bibinfo{person}{Kilian~Q.
  Weinberger}, {and} \bibinfo{person}{Yoav Artzi}.}
  \bibinfo{year}{2020}\natexlab{}.
\newblock \showarticletitle{BERTScore: Evaluating Text Generation with BERT}.
  In \bibinfo{booktitle}{\emph{International Conference on Learning
  Representations}}.
\newblock
\urldef\tempurl%
\url{https://openreview.net/forum?id=SkeHuCVFDr}
\showURL{%
\tempurl}


\bibitem[\protect\citeauthoryear{Zhao, Peyrard, Liu, Gao, Meyer, and Eger}{Zhao
  et~al\mbox{.}}{2019}]%
        {zhao-etal-2019-moverscore}
\bibfield{author}{\bibinfo{person}{Wei Zhao}, \bibinfo{person}{Maxime Peyrard},
  \bibinfo{person}{Fei Liu}, \bibinfo{person}{Yang Gao},
  \bibinfo{person}{Christian~M. Meyer}, {and} \bibinfo{person}{Steffen Eger}.}
  \bibinfo{year}{2019}\natexlab{}.
\newblock \showarticletitle{{M}over{S}core: Text Generation Evaluating with
  Contextualized Embeddings and Earth Mover Distance}. In
  \bibinfo{booktitle}{\emph{Proceedings of the 2019 Conference on Empirical
  Methods in Natural Language Processing and the 9th International Joint
  Conference on Natural Language Processing (EMNLP-IJCNLP)}}.
  \bibinfo{publisher}{Association for Computational Linguistics},
  \bibinfo{address}{Hong Kong, China}, \bibinfo{pages}{563--578}.
\newblock
\urldef\tempurl%
\url{https://doi.org/10.18653/v1/D19-1053}
\showDOI{\tempurl}


\bibitem[\protect\citeauthoryear{Zhou, Qiao, and Xiang}{Zhou
  et~al\mbox{.}}{2017}]%
        {zhou2017deep}
\bibfield{author}{\bibinfo{person}{Kaiyang Zhou}, \bibinfo{person}{Yu Qiao},
  {and} \bibinfo{person}{Tao Xiang}.} \bibinfo{year}{2017}\natexlab{}.
\newblock \showarticletitle{Deep reinforcement learning for unsupervised video
  summarization with diversity-representativeness reward}.
\newblock \bibinfo{journal}{\emph{arXiv preprint arXiv:1801.00054}}
  (\bibinfo{year}{2017}).
\newblock


\bibitem[\protect\citeauthoryear{Zhou, Palangi, Zhang, Hu, Corso, and Gao}{Zhou
  et~al\mbox{.}}{2020}]%
        {zhou2020unified}
\bibfield{author}{\bibinfo{person}{Luowei Zhou}, \bibinfo{person}{Hamid
  Palangi}, \bibinfo{person}{Lei Zhang}, \bibinfo{person}{Houdong Hu},
  \bibinfo{person}{Jason~J Corso}, {and} \bibinfo{person}{Jianfeng Gao}.}
  \bibinfo{year}{2020}\natexlab{}.
\newblock \showarticletitle{Unified Vision-Language Pre-Training for Image
  Captioning and VQA.}. In \bibinfo{booktitle}{\emph{AAAI}}.
  \bibinfo{pages}{13041--13049}.
\newblock


\bibitem[\protect\citeauthoryear{Zhu, Li, Liu, Zhou, Zhang, and Zong}{Zhu
  et~al\mbox{.}}{2018}]%
        {zhu2018msmo}
\bibfield{author}{\bibinfo{person}{Junnan Zhu}, \bibinfo{person}{Haoran Li},
  \bibinfo{person}{Tianshang Liu}, \bibinfo{person}{Yu Zhou},
  \bibinfo{person}{Jiajun Zhang}, {and} \bibinfo{person}{Chengqing Zong}.}
  \bibinfo{year}{2018}\natexlab{}.
\newblock \showarticletitle{MSMO: Multimodal Summarization with Multimodal
  Output}. In \bibinfo{booktitle}{\emph{Proceedings of the 2018 Conference on
  Empirical Methods in Natural Language Processing}}.
  \bibinfo{pages}{4154--4164}.
\newblock


\bibitem[\protect\citeauthoryear{Zhu, Zhou, Zhang, Li, Zong, and Li}{Zhu
  et~al\mbox{.}}{2020a}]%
        {zhu3multimodal}
\bibfield{author}{\bibinfo{person}{Junnan Zhu}, \bibinfo{person}{Yu Zhou},
  \bibinfo{person}{Jiajun Zhang}, \bibinfo{person}{Haoran Li},
  \bibinfo{person}{Chengqing Zong}, {and} \bibinfo{person}{Changliang Li}.}
  \bibinfo{year}{2020}\natexlab{a}.
\newblock \showarticletitle{Multimodal Summarization with Guidance of
  Multimodal Reference}. In \bibinfo{booktitle}{\emph{Proceedings of the AAAI
  Conference on Artificial Intelligence}}, Vol.~\bibinfo{volume}{34}.
  \bibinfo{pages}{9749--9756}.
\newblock


\bibitem[\protect\citeauthoryear{Zhu, Zhou, Zhang, Li, Zong, and Li}{Zhu
  et~al\mbox{.}}{2020b}]%
        {zhu2020multimodal}
\bibfield{author}{\bibinfo{person}{Junnan Zhu}, \bibinfo{person}{Yu Zhou},
  \bibinfo{person}{Jiajun Zhang}, \bibinfo{person}{Haoran Li},
  \bibinfo{person}{Chengqing Zong}, {and} \bibinfo{person}{Changliang Li}.}
  \bibinfo{year}{2020}\natexlab{b}.
\newblock \showarticletitle{Multimodal summarization with guidance of
  multimodal reference}. In \bibinfo{booktitle}{\emph{Proceedings of the AAAI
  Conference on Artificial Intelligence}}, Vol.~\bibinfo{volume}{34}.
  \bibinfo{pages}{9749--9756}.
\newblock


\bibitem[\protect\citeauthoryear{Zhuge}{Zhuge}{2016}]%
        {zhuge2016multi}
\bibfield{author}{\bibinfo{person}{Hai Zhuge}.}
  \bibinfo{year}{2016}\natexlab{}.
\newblock \bibinfo{booktitle}{\emph{Multi-dimensional summarization in
  cyber-physical society}}.
\newblock \bibinfo{publisher}{Morgan Kaufmann}.
\newblock


\bibitem[\protect\citeauthoryear{Zlatintsi, Koutras, Evangelopoulos,
  Malandrakis, Efthymiou, Pastra, Potamianos, and Maragos}{Zlatintsi
  et~al\mbox{.}}{2017}]%
        {zlatintsi2017cognimuse}
\bibfield{author}{\bibinfo{person}{Athanasia Zlatintsi},
  \bibinfo{person}{Petros Koutras}, \bibinfo{person}{Georgios Evangelopoulos},
  \bibinfo{person}{Nikolaos Malandrakis}, \bibinfo{person}{Niki Efthymiou},
  \bibinfo{person}{Katerina Pastra}, \bibinfo{person}{Alexandros Potamianos},
  {and} \bibinfo{person}{Petros Maragos}.} \bibinfo{year}{2017}\natexlab{}.
\newblock \showarticletitle{COGNIMUSE: A multimodal video database annotated
  with saliency, events, semantics and emotion with application to
  summarization}.
\newblock \bibinfo{journal}{\emph{EURASIP Journal on Image and Video
  Processing}} \bibinfo{volume}{2017}, \bibinfo{number}{1}
  (\bibinfo{year}{2017}), \bibinfo{pages}{1--24}.
\newblock


\end{thebibliography}

\end{document}